\documentclass[lettersize,journal]{IEEEtran}
\usepackage{amsmath,amsfonts}
\usepackage{algorithmic}
\usepackage{array}
\usepackage[caption=false,font=normalsize,labelfont=sf,textfont=sf]{subfig}
\usepackage{textcomp}
\usepackage{stfloats}
\usepackage{url}
\usepackage{verbatim}
\usepackage{graphicx}

\def\BibTeX{{\rm B\kern-.05em{\sc i\kern-.025em b}\kern-.08em
    T\kern-.1667em\lower.7ex\hbox{E}\kern-.125emX}}
\usepackage{balance}

\usepackage{mathtools}
\usepackage{tabularx}
\usepackage[numbers]{natbib}
\newcommand{\textvareps}{ \scalebox{1.5}{$\epsilon$}}

\usepackage{amsmath, amsfonts, amssymb}
\usepackage{array}
\usepackage{booktabs}
\usepackage{enumitem}
\usepackage{flushend}
\usepackage{graphicx}
\usepackage{microtype}
\usepackage{multirow}
\usepackage{nicefrac}
\usepackage{printlen}
\usepackage{soul}
\usepackage{svg}
\usepackage{wrapfig}
\usepackage{lipsum}
\usepackage{xcolor}
\usepackage{lipsum}

\usepackage{framed}
\colorlet{shadecolor}{blue!20}

\newcommand{\rulesep}{\unskip\ \vrule\ }

\newcommand{\com}[1]{\iffalse #1 \fi}%

\newcommand{\noimage}{%
  \setlength{\fboxsep}{-\fboxrule}%
  \fbox{\phantom{\rule{100pt}{100pt}}File missing\phantom{\rule{100pt}{100pt}}}
}
\let\includegraphicsoriginal\includegraphics
\renewcommand{\includegraphics}[2][width=\textwidth]{\IfFileExists{#2}{\includegraphicsoriginal[#1]{#2}}{\noimage}}

\newcolumntype{+}{>{\global\let\currentrowstyle\relax}}
\newcolumntype{^}{>{\currentrowstyle}}

\newenvironment{aequation}
{\begin{equation} \begin{aligned}}
{\end{aligned} \end{equation}}

\emergencystretch=\maxdimen
\everypar{\looseness=-1 }

\newcounter{descriptcount}

\definecolor{Overlapped}{RGB}{180, 22 , 0   }
\definecolor{Fractured}{RGB}{23 , 77 , 127 }
\definecolor{Clustered}{RGB}{55 , 144, 48  }
\definecolor{hl}{RGB}{255,20,147}

\definecolor{yellow}{HTML}{FF7F00} 
\definecolor{green}{HTML}{4D06AF}
\definecolor{pink}{HTML}{93256B}

\usepackage{tikz}

\newrobustcmd*{\mysquare}[1]{\tikz{\filldraw[draw=#1,fill=#1] (0,0)
rectangle (0.2cm,0.2cm);}}

\newrobustcmd*{\mycircle}[1]{\tikz{\filldraw[draw=#1,fill=#1] (0,0) circle [radius=0.1cm];}}

\newrobustcmd*{\mytriangle}[1]{\tikz{\filldraw[draw=#1,fill=#1] (0,0) --
(0.2cm,0) -- (0.1cm,0.2cm);}}

\newcommand{\Fractured}{\textcolor{Fractured}{$\bigstar$}}
\newcommand{\Overlapped}{\textcolor{Overlapped}{$\clubsuit$}}
\newcommand{\Clustered}{\textcolor{Clustered}{$\spadesuit$}}

\newcommand{\PatternA}{\textcolor{Fractured}{\textbf{Pattern~A}~(\Fractured)}}
\newcommand{\PatternB}{\textcolor{Overlapped}{\textbf{Pattern~B}~(\Overlapped)}}
\newcommand{\PatternC}{\textcolor{Clustered}{\textbf{Pattern~C}~(\Clustered)}}

\newcommand{\overlapped}{\textcolor{Overlapped}{\textbf{Overlapped}}}
\newcommand{\clustered}{\textcolor{Clustered}{\textbf{Clustered}}}
\newcommand{\fractured}{\textcolor{Fractured}{\textbf{Fractured}}}

\newcommand{\ImageNet}{ImageNet\nobreakdash-1k}
\newcommand{\SubImageNet}{$16$\nobreakdash-class\nobreakdash-ImageNet}

\sethlcolor{hl}
\makeatletter
\def\SOUL@hlpreamble{%
    \setul{\dp\strutbox}{\dimexpr\ht\strutbox+\dp\strutbox\relax}%
    \let\SOUL@stcolor\SOUL@hlcolor
    \SOUL@stpreamble
}
\makeatother

{\color{hl}}%
{}

\definecolor{Gray}{gray}{0.85}
\definecolor{cite}{HTML}{53769A}
\definecolor{ref}{HTML}{379030}
\definecolor{lightcornflowerblue}{rgb}{0.6, 0.81, 0.93}
\definecolor{lightkhaki}{rgb}{0.94, 0.9, 0.55}
\definecolor{lightmauve}{rgb}{0.86, 0.82, 1.0}
\definecolor{lightgreen}{rgb}{0.56, 0.93, 0.56}

\definecolor{royalpurple}{RGB}{207,199,216}
\definecolor{forestgreen}{RGB}{202,225,204}
\usepackage[colorlinks=true, citecolor=cite, linkcolor=ref]{hyperref}

\usepackage{collcell}
\usepackage{xstring}
\usepackage{datatool}
\usepackage{booktabs}
\usepackage{environ}

\newcounter{CurrentRow}
\newcounter{CurrentColumn}
\newtoggle{DoneWithFirstRow}

\newcommand*{\FirstColumn}[1]{%
    \IfEq{\arabic{CurrentColumn}}{0}{%
        \global\togglefalse{DoneWithFirstRow}%
        \setcounter{CurrentRow}{1}
    }{%
        \global\toggletrue{DoneWithFirstRow}%
        \stepcounter{CurrentRow}%
    }%
    \setcounter{CurrentColumn}{0}%
    \NewData{#1}%
}
\newcommand*{\NewData}[1]{%
    \dtlexpandnewvalue%
    \stepcounter{CurrentColumn}%
    \iftoggle{DoneWithFirstRow}{%
        \dtlgetrow{TransposedTabularDB}{\arabic{CurrentColumn}}%
        \dtlappendentrytocurrentrow{\Alph{CurrentRow}}{#1}%
        \dtlrecombine%
    }{%
        \DTLnewrow{TransposedTabularDB}%
        \DTLnewdbentry{TransposedTabularDB}{\Alph{CurrentRow}}{#1}%
    }%
}%

\newcolumntype{F}{>{\collectcell\FirstColumn}c<{\endcollectcell}}
\newcolumntype{C}{>{\collectcell\NewData}{c}<{\endcollectcell}}

\newtoggle{EncounteredDataRow}

\newsavebox{\TempBox}
\DTLnewdb{TransposedTabularDB}

\NewEnviron{Ttabular}[1]{%
    \setcounter{CurrentColumn}{0}%
    \global\togglefalse{EncounteredDataRow}%
    \savebox{\TempBox}{%
        \begin{tabular}{FCCCCCC}
            \BODY%
        \end{tabular}%
    }%
    \begin{tabular}{#1}\toprule%
    \DTLforeach*{TransposedTabularDB}{\Aa=A, \Ba=B, \Ca=C}{%
      \DTLiffirstrow{}{\\\midrule}%
        \Aa & \Ba & \Ca %
    }\\\bottomrule%
    \end{tabular}%
}%

\newcolumntype{H}{>{\setbox0=\hbox\bgroup}c<{\egroup}@{}}

\newcommand{\figleft}{{\em (Left)}}

\newcommand{\figright}{{\em (Right)}}

\newcommand{\figtopleft}{{\em (Top Left)}}
\newcommand{\figbottomright}{{\em (Bottom Right)}}
\newcommand{\figtopright}{{\em (Top Right)}}
\newcommand{\figbottomleft}{{\em (Bottom Left)}}

\usepackage{mathabx,graphicx}

\def\CircleArrowright{\ensuremath{%
  \reflectbox{\rotatebox[origin=c]{180}{$\circlearrowright$}}}}

\def\Figref#1{Figure~\ref{#1}}

\def\Tableref#1{Table~\ref{#1}}

\def\Twotablesref#1#2{Tables \ref{#1} and \ref{#2}}

\def\eqref#1{equation~\ref{#1}}

\def\1{\bm{1}}

\def\eps{{\epsilon}}

\DeclareMathAlphabet{\mathsfit}{\encodingdefault}{\sfdefault}{m}{sl}
\SetMathAlphabet{\mathsfit}{bold}{\encodingdefault}{\sfdefault}{bx}{n}

\def\gN{{\mathcal{N}}}

\DeclareMathOperator*{\argmin}{arg\,min}

\usepackage{orcidlink}

\begin{document}
\title{{k*~Distribution}: Evaluating the Latent Space of Deep Neural Networks using Local Neighborhood Analysis}

\author{
    Shashank Kotyan~
    \orcidlink{0000-0003-1235-3316},~
    \IEEEmembership{Student Member, IEEE}, 
    Tatsuya Ueda, 
    and 
    Danilo Vasconcellos Vargas 
    
    \thanks{
         This work was supported by JSPS Grant-in-Aid for Challenging Exploratory Research - Grant Number JP22K19814, JST Strategic Basic Research Promotion Program (AIP Accelerated Research) - Grant Number JP22584686, JSPS Research on Academic Transformation Areas (A) - Grant Number JP22H05194. \textit{(Shashank Kotyan and Ueda Tatsuya contributed equally to this work) (Corresponding Author: Shashank Kotyan)}
         }\thanks{
         Shashank Kotyan and Danilo Vasconcellos Vargas are with the Laboratory of Intelligent Systems, Kyushu University, Fukuoka, Japan. 
         Tatsuya Ueda is with SoftBank Group Corporation, Tokyo, Japan.
         Danilo Vasconcellos Vargas is also with Department of Electrical Engineering and Information Systems, School of Engineering, The University of Tokyo, Tokyo, Japan and MiraiX.
         }\thanks{
         Project Website is available online at \href{https://shashankkotyan.github.io/k-Distribution/}{https://shashankkotyan.github.io/k-Distribution/}.
         
    }
}

\maketitle

\begin{abstract}

    Most examinations of neural networks' learned latent spaces typically employ dimensionality reduction techniques such as t-SNE or UMAP.
    These methods distort the local neighborhood in the visualization, making it hard to distinguish the structure of a subset of samples in the latent space.
    In response to this challenge, we introduce the {k*~distribution} and its corresponding visualization technique
    This method uses local neighborhood analysis to guarantee the preservation of the structure of sample distributions for individual classes within the subset of the learned latent space. 
    This facilitates easy comparison of different {k*~distribution}s, enabling analysis of how various classes are processed by the same neural network.
    Our study reveals three distinct distributions of samples within the learned latent space subset:
        a) Fractured, 
        b) Overlapped, and 
        c) Clustered, 
    providing a more profound understanding of existing contemporary visualizations.
    Experiments show that the distribution of samples within the network's learned latent space significantly varies depending on the class.
    Furthermore, we illustrate that our analysis can be applied to explore the latent space of
        diverse neural network architectures,
        various layers within neural networks,
        transformations applied to input samples,
        and the distribution of training and testing data
    for neural networks.
    Thus, the k* distribution should aid in visualizing the structure inside neural networks and further foster their understanding.
    

    
\end{abstract}

\begin{IEEEkeywords}
Neural Networks,
Latent Space Visualization,
Local Neighborhood Analysis, 
Class Representation, 
Cluster Analysis
\end{IEEEkeywords}

\section{Introduction}

    \begin{figure}[!ht]
        \centering
        \includegraphics[width=\linewidth]{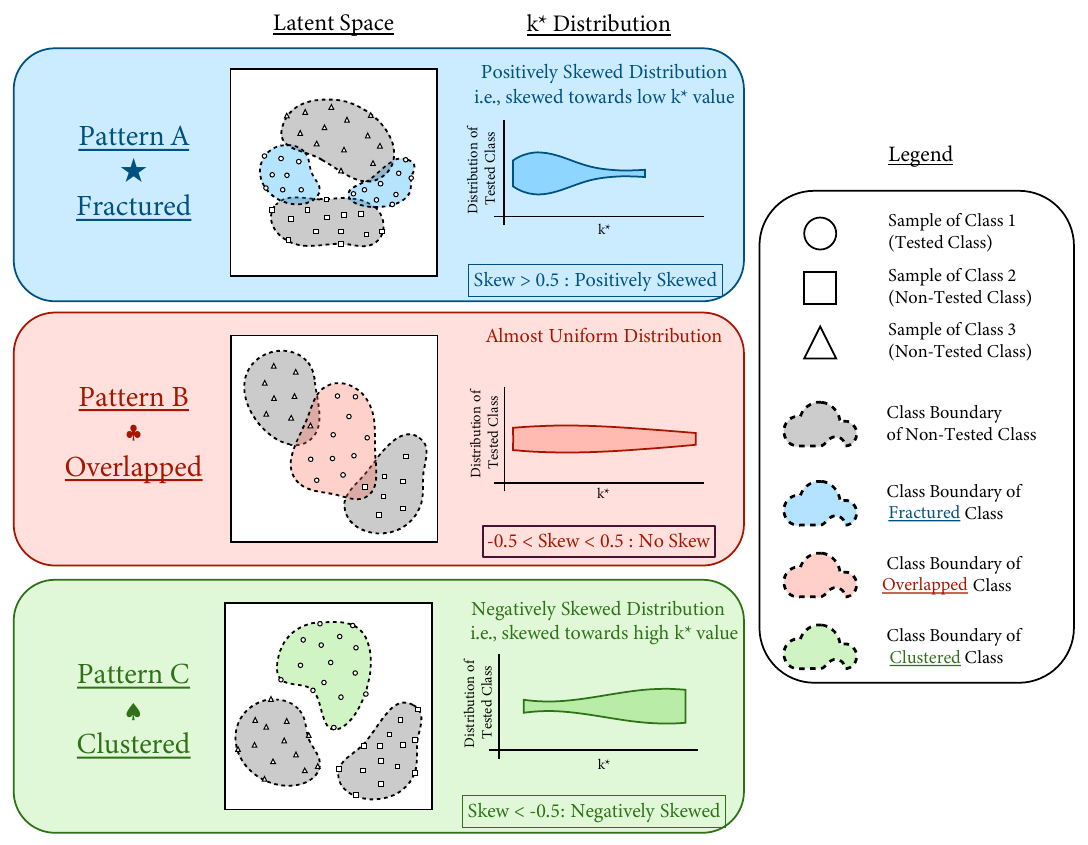}
        \caption{
        Overview of three distinct basic patterns of {k*~Distribution}.
        Here, we define the k* value of a sample point as the k\textsuperscript{th}-closest neighbor, which differs in class compared to the test point, i.e., the neighbor (sample) which breaks homogeneity in the local neighborhood of the test point.
        \PatternA~which has positively skewed {k*~distribution} (skewed towards low k* value) representing  an `\fractured' distribution of samples in latent space;
        \PatternB~which has almost uniform {k*~distribution} representing a `\overlapped' distribution of samples in latent space;
        \PatternC~which has negatively skewed {k*~distribution} (skewed towards high k* value) representing a `\clustered' distribution of samples in latent space.
        }
        \label{fig:illustration}
    \end{figure}

    \IEEEPARstart{A} significant portion of neural network research relies on creating tools to comprehend the acquired latent space and unveil the inner workings of neural networks, often viewed as black boxes.
    Nevertheless, if researchers are equipped with the essential tools to grasp the learned latent space, it becomes feasible to delve deeper, uncovering insights and reasons that can guide further research in neural networks.
    Analyzing the neural network's latent space poses challenges due to its intricate non-convex characteristics.
    However, directly evaluating and comparing the configuration and distribution of data within high-dimensional latent spaces persists despite the development of various visualization tools.

    Existing tools often rely on dimensionality reduction techniques like t-SNE \cite{maaten2008visualizing} and UMAP \cite{2018arXivUMAP} to create $2$ or $3$-dimensional scatter plots for individual latent spaces, offering a broad overview.
    They rely on dimensionality reduction methods rooted in manifold learning, which considers specific manifold characteristics and preserves them while modifying other factors and attributes.

    Analyzing the configurations and structures of various sample distributions associated with a specific class within the latent space presents a challenge using existing visualization methods.
    This challenge extends to the comparison of multiple sample distributions, limiting the analysis of the latent spaces.
    Additionally, visually comparing multiple latent spaces side by side can be overwhelming and confusing, especially when dealing with numerous points and varying degrees of distortion  \cite{gleicher2011Visual,arendt2020Parallel,cutura2020Comparing,boggust2022Embedding,sivaraman2022Emblaze}.

    If there is too much information kept, it is hard to understand.
    At the same time, if the information is filtered enough, some perspective is always lacking.
    Therefore, some complementary tools might be crucial in interpreting these complex latent spaces.

    \begin{figure*}[!ht]
        \centering
        \includegraphics[width=\linewidth]{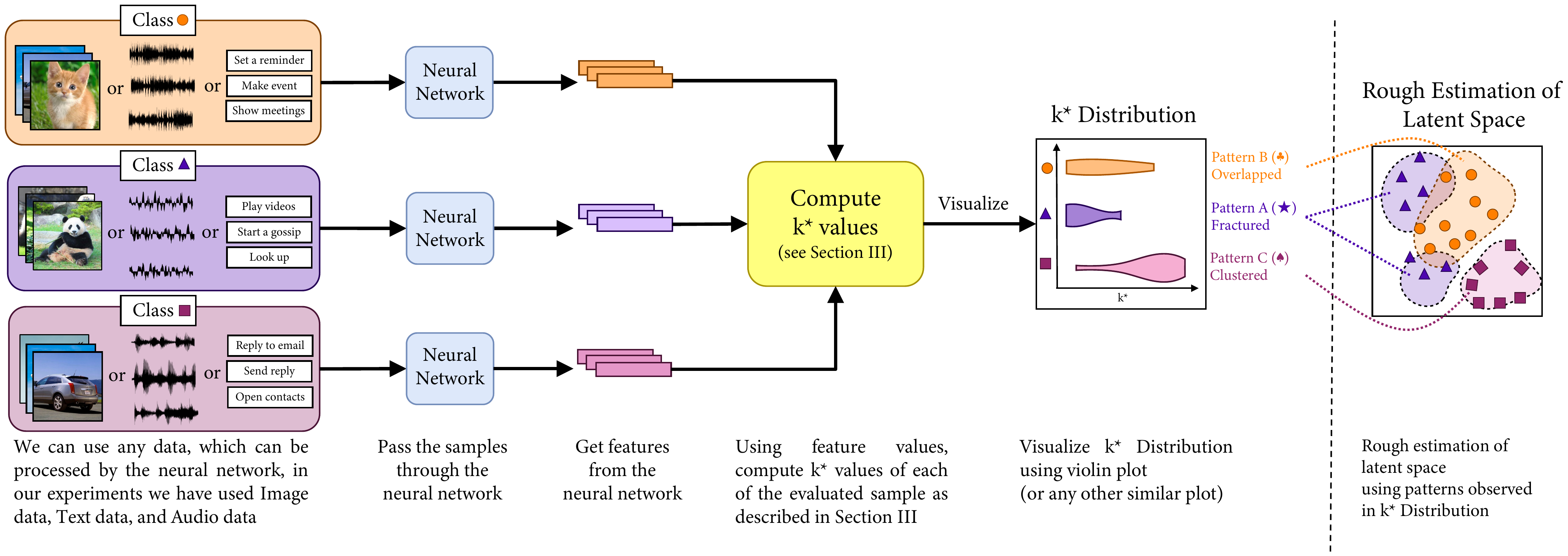}
        \caption{
        Overview of the framework to create {k*~Distribution}.
        We use the learned features of a neural network to compute k* values of individual evaluated samples and then compute the {k*~distribution} for a particular class.
        }
        \label{fig:framework}
    \end{figure*}

    This article introduces a tool designed to enhance understanding of the distribution of samples associated with `classes/features' in the learned latent space generated by neural networks.
    The proposed approach involves leveraging the local neighborhood relationships among these samples in the latent space.
    More precisely, the nearest-neighbor method is applied to the learned latent space to identify a k*-nearest neighbor (sample) within the local neighborhood featuring a different class from the test sample.
    This process evaluates the disruption of the homogeneity within the local neighborhood of the tested sample.
    Subsequently, a distribution of k* values referred to as the {k*~distribution} is generated, incorporating all the samples belonging to a given class.
    Through an analysis of the {k*~distribution}, three distinct patterns of distribution of samples in the learned latent space are identified as illustrated in \Figref{fig:illustration}:
    
    \begin{description}[leftmargin=*] 
        
        \item \PatternA~representing \fractured~distribution of samples in latent space,
        \item \PatternB~representing \overlapped~distribution of samples in latent space, or
        \item \PatternC~representing \clustered~distribution of samples in latent space.
        
    \end{description}
    
\begin{figure*}[!ht]
        \centering
        \includegraphics[width=\linewidth]{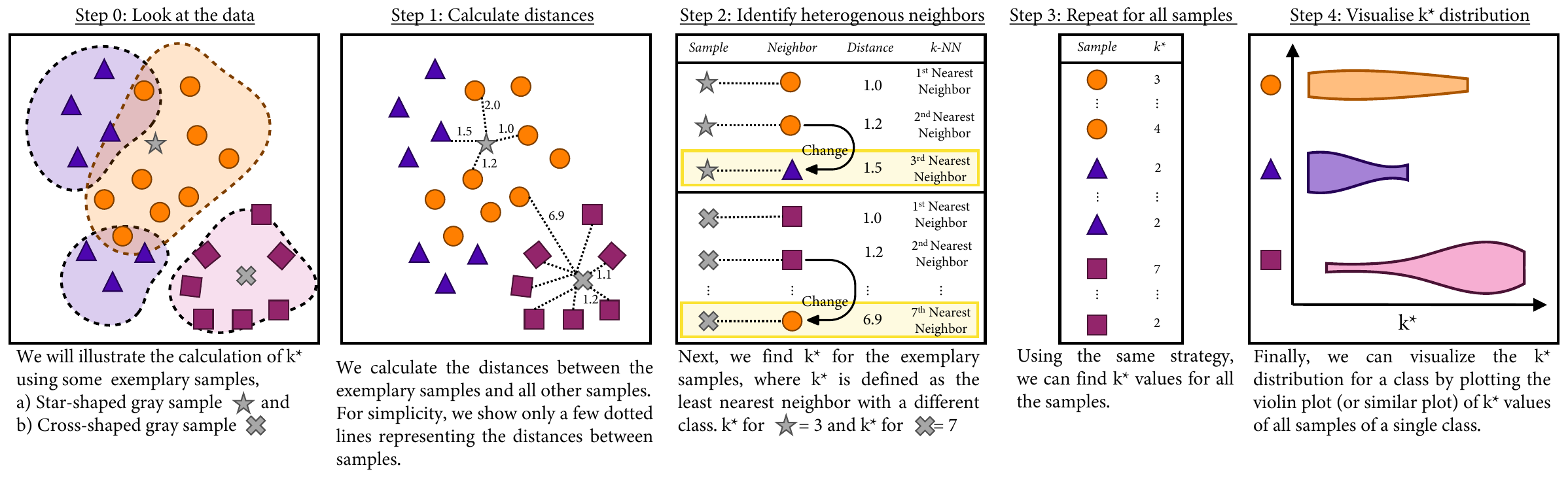}
        \caption{
        Illustration of calculating k* value of a sample and correspondingly {k*~distribution} of class.
        For all the samples, in the evaluating data, first find the index of the nearest neighbor that differs in class, i.e., where the local homogeneity of the neighbors breaks.
        We call this as k* value of the sample.
        Further, one can gather k* values for all the samples belonging to a single class and plot its distribution.
        In our example, note that the distribution of k* values for all the samples of \mycircle{yellow} class is almost uniform, corresponding to \PatternB~\overlapped~distribution of samples in latent space.
        Similarly, the distribution of k* values for all the samples of \mytriangle{green} is positively skewed, which corresponds to \PatternA~\fractured~distribution of samples in latent space, and the distribution of k* values for all the samples of \mysquare{pink} is negatively skewed which corresponds to \PatternC~\clustered~distribution of samples in latent space.
        }
        \label{fig:explanation}
    \end{figure*}

    \noindent
    \textbf{Contributions:}
    This article provides,
    \begin{description}[font=$\bullet~$, leftmargin=*] 
        \item A new interpretation of latent space learned by the neural network, relying on local neighborhood relationships and homogeneity.
        \item Identification of various distribution patterns of samples in the latent space based on neighborhood characteristics (see \Figref{fig:illustration}).
        \item A model-agnostic latent space analysis of neural networks, focusing on samples from a single class (see \Figref{fig:framework}).
        \item A method for straightforwardly comparing different classes and understanding how samples from various classes are distributed in the learned latent space (see \Figref{fig:explanation}).
    \end{description}

\section{Related Works}

    \subsection{Visualizing the Latent Space of Neural Networks using Dimensionality Reduction Techniques:}
    Given that neural networks operate in a high-dimensional space, visualizing their latent space directly is challenging.
    Various researchers employ dimensionality reduction techniques to overcome this limitation to represent the latent space in 2 or 3 dimensions.
    There are a lot of techniques for visualizing the latent space, like, \cite{maaten2008visualizing,2018arXivUMAP,hotelling1933analysis,kruskal1964multidimensional,sammon1969nonlinear,tenenbaum2000global,belkin2001laplacian,coifman2006diffusion,van2009dimensionality,villegas2011dimensionality,timofte2012iterative,im2013GPLOM,gleicher2013Explainers,kramer2015unsupervised,kim2016InterAxis,smilkov2016Embedding,tang2016visualizing,kahng2018ActiVis,li2018EmbeddingVis,wei2018reconstructible,dowling2019SIRIUS,liu2019-lsc,ovchinnikova2020Exploring,hohman2020Summit,amid2022TriMap}.
    Many of these algorithms require hyperparameter tuning for optimal visualization, making it challenging to fairly compare latent spaces of varying dimensions.
    Additionally, while these visualizations effectively capture some perspective on the structure of the learned latent space, using them to draw a clear comparison between different local structures is challenging \cite{gleicher2011Visual,arendt2020Parallel,cutura2020Comparing,boggust2022Embedding,sivaraman2022Emblaze}.

    The effectiveness of dimensionality reduction techniques becomes apparent in instances when the latent space is well-organized and has successfully assimilated the intended information.
    In such cases, these visualizations demonstrate utility by aligning with pre-established interpretations.
    Conversely, when applied to latent space lacking clearly defined pre-established knowledge, the efficacy diminishes for such visualizations where the lack of known structure leads to a `blob of points' in the visualization \cite{sivaraman2022Emblaze}.

    \subsection{Visualizing Association of Features in the Latent Space of Neural Network:}

    Various approaches employ visualizations to understand how neural networks interact with features.
    Analyzing the responses of units in hidden layers to features provides insights into the learned latent space, elucidating which features the network prioritizes \cite{visualizing_activation_1, visualizing_activation_2}.
    This involves evaluating individual unit responses or combinations of specific inputs \cite{activation-atlas}.

    Another visualization method focuses on understanding neural network attention, revealing which parts of an image are emphasized.
    Salient regions in the input that influence the network's output are identified using saliency maps \cite{itti1998model} and gradient-weighted class-activation maps \cite{selvaraju2017grad}.
    These maps quantify conspicuity at each visual field location, guiding the selection of attended locations based on saliency distribution.

    Additionally, interpretable concepts, defined as groups of latent variables in the space that are meaningful, may manifest in the latent space, further contributing to our understanding \cite{visualizing_activation_1,visualizing_activation_2,unit_features1,unit_features3,unit_features5}.
    In natural language processing, these interpretations often provide insight into learning semantic and linguistic concepts provided by the lexical object \cite{dalvi2021Discoveringa,durrani2022Transformationa,anelli2022Interpretability}.
    These interpretations are often effectively visualized using analytic systems like \cite{liu2019NLIZE,strobelt2019Seq2seqVis,vig2019Multiscale,park2019SANVis,hoover2020exBERT,chauhan2022BERTops,sevastjanova2023Visual}; however, the challenge to draw a clear comparison between different local structures and multiple latent spaces persists.

    We have seen a few methods significantly contribute to the understanding and visualization of latent spaces using neighborhood-based analyses \cite{bressan2003nonparametric,goldberger2004neighbourhood,weinberger2009distance,pang2019simultaneously}, facilitating the development of more sophisticated models and techniques. 
    They have paved the way for metric learning in neural networks, a domain that focuses on learning distance metrics directly from data \cite{gao2014dimensionality,plotz2018neural,chen2021neighborhood,huang2021neighbor2neighbor,lee2022knn}. 
    
    We suggest employing the {k*~distribution} to examine the distribution of samples in the neighborhood of a specific sample and analyze the distribution of samples sharing a predefined label.
    Our proposed approach presents a framework for effectively comparing latent spaces and sub-spaces with diverse sample distributions (see \Figref{fig:framework}).
    This framework enables meaningful comparisons between the distribution of samples belonging to different classes and multiple latent spaces, offering additional insights into latent spaces beyond what existing visualizations provide.
    We hope that our proposed visualization technique using {k*~distribution} can further the advances by providing more insightful analyses of the latent space. 



\section{{K*~Distribution}: Analyzing Homogeneity in the Local Neighborhood of Samples}

    We propose a methodology based on the local neighborhood to analyze the hyper-dimensional latent space learned by neural networks (see \Figref{fig:framework}).
    The distribution of samples and clusters in the learned latent space is analyzed by associating them with classes.
    Through this approach, we gain insights into the distribution patterns within the learned latent space and identify the clusters formed (see \Figref{fig:illustration}), thereby enhancing our comprehension of the latent space.


    In this context, we introduce the {k*~distribution} by exploring the concept of neighborhoods and their utility in validating the local relationship among features in latent space.
    The {k*~distribution} is constructed by taking a latent space as input and analyzing the neighborhood of a sample (see \Figref{fig:explanation}).

    The Nearest Neighbor method, a widely recognized non-parametric technique, enables us to understand the positioning of latent variables in space near a latent variable and their corresponding class distribution.
    This method aids in gauging the relative distances between latent variables and grouping them into clusters.
    The underlying principle is grounded in the notion that local neighborhoods offer a reasonable estimate of the sample distribution within the latent space.

    The index of the k\textsuperscript{th} nearest neighbor that belongs to a different class than the test sample, disrupting the uniformity of the test sample's neighborhood, is quantified and  termed as the k* value.
    Essentially, a high k* value indicates that the sample is surrounded by similar points that share the same class.
    In other words, a high k* value indicates that a neighbor of a different class will be situated a considerable distance from the neighborhood of the specified sample.
    In particular, the homogeneity of each class cluster within the learned latent space can be assessed using k* (see \Figref{fig:explanation}).
    Additionally, one can ascertain whether a cluster remains cohesive for a given class or if it is fragmented across various spatial locations.
    
    The measurement of the nearest neighbor index allows us to address the sparsity inherent in high-dimensional space.
    Rather than relying on absolute distance values between points, a relative measure, the neighborhood concept, is utilized.
    This approach facilitates the comparison of two distinct latent spaces with varying dimensionalities.
    Importantly, the neighborhood concept is dimensionality-independent, making this technique applicable and effective across low and high dimensions.

    Mathematically speaking,
    let us consider a collection of sample-label pairs $X$:
    $(x_1, Y_1), (x_2, Y_2), ..., (x_n, Y_n)$
    where $x$ is the input sample, and $Y$ is the label for the input sample.
    Here, the latent space is embedded with such $x$ points.
    Let $S$ be the set of all samples $x_p \in X$ such that they have the same label $c$, i.e.,
    \begin{aequation}
        S_c = \{x_i \mid \forall x_i \in X ~\text{such that}~ Y_{i} = c \}.
    \end{aequation}
    The distance $D$ of sample $x_p$ from all other samples can be formulated as follows:
    \begin{aequation}
        D(x_p) = \{ \text{distance}(x_p, x_i) \mid \forall (x_i, y_i) \in X \},
    \end{aequation}
    where $\text{distance}(a,b)$ is the distance between two samples $a$ and $b$.
    The commonly employed distance function is the Minkowski distance, a generalization of various distance metrics, including Euclidean and City-block distances.
    The Minkowski distance of order $r$ between two points
    $a = (a_1, a_2 \ldots a_d)$ and
    $b = (b_1, b_2 \ldots b_d)$ in $d$ dimensional space is given by,
    \begin{aequation}
        \text{distance}(a, b)_{\text{Minkowski}, r} = \left( \sum_{i=1}^{d} \vert a_i - b_i \vert ^r \right)^{1/r},
    \end{aequation}
    here, it represents City-block distance ($l_1$ norm) when $r=1$;
    it represents Euclidean distance ($l_2$ norm) when $r=2$;
    and
    it represents Maximum Norm distance ($l_\infty$ norm) when $r=\infty$.
    $k^{\text{th}}$ neighbour sample $x_{k}^{p}$ to sample $x_p$ is defined as,
    \begin{aequation}
    x_{k}^{p} = x_q \quad \text{where,} \quad q \in \argmin_{x_q \in P_i}~ \text{distance}(x_q, x_p) \\
    \text{such that} \quad P_i = X - \{ x_{j}^{p} \mid \forall j < i\}.
    \end{aequation}
    Similarly, we can define a sorted local neighborhood space $N_p$ of sample $(x_p)$ based on the distance,
    \begin{aequation}
    N_p = (x_{0}^{p}, x_{1}^{p} \ldots x_{n}^{p})
    \end{aequation}
    such that, $\text{distance}(x_{i}^{p}, x_p) < \text{distance}(x_{j}^{p}, x_p)$, where $i < j$.
    Using this local neighborhood space $N_p$,
    we can define k* of a test sample point $(x_p, Y_p)$ as k$^{\text{th}}$-closest neighbor which differs in label compared to $Y_p$.
    Mathematically, it can be written as,
    \begin{aequation}
    \text{k}^{*}_p = \argmin_{(x_p, Y_p)} \{ x_{i}^{p} \mid x_{i}^{p} \in N_p, ~ Y_{i}^{p} \ne Y_p \},
    \end{aequation}
    where $i$ is the index of the nearest neighbor,
    $Y_p$ is the label of test sample $x_p$ and
    $Y_{i}^{p}$ is the label of the nearest neighbor (sample) $x_{i}^{p}$ that differs compared to label $Y_p$.
    Then, the {k*~distribution} $\text{k}^{*}(\cdot)$ of class $c$ can be defined as,
    \begin{aequation}
    \text{k}^{*}(S_c) = \left\{ \frac{\text{k}^{*}_p}{\vert S_c \vert} \mid \forall x_p \in S_c \right\},
    \end{aequation}
    here $\vert S_c \vert$ is the cardinality of set $S_c$ representing the number of samples belonging to class $c$.

    \begin{figure*}[!t]
        \centering
        \includegraphics[width=0.49\linewidth]{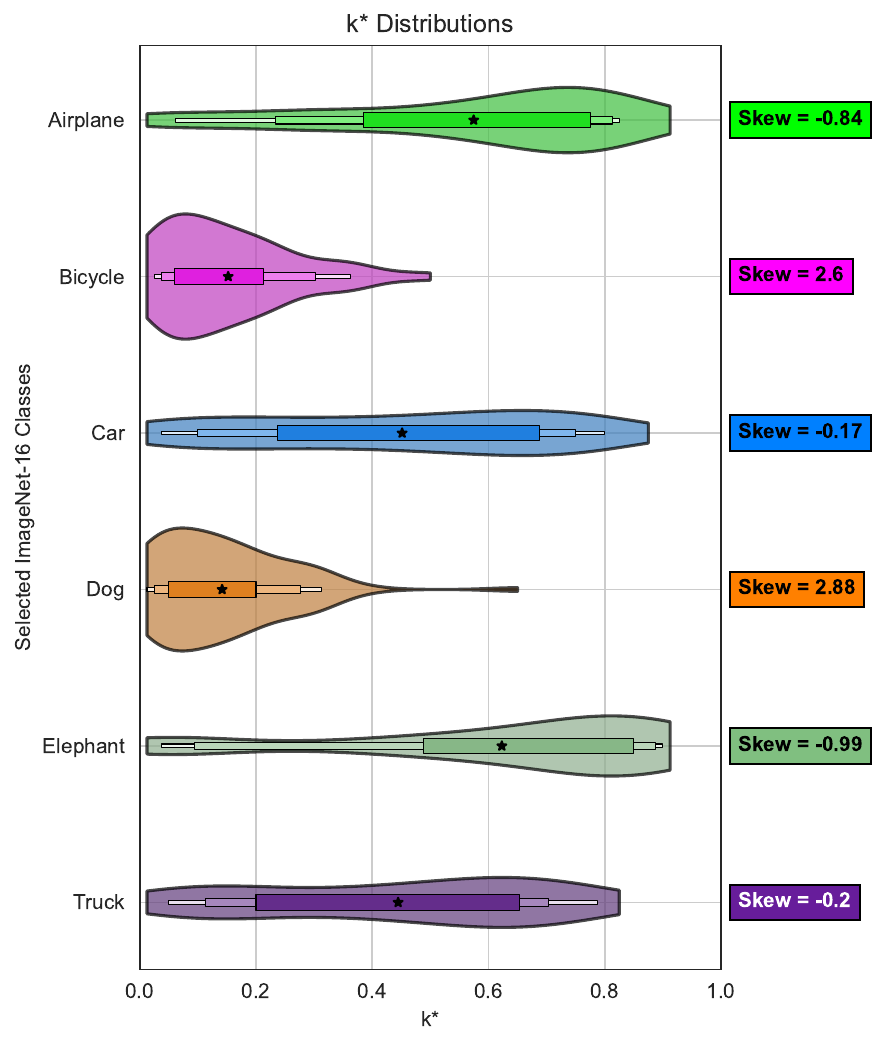}
        \rulesep
        \includegraphics[width=0.49\linewidth]{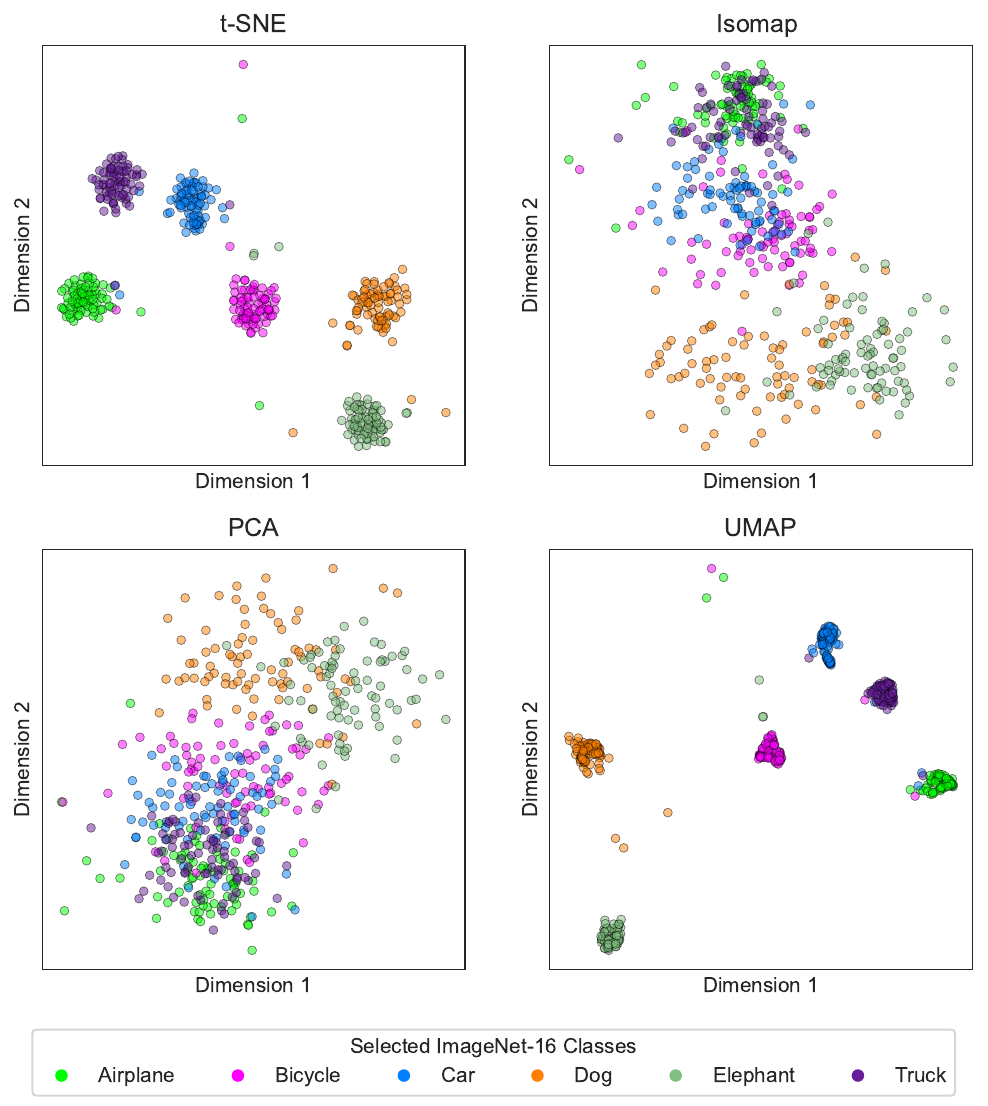}
        \caption{
                Visualization of the distribution of samples in latent space using, 
                \figleft~{k*~distribution}, and 
                \figright~Dimensionality Reduction techniques like 
                	t-SNE \figtopleft, 
                	Isomap \figtopright, 
                	PCA \figbottomleft, and 
                	UMAP \figbottomright~ 
                of all classes of 16-class-ImageNet for the Logit Layer of ResNet-50 Architecture.
                Note that the distribution of samples for a particular class is easier to compare using {k*~distribution} than dimensionality reduction techniques.
        }
        \label{fig:architecture}
    \end{figure*}
    
    Based on the defined {k*~distribution}, we can define certain metrics over it as mentioned below,
    \begin{description}[leftmargin=*]

        \item[Mean of {k*~distribution} ($\mu_{k^*}$): ]
        \begin{aequation}
        \mu_{k^*} = \frac{1}{\vert S \vert} \sum \text{k}^{*}(S)
        \end{aequation}

        \item[Standard Deviation of {k*~distribution} ($\sigma_{k^*}$):]
        \begin{aequation}
        \sigma_{k^*} =  \left( \frac{1}{\vert S \vert} \sum (\text{k}^{*}(S) - \mu_{k^*})^2 \right)^{\frac{1}{2}}
        \end{aequation}

        \item[Skewness Coefficient of {k*~distribution} ($\gamma_{k^*}$):]
        \begin{aequation}
        \gamma_{k^*} = \frac{ \frac{1}{\vert S \vert} \sum (\text{k}^{*}(S) - \mu_{k^*})^3}{ \left(\frac{1}{\vert S \vert} \sum (\text{k}^{*}(S) - \mu_{k^*})^2 \right)^{3/2}}
        \end{aequation}
        It measures the asymmetry of the {k*~distribution} about its mean $\mu_{k^*}$.
        The skewness coefficient can be positive, negative, or zero.
        A positive skewness indicates a distribution that is skewed to the left, i.e., towards lower k* metric values, while a negative skewness indicates a distribution that is skewed to the right, i.e., towards higher k* metric values.

    \end{description}

    Based on the {k*~distribution}s, we observe three distinct patterns (\Figref{fig:illustration}) of sample distribution in the latent space:

    \vspace{0.5em}
    \begin{description}[style=nextline, leftmargin=*, itemsep=0.5em]

        \item[\PatternA~\fractured~distribution of samples:]
        In this latent space configuration, multiple clusters of testing samples are discernible, each separated in the latent space.
        Consequently, most points exhibit low k* metric values, as they belong to smaller clusters.
        Conversely, no points display high k* metric values, given the presence of points from another class distribution situated between the various sub-clusters of the testing class.
        The {k*~distribution} for this clustered distribution of samples in latent space is markedly positively skewed ($\gamma_{k^*} > 0.5$)\footnotemark[1], i.e., skewed towards lower k* metric values, indicating the existence of multiple clusters and the interference of another class distribution amid these clusters.
        As illustrated in \Figref{fig:explanation}, the {k*~distribution} of \mytriangle{green} class follows \PatternA~classifying the distribution of samples in latent space as \fractured.

        \item[\PatternB~\overlapped~distribution of samples:]
        This latent space configuration overlaps samples from two or more classes.
        Consequently, some points possess low k* metric values, suggesting their location in the overlapping region, while others have high k* metric values, signifying their deep embedding within a class cluster.
        Due to the diverse distribution of samples in this latent space, the {k*~distribution} appears nearly uniform ($-0.5 < \gamma_{k^*} < 0.5$)\footnotemark[1].
        As illustrated in \Figref{fig:explanation}, {k*~distribution} of \mycircle{yellow} class follows \PatternB~classifying the distribution of samples in latent space as \overlapped.
        \footnotetext[1]{{From our observation, we note that the most accurate representation of overlapped classes is achieved when the $ \gamma_{k^*} \in [-0.5, 0.5] $.}}

        \item[\PatternC~\clustered~distribution of samples:]
        A homogeneous cluster of testing samples is prevalent in this latent space arrangement.
        As a result, most points boast high k* metric values, indicating their deep placement within the cluster.
        Simultaneously, some points may exhibit low k* metric values as they reside on the cluster's periphery; these edge samples might be closer to points from another class distribution than the majority within the cluster.
        Owing to this concentrated distribution of samples, the {k*~distribution} for this clustered distribution of samples in latent space is strongly negatively biased ($\gamma_{k^*} < -0.5$)\footnotemark[1], i.e., skewed towards higher k* metric values, symbolizing a dense cluster.
        As illustrated in \Figref{fig:explanation}, {k*~distribution} of \mysquare{pink} class follows \PatternC~classifying the distribution of samples in latent space as \clustered.

    \end{description}

    \vspace{0.5em}
    \noindent
    \textbf{Note:} 
    The k* distribution is not a indicator for measuring a neural network’s classification accuracy.
    The k* distribution visualizes the learned latent space from a local neighborhood perspective, while metrics like accuracy evaluate it from a different perspective.
    Both multiple well-separated fractured distributions (indication of overclustering) and well-separated homogeneous clusters (indication of optimal clustering) can lead to high classification accuracy. 

    
    
    
     
    \vspace{0.5em}
    \noindent
    \textbf{Limitations:}
    As our approach revolves around identifying surrounding neighbor samples, we inherit the limitations inherent in the nearest-neighbor method.
    One such trade-off is sacrificing information about the absolute distances between samples and between the distribution of samples.

\section{Experimental Results and Visualization}
    \begin{table*}[!t]
        \centering
        \caption{%
        Multi-category evaluation of various neural architectures by using statistical metrics across object categories.
        }
        \label{tab:architectures}
        \begin{tabular}{l|rrrrr|rrrrr|rrrrr}
        \toprule
        & \multicolumn{5}{c|}{\textbf{Airplane}}
        & \multicolumn{5}{c|}{\textbf{Bicycle}}
        & \multicolumn{5}{c}{\textbf{Car}}
        \\ \textbf{Architectures}
        & $\mu_{k^*}$ & $\sigma_{k^*}$ & $\gamma_{k^*}$ & Acc & Pat
        & $\mu_{k^*}$ & $\sigma_{k^*}$ & $\gamma_{k^*}$ & Acc & Pat
        & $\mu_{k^*}$ & $\sigma_{k^*}$ & $\gamma_{k^*}$ & Acc & Pat
        \\ \midrule

        ResNet-50 \cite{he2016deep}
        & 0.57 & 0.25 & -0.84 & 100.00 & \Clustered
        & 0.15 & 0.11 & 2.60  & 97.50  & \Fractured
        & 0.45 & 0.26 & -0.17 & 93.75  & \Overlapped
        \\ ResNeXt-101 \cite{xie2017aggregated}
        & 0.54 & 0.23 & -0.71 & 100.00 & \Clustered
        & 0.22 & 0.17 & 1.23  & 100.00 & \Fractured
        & 0.50 & 0.28 & -0.32 & 97.50  & \Overlapped
        \\ EfficientNet-B0 \cite{tan2019efficientnet}
        & 0.80 & 0.19 & -3.03 & 100.00 & \Clustered
        & 0.44 & 0.20 & -0.53 & 96.25  & \Clustered
        & 0.54 & 0.20 & -0.93 & 96.25  & \Clustered
        \\ ViT \cite{dosovitskiy2020image}
        & 0.89 & 0.12 & -5.08 & 100.00 & \Clustered
        & 0.41 & 0.15 & -0.20 & 98.75  & \Overlapped
        & 0.52 & 0.20 & -0.44 & 93.75  & \Overlapped

        \\ \bottomrule \toprule
        & \multicolumn{5}{c|}{\textbf{Dog}}
        & \multicolumn{5}{c|}{\textbf{Elephant}}
        & \multicolumn{5}{c}{\textbf{Truck}}
        \\ \textbf{Architecture}
        & $\mu_{k^*}$ & $\sigma_{k^*}$ & $\gamma_{k^*}$ & Acc & Pat
        & $\mu_{k^*}$ & $\sigma_{k^*}$ & $\gamma_{k^*}$ & Acc & Pat
        & $\mu_{k^*}$ & $\sigma_{k^*}$ & $\gamma_{k^*}$ & Acc & Pat
        \\ \midrule

        ResNet-50 \cite{he2016deep}
        & 0.14 & 0.11 & 2.88 & 95.00 & \Fractured
        & 0.62 & 0.29 & -0.99 & 98.75 & \Clustered
        & 0.44 & 0.25 & -0.20 & 98.75 & \Overlapped
        \\ ResNeXt-101 \cite{xie2017aggregated}
        & 0.14 & 0.12 & 2.52 & 97.50 & \Fractured
        & 0.71 & 0.27 & -1.38 & 100.00 & \Clustered
        & 0.49 & 0.24 & -0.45 & 98.75 & \Overlapped
        \\ EfficientNet-B0 \cite{tan2019efficientnet}
        & 0.05 & 0.03 & 7.89 & 100.00 & \Fractured
        & 0.77 & 0.18 & -2.83 & 100.00 & \Clustered
        & 0.48 & 0.20 & -0.86 & 98.75 & \Clustered
        \\ ViT-B \cite{dosovitskiy2020image}
        & 0.05 & 0.03 & 8.06 & 98.75 & \Fractured
        & 0.92 & 0.09 & -5.24 & 100.00 & \Clustered
        & 0.53 & 0.19 & -0.67 & 100.00 & \Clustered

        \\ \bottomrule
        \end{tabular}
    \end{table*}

    \begin{table*}[!t]
        \centering
        \caption{%
        Performance by various neural architectures across varied visual patterns.
        }
        \label{tab:architectures:summary}
        \begin{tabular}{l|rrrr|rrrr|rrrr|rrrH}
        \toprule
        & \multicolumn{4}{c|}{\textbf{\shortstack{Average of Classes with\\\PatternA~(\fractured)}}}
        & \multicolumn{4}{c|}{\textbf{\shortstack{Average of Classes with\\\PatternB~(\overlapped)}}}
        & \multicolumn{4}{c|}{\textbf{\shortstack{Average of Classes with\\\PatternC~(\clustered)}}}
        & \multicolumn{3}{c}{\textbf{\shortstack{Average of 1,000\\ImageNet1k Classes}}}
        \\ \textbf{Architecture}
        & $\mu_{k^*}$ & $\gamma_{k^*}$ & Acc & N
        & $\mu_{k^*}$ & $\gamma_{k^*}$ & Acc & N
        & $\mu_{k^*}$ & $\gamma_{k^*}$ & Acc & N
        & $\mu_{k^*}$ & $\gamma_{k^*}$ & Acc & N
        \\ \midrule

        ResNet-50 \cite{he2016deep}
        & 0.10 & 4.51 & 74.88 & 935
        & 0.37 & 0.14 & 93.96 & 56
        & 0.61 & -1.04 & 97.11 & 9
        & 0.12 & 4.21 & 76.15 & 1,000
        \\ ResNeXt-101 \cite{xie2017aggregated}
        & 0.11 & 4.23 & 75.05 & 865
        & 0.38 & 0.08 & 93.44 & 104
        & 0.60 & -1.12 & 96.06 & 31
        & 0.15 & 3.63 & 77.62 & 1,000
        \\ EfficientNet-B0 \cite{tan2019efficientnet}
        & 0.10 & 4.56 & 74.66 & 840
        & 0.37 & 0.01 & 92.50 & 105
        & 0.59 & -1.40 & 95.45 & 55
        & 0.15 & 3.76 & 77.68 & 1,000
        \\ ViT \cite{dosovitskiy2020image}
        & 0.12 & 3.83 & 74.33 & 604
        & 0.35 & 0.01 & 87.78 & 153
        & 0.64 & -1.85 & 93.59 & 243
        & 0.28 & 1.86 & 81.07 & 1,000


        \\ \bottomrule
        \end{tabular}
    \end{table*}

\subsection{Common Experimental Setup}

    \vspace{0.5em}
    \noindent
    \textbf{Datasets:}
    Initially, we assess the {k*~distribution} using the entire set of $1280$ images from the \SubImageNet~dataset curated by \citet{geirhos2018Generalisationa}.
    This dataset contains $80$ images per each of the $16$ classes of the dataset.
    These are the $16$ entry-level categories from MS-COCO that have the highest number of ImageNet classes mapped via the WordNet hierarchy, making them compatible with the $1,000$ classes of \ImageNet~dataset \cite{russakovsky2015imagenet}.
    The purpose of employing this dataset is to evaluate individual classes organized naturally.
    In addition to the \SubImageNet~dataset, we evaluate {k*~distribution} comprehensively using $50,000$ validation images of the original \ImageNet~dataset \cite{russakovsky2015imagenet}.
    This dataset is divided into $1,000$ classes, each with $50$ images.

    We apply various transformations to the samples from the \SubImageNet~dataset and \ImageNet~dataset to further scrutinize the models' latent spaces.
    These transformations include Image Cropping, adding Gaussian Noise, rotating the images, and generating Adversarial and Stylized Versions of the samples. 
    Note that the pre-trained models are not trained on these transformations.

    Moreover, to assess the {k*~distribution} across various other tasks, we also use various other datasets for different tasks.
    We use the English subset of the MASSIVE \cite{fitzgerald2022massive} dataset for intent classification.
    The testing samples contained $2,970$ samples split across $60$ distinct intent classes.
    For the keyword spotting task, we used the Speech Commands v0.02 dataset \cite{speechcommandsv2}, which contains $4,890$ samples split across $36$ command categories.
    We utilize these datasets to analyze (see \Figref{fig:framework}) and comprehend the characteristics associated with each pattern (see \Figref{fig:illustration}).

    \vspace{0.5em}
    \noindent
    \textbf{Deep Neural Network Architectures:}
    We analyze the latent space of pre-trained weights from various open-source neural networks referenced as we have used them in the experiments.

    \vspace{0.5em}
    \noindent
    \textbf{Adversarial Attacks:}
    We subject the trained ResNet-50 model to a Projected Gradient Descent (PGD) attack, as outlined in \cite{pgd_AEs}, employing a perturbation magnitude of $\eps=4/255$.
    This attack aims to gauge the extent of the effective alteration in representation induced by the adversarial samples.
    Furthermore, we assess the model that has undergone adversarial training using the PGD attack.
    This evaluation compares and contrasts the robust model's latent space with its non-robust counterpart.

    \vspace{0.5em}
    \noindent
    \textbf{Metrics:} 
    For our analyses of individual classes, we report; 
    Mean ($\mu_{k^*}$), 
    Standard deviation ($\sigma_{k^*}$), 
    Skewness Coefficient ($\gamma_{k^*}$), 
    Accuracy (Acc),  
    Number of Classes (N), and 
    prevailing Pattern (Pat) in the context of diverse visual patterns.
    

    \begin{table*}[!t]
        \centering
        \caption{
        Multi-category evaluation of various layers of ResNet-50 by using statistical metrics across object categories.
        }
        \label{tab:layer}
        \begin{tabular}{l|rrrHr|rrrHr|rrrHr}
        \toprule
        & \multicolumn{5}{c|}{\textbf{Airplane}}
        & \multicolumn{5}{c|}{\textbf{Bicycle}}
        & \multicolumn{5}{c}{\textbf{Car}}
        \\ \textbf{Layer Name}
        & $\mu_{k^*}$ & $\sigma_{k^*}$ & $\gamma_{k^*}$ & Acc & Pat
        & $\mu_{k^*}$ & $\sigma_{k^*}$ & $\gamma_{k^*}$ & Acc & Pat
        & $\mu_{k^*}$ & $\sigma_{k^*}$ & $\gamma_{k^*}$ & Acc & Pat
        \\ \midrule

        Logit Layer
        & 0.57 & 0.25 & -0.84 & 100.00 & \Clustered
        & 0.15 & 0.11 & 2.60 & 97.50 & \Fractured
        & 0.45 & 0.26 & -0.17 & 93.75 & \Overlapped
        \\ Average Pooling
        & 0.59 & 0.25 & -0.93 & 0.00 & \Overlapped
        & 0.15 & 0.12 & 2.59 & 0.00 & \Fractured
        & 0.42 & 0.26 & -0.07 & 0.00 & \Overlapped
        \\ Stage4 Block3 Conv3
        & 0.27 & 0.16 & 0.76 & 0.00 & \Overlapped
        & 0.08 & 0.06 & 5.50 & 0.00 & \Fractured
        & 0.06 & 0.06 & 6.12 & 0.00 & \Fractured
        \\ Stage3 Block6 Conv3
        & 0.07 & 0.08 & 4.78 & 0.00 & \Fractured
        & 0.03 & 0.02 & 8.37 & 0.00 & \Fractured
        & 0.03 & 0.03 & 8.14 & 0.00 & \Fractured
        \\ Stage2 Block4 Conv3
        & 0.01 & 0.00 & 8.88 & 0.00 & \Fractured
        & 0.01 & 0.00 & 8.89 & 0.00 & \Fractured
        & 0.01 & 0.00 & 8.89 & 0.00 & \Fractured

        \\ \bottomrule \toprule
        & \multicolumn{5}{c|}{\textbf{Dog}}
        & \multicolumn{5}{c|}{\textbf{Elephant}}
        & \multicolumn{5}{c}{\textbf{Truck}}
        \\ \textbf{Layer Name}
        & $\mu_{k^*}$ & $\sigma_{k^*}$ & $\gamma_{k^*}$ & Acc & Pat
        & $\mu_{k^*}$ & $\sigma_{k^*}$ & $\gamma_{k^*}$ & Acc & Pat
        & $\mu_{k^*}$ & $\sigma_{k^*}$ & $\gamma_{k^*}$ & Acc & Pat
        \\ \midrule

        Logit Layer
        & 0.14 & 0.11 & 2.88 & 95.00 & \Fractured
        & 0.62 & 0.29 & -0.99 & 98.75 & \Clustered
        & 0.44 & 0.25 & -0.20 & 98.75 & \Overlapped
        \\ Average Pooling
        & 0.09 & 0.07 & 4.98 & 0.00 & \Fractured
        & 0.62 & 0.28 & -1.06 & 0.00 & \Clustered
        & 0.43 & 0.25 & -0.19 & 0.00 & \Overlapped
        \\ Stage4 Block3 Conv3
        & 0.01 & 0.00 & 8.87 & 0.00 & \Fractured
        & 0.10 & 0.08 & 4.41 & 0.00 & \Fractured
        & 0.06 & 0.05 & 6.55 & 0.00 & \Fractured
        \\ Stage3 Block6 Conv3
        & 0.02 & 0.01 & 8.82 & 0.00 & \Fractured
        & 0.03 & 0.02 & 8.24 & 0.00 & \Fractured
        & 0.02 & 0.01 & 8.71 & 0.00 & \Fractured
        \\ Stage2 Block4 Conv3
        & 0.01 & 0.00 & 8.89 & 0.00 & \Fractured
        & 0.01 & 0.00 & 8.89 & 0.00 & \Fractured
        & 0.01 & 0.00 & 8.89 & 0.00 & \Fractured
        \\ \bottomrule
        \end{tabular}
    \end{table*}

    \begin{table*}[!t]
        \centering
        \caption{
            Performance by different layers of ResNet-50 across varied visual patterns.
        }
        \label{tab:layer:summary}
        \begin{tabular}{l|rrHr|rrHr|rrHr|rrHH}
            \toprule
            & \multicolumn{4}{c|}{\textbf{\shortstack{Average of\\ Classes with\\\PatternA\\~(\fractured)}}}
            & \multicolumn{4}{c|}{\textbf{\shortstack{Average of\\ Classes with\\\PatternB\\~(\overlapped)}}}
            & \multicolumn{4}{c|}{\textbf{\shortstack{Average of\\ Classes with\\\PatternC\\~(\clustered)}}}
            & \multicolumn{3}{c}{\textbf{\shortstack{Average of\\ 1,000\\ImageNet1k\\ Classes}}}
            \\ \textbf{Layer Name}
            & $\mu_{k^*}$ & $\gamma_{k^*}$ & Acc & N
            & $\mu_{k^*}$ & $\gamma_{k^*}$ & Acc & N
            & $\mu_{k^*}$ & $\gamma_{k^*}$ & Acc & N
            & $\mu_{k^*}$ & $\gamma_{k^*}$ & Acc & N
            \\ \midrule
            
            Logit Layer
            & 0.10 & 4.51 & 74.88 & 935
            & 0.37 & 0.14 & 93.96 & 56
            & 0.61 & -1.04 & 97.11 & 9
            & 0.12 & 4.21 & 76.15 & 1,000
            \\ Average Pooling
            & 0.09 & 4.59 & 0.00 & 837
            & 0.34 & 0.35 & 0.00 & 157
            & 0.72 & -1.73 & 0.00 & 6
            & 0.13 & 3.89 & 0.00 & 1,000
            \\ Stage4 Block3 Conv3
            & 0.05 & 5.90 & 0.00 & 998
            & 0.33 & 0.48 & 0.00 & 2
            & --- & --- & --- & 0
            & 0.05 & 5.89 & 0.00 & 1,000
            \\ Stage3 Block6 Conv3
            & 0.02 & 6.96 & 0.00 & 1,000
            & --- & --- & --- & 0
            & --- & --- & --- & 0
            & 0.02 & 6.96 & 0.00 & 1,000
            \\ Stage2 Block4 Conv3
            & 0.02 & 7.00 & 0.00 & 1,000
            & --- & --- & --- & 0
            & --- & --- & --- & 0
            & 0.02 & 7.00 & 0.00 & 1,000

            \\ \bottomrule
        \end{tabular}
    \end{table*}
    
\subsection{Analysis of Latent Space of Different Neural Architectures}

    We have the option to visualize the learned latent space of ResNet-50 (logit layer) in two ways: 
    
    a) Using {k*~distribution}, as illustrated in \Figref{fig:architecture} \figleft, or, 
    
    b) By employing dimensionality reduction techniques, showcased in \Figref{fig:architecture} \figright.
    
    Intriguingly, the t-SNE \figtopleft~and UMAP \figbottomright~visualizations indicate a highly clustered distribution of samples in latent space for ResNet-50, while Isomap \figtopright~and PCA \figbottomleft~show an overlapping distribution of samples in latent space. This creates uncertainty in distinguishing which classes are well-represented and which are fragmented in the distribution of samples in latent space.

    To address this ambiguity and better understand the local latent space, we turn to the {k*~distribution}, visualized in \Figref{fig:architecture} \figleft.
    By utilizing the {k*~distribution}, we can distinctly identify that the local spaces of the six visualized classes differ.
    For instance, the Airplane and Elephant classes exhibit more \PatternC~clustered distribution of samples in latent space with a negatively skewed {k*~distribution}.

    In contrast, Bicycle and Dog classes showcase \PatternA~fractured distribution of samples in latent space with a positively skewed {k*~distribution}, while Car and Truck have \PatternB~overlapped distribution of samples in latent space with an almost uniform {k*~distribution}.

    Furthermore, we can comprehensively compare various neural architectures, as presented in \Tableref{tab:architectures}.
    This allows us to observe how specific classes are distributed differently across architectures.
    Notably, the Bicycle class displays \PatternA~(fractured distribution of samples) in ResNet and ResNeXt architectures, \PatternB~ (overlapped distribution of samples) in ViT, and \PatternC~(clustered distribution of samples) in EfficientNet-B0.
    Similarly, the Truck class is \PatternB~(overlapped distribution of samples) in ResNet and ResNeXt architectures and \PatternC~(clustered distribution of samples) in ViT and EfficientNet-B0.
    Conversely, Airplane, Dog, and Elephant classes exhibit consistent distribution across all architectures.

    Additionally, a comprehensive comparison of neural architectures can be made by calculating the averages of \fractured, \overlapped, and \clustered~classes, as detailed in \Tableref{tab:architectures:summary}.
    The results indicate distinct distributions of classes across various architectures.
    For instance, ResNet tends to fracture the latent space, with only $9$ clustered classes.
    In contrast, ViT leans towards clustering of samples in the latent space, with a significant number of classes ($243$) exhibiting clustering, i.e., \PatternA.

    A general trend is also observed: 
    an increase in the mean of {k*~distribution} $\mu_{k^*}$ with improvements in model accuracy.
    Additionally, a declining general trend in the skewness coefficient of {k*~distribution} $\gamma_{k^*}$ suggests a transition from fractured to the more overlapped distribution of samples in latent space across tested models.

\subsection{Analysis of Latent Space of Different Layers of a Network}

    \begin{table*}[!t]
        \centering
        \caption{
            Multi-category evaluation of models trained on different 
            distributions by using statistical metrics across object categories.
        }
        \label{tab:training}
        \resizebox{\textwidth}{!}{
            \begin{tabular}{l|rrrrr|rrrrr|rrrrr}
                \toprule
                & \multicolumn{5}{c|}{\textbf{Airplane}}
                & \multicolumn{5}{c|}{\textbf{Bicycle}}
                & \multicolumn{5}{c}{\textbf{Car}}
                \\ \textbf{Training Distribution}
                & $\mu_{k^*}$ & $\sigma_{k^*}$ & $\gamma_{k^*}$ & Acc & Pat
                & $\mu_{k^*}$ & $\sigma_{k^*}$ & $\gamma_{k^*}$ & Acc & Pat
                & $\mu_{k^*}$ & $\sigma_{k^*}$ & $\gamma_{k^*}$ & Acc & Pat
                \\ \midrule
                ImageNet1k \cite{russakovsky2015imagenet}
                & 0.57 & 0.25 & -0.84 & 100.00 & \Clustered
                & 0.15 & 0.11 & 2.60 & 97.50 & \Fractured
                & 0.45 & 0.26 & -0.17 & 93.75 & \Overlapped
                \\ Stylised ImageNet \cite{geirhos2018}
                & 0.41 & 0.24 & -0.29 & 95.00 & \Overlapped
                & 0.13 & 0.09 & 3.72 & 93.75 & \Fractured
                & 0.23 & 0.18 & 1.12 & 87.50 & \Fractured
                \\ ImageNet1k + Stylised \cite{geirhos2018}
                & 0.58 & 0.24 & -0.97 & 100.00 & \Clustered
                & 0.17 & 0.13 & 1.97 & 98.75 & \Fractured
                & 0.43 & 0.27 & -0.02 & 92.50 & \Overlapped
                
                \\ \bottomrule \toprule
                & \multicolumn{5}{c|}{\textbf{Dog}}
                & \multicolumn{5}{c|}{\textbf{Elephant}}
                & \multicolumn{5}{c}{\textbf{Truck}}
                \\ \textbf{Training Distribution}
                & $\mu_{k^*}$ & $\sigma_{k^*}$ & $\gamma_{k^*}$ & Acc & Pat
                & $\mu_{k^*}$ & $\sigma_{k^*}$ & $\gamma_{k^*}$ & Acc & Pat
                & $\mu_{k^*}$ & $\sigma_{k^*}$ & $\gamma_{k^*}$ & Acc & Pat
                \\ \midrule
                ImageNet1k \cite{russakovsky2015imagenet}
                & 0.14 & 0.11 & 2.88 & 95.00 & \Fractured
                & 0.62 & 0.29 & -0.99 & 98.75 & \Clustered
                & 0.44 & 0.25 & -0.20 & 98.75 & \Overlapped
                \\ Stylised ImageNet \cite{geirhos2018}
                & 0.09 & 0.07 & 5.20 & 95.00 & \Fractured
                & 0.41 & 0.28 & 0.01 & 98.75 & \Overlapped
                & 0.26 & 0.21 & 0.75 & 97.50 & \Fractured
                \\ ImageNet1k + Stylised \cite{geirhos2018}
                & 0.16 & 0.13 & 2.22 & 96.25 & \Fractured
                & 0.64 & 0.29 & -0.85 & 100.00 & \Clustered
                & 0.40 & 0.22 & -0.07 & 100.00 & \Overlapped
                \\ \bottomrule
        \end{tabular}}
    \end{table*}
    
    \begin{table*}[!t]
        \centering
        \caption{%
            Performance by models trained on different training datasets across varied visual patterns.
        }
        \label{tab:training:summary}
        \resizebox{\textwidth}{!}{
            \begin{tabular}{l|rrrr|rrrr|rrrr|rrrH}
                \toprule
                & \multicolumn{4}{c|}{\textbf{\shortstack{Average of Classes with\\\PatternA~(\fractured)}}}
                & \multicolumn{4}{c|}{\textbf{\shortstack{Average of Classes with\\\PatternB~(\overlapped)}}}
                & \multicolumn{4}{c|}{\textbf{\shortstack{Average of Classes with\\\PatternC~(\clustered)}}}
                & \multicolumn{3}{c}{\textbf{\shortstack{Average of 1,000\\ImageNet1k Classes}}}
                \\ \textbf{Training Distribution}
                & $\mu_{k^*}$ & $\gamma_{k^*}$ & Acc & N
                & $\mu_{k^*}$ & $\gamma_{k^*}$ & Acc & N
                & $\mu_{k^*}$ & $\gamma_{k^*}$ & Acc & N
                & $\mu_{k^*}$ & $\gamma_{k^*}$ & Acc & N
                \\ \midrule
                ImageNet1k \cite{russakovsky2015imagenet}
                & 0.10 & 4.51 & 74.88 & 935
                & 0.37 & 0.14 & 93.96 & 56
                & 0.61 & -1.04 & 97.11 & 9
                & 0.12 & 4.21 & 76.15 & 1,000
                \\ Stylised ImageNet \cite{geirhos2018}
                & 0.05 & 5.84 & 59.99 & 994
                & 0.35 & 0.19 & 91.33 & 6
                & --- & --- & --- & 0
                & 0.06 & 5.80 & 60.18 & 1,000
                \\ ImageNet1k + Stylised \cite{geirhos2018}
                & 0.09 & 4.67 & 73.36 & 939
                & 0.37 & 0.12 & 92.90 & 51
                & 0.57 & -0.98 & 96.20 & 10
                & 0.11 & 4.38 & 74.59 & 1,000
                \\ \bottomrule
        \end{tabular}}
    \end{table*}
    
    \begin{table*}[!t]
        \centering
        \caption{
            Multi-category evaluation of Robust (Adversarially Trained: AT) and Non-Robust (Standard Trained: ST) models by using statistical metrics across object categories.
        }
        \label{tab:adversarial}
        \begin{tabular}{lr|rrrHr|rrrHr|rrrHr}
            \toprule
            && \multicolumn{5}{c|}{\textbf{Airplane}}
            & \multicolumn{5}{c|}{\textbf{Bicycle}}
            & \multicolumn{5}{c}{\textbf{Car}}
            \\ \textbf{Architecture} & \textbf{Type}
            & $\mu_{k^*}$ & $\sigma_{k^*}$ & $\gamma_{k^*}$ & Acc & Pat
            & $\mu_{k^*}$ & $\sigma_{k^*}$ & $\gamma_{k^*}$ & Acc & Pat
            & $\mu_{k^*}$ & $\sigma_{k^*}$ & $\gamma_{k^*}$ & Acc & Pat
            \\ \midrule
            
            \multirow{2}{*}{ResNet-50 \cite{he2016deep}} & AT \cite{dong2020Benchmarking}
            & 0.40 & 0.22 & -0.40 & 100.00 & \Overlapped
            & 0.13 & 0.09 & 3.56 & 97.50 & \Fractured
            & 0.35 & 0.23 & 0.18 & 92.50 & \Overlapped
            \\ & ST \cite{he2016deep}
            & 0.57 & 0.25 & -0.84 & 100.00 & \Clustered
            & 0.15 & 0.11 & 2.60 & 97.50 & \Fractured
            & 0.45 & 0.26 & -0.17 & 93.75 & \Overlapped
            \\ \midrule \multirow{2}{*}{WideResNet-50 \cite{zagoruyko2017Widea}} & AT \cite{dong2020Benchmarking}
            & 0.64 & 0.26 & -1.51 & 100.00 & \Clustered
            & 0.16 & 0.10 & 2.76 & 96.25 & \Fractured
            & 0.57 & 0.27 & -0.81 & 92.50 & \Clustered
            \\ & ST \cite{zagoruyko2017Widea}
            & 0.87 & 0.17 & -3.75 & 100.00 & \Clustered
            & 0.21 & 0.14 & 1.30 & 98.75 & \Fractured
            & 0.57 & 0.20 & -0.57 & 97.50 & \Clustered
            \\ \midrule \multirow{2}{*}{ViT-B \cite{dosovitskiy2020image}} & AT \cite{dong2020Benchmarking}
            & 0.76 & 0.17 & -3.26 & 100.00 & \Clustered
            & 0.25 & 0.16 & 0.84 & 98.75 & \Fractured
            & 0.62 & 0.23 & -1.26 & 98.75 & \Clustered
            \\ & ST \cite{dosovitskiy2020image}
            & 0.91 & 0.19 & -3.27 & 100.00 & \Clustered
            & 0.30 & 0.17 & 0.25 & 96.25 & \Overlapped
            & 0.56 & 0.26 & -0.75 & 97.50 & \Clustered
            
            \\ \bottomrule \toprule
            && \multicolumn{5}{c|}{\textbf{Dog}}
            & \multicolumn{5}{c|}{\textbf{Elephant}}
            & \multicolumn{5}{c}{\textbf{Truck}}
            \\ \textbf{Architecture} & \textbf{Type}
            & $\mu_{k^*}$ & $\sigma_{k^*}$ & $\gamma_{k^*}$ & Acc & Pat
            & $\mu_{k^*}$ & $\sigma_{k^*}$ & $\gamma_{k^*}$ & Acc & Pat
            & $\mu_{k^*}$ & $\sigma_{k^*}$ & $\gamma_{k^*}$ & Acc & Pat
            \\ \midrule
            
            \multirow{2}{*}{ResNet-50 \cite{he2016deep}} & AT \cite{dong2020Benchmarking}
            & 0.06 & 0.05 & 6.34 & 63.75 & \Fractured
            & 0.33 & 0.23 & 0.10 & 96.25 & \Overlapped
            & 0.24 & 0.17 & 0.71 & 93.75 & \Fractured
            \\ & ST \cite{he2016deep}
            & 0.14 & 0.11 & 2.87 & 80.00 & \Fractured
            & 0.62 & 0.29 & -0.99 & 100.00 & \Clustered
            & 0.44 & 0.25 & -0.20 & 93.75 & \Overlapped
            \\ \midrule \multirow{2}{*}{WideResNet-50 \cite{zagoruyko2017Widea}} & AT \cite{dong2020Benchmarking}
            & 0.09 & 0.08 & 4.66 & 77.50 & \Fractured
            & 0.50 & 0.25 & -0.94 & 98.75 & \Clustered
            & 0.50 & 0.25 & -0.85 & 95.00 & \Clustered
            \\ & ST \cite{zagoruyko2017Widea}
            & 0.06 & 0.05 & 6.86 & 36.25 & \Fractured
            & 0.83 & 0.15 & -3.92 & 100.00 & \Clustered
            & 0.52 & 0.18 & -0.73 & 97.50 & \Clustered
            \\ \midrule \multirow{2}{*}{ViT-B \cite{dosovitskiy2020image}} & AT \cite{dong2020Benchmarking}
            & 0.06 & 0.04 & 6.96 & 73.75 & \Fractured
            & 0.56 & 0.21 & -1.60 & 100.00 & \Clustered
            & 0.47 & 0.22 & -0.52 & 98.75 & \Clustered
            \\ & ST \cite{dosovitskiy2020image}
            & 0.17 & 0.12 & 2.28 & 95.00 & \Fractured
            & 0.78 & 0.25 & -1.95 & 100.00 & \Clustered
            & 0.64 & 0.22 & -1.11 & 98.75 & \Clustered
            
            \\ \bottomrule
        \end{tabular}
    \end{table*}

    \begin{table*}[!t]
        \centering
        \caption{
            Performance of Robust (Adversarially Trained: AT) and Non-Robust (Standard Trained: ST) models across varied visual patterns. 
        }
        \label{tab:adversarial:summary}
        \resizebox{\textwidth}{!}{
            \begin{tabular}{lr|rrrr|rrrr|rrrr|rrrH}
                \toprule
                && \multicolumn{4}{c|}{\textbf{\shortstack{Average of Classes with\\\PatternA~(\fractured)}}}
                & \multicolumn{4}{c|}{\textbf{\shortstack{Average of Classes with\\\PatternB~(\overlapped)}}}
                & \multicolumn{4}{c|}{\textbf{\shortstack{Average of Classes with\\\PatternC~(\clustered)}}}
                & \multicolumn{3}{c}{\textbf{\shortstack{Average of 1,000\\ImageNet1k Classes}}}
                \\ \textbf{Architecture} & \textbf{Type}
                & $\mu_{k^*}$ & $\gamma_{k^*}$ & Acc & N
                & $\mu_{k^*}$ & $\gamma_{k^*}$ & Acc & N
                & $\mu_{k^*}$ & $\gamma_{k^*}$ & Acc & N
                & $\mu_{k^*}$ & $\gamma_{k^*}$ & Acc & N
                \\ \midrule
                
                \multirow{2}{*}{ResNet-50 \cite{he2016deep}} & AT \cite{dong2020Benchmarking}
                & 0.06 & 5.77 & 64.52 & 975
                & 0.35 & 0.04 & 94.60 & 20
                & 0.57 & -1.22 & 95.60 & 5
                & 0.07 & 5.62 & 65.28 & 1,000
                \\ & ST \cite{he2016deep}
                & 0.10 & 4.51 & 74.88 & 935
                & 0.37 & 0.14 & 93.96 & 56
                & 0.61 & -1.04 & 97.11 & 9
                & 0.12 & 4.22 & 76.15 & 1,000
                \\ \midrule \multirow{2}{*}{WideResNet-50 \cite{zagoruyko2017Widea}} & AT \cite{dong2020Benchmarking}
                & 0.08 & 5.12 & 66.53 & 895
                & 0.35 & -0.01 & 89.91 & 66
                & 0.60 & -1.30 & 95.90 & 39
                & 0.11 & 4.53 & 69.22 & 1,000
                \\ & ST \cite{zagoruyko2017Widea}
                & 0.12 & 3.85 & 75.53 & 650
                & 0.35 & -0.02 & 88.59 & 135
                & 0.64 & -1.70 & 94.67 & 215
                & 0.26 & 2.13 & 81.41 & 1,000
                \\ \midrule \multirow{2}{*}{ViT-B \cite{dosovitskiy2020image}} & AT \cite{dong2020Benchmarking}
                & 0.08 & 5.01 & 68.87 & 823
                & 0.34 & -0.04 & 88.29 & 83
                & 0.59 & -1.49 & 94.45 & 94
                & 0.15 & 3.98 & 72.88 & 1,000
                \\ & ST \cite{dosovitskiy2020image}
                & 0.12 & 3.77 & 68.12 & 652
                & 0.35 & 0.05 & 84.97 & 149
                & 0.67 & -1.59 & 92.73 & 199
                & 0.26 & 2.15 & 75.53 & 1,000

                \\ \bottomrule
        \end{tabular}}
    \end{table*}
    
    We analyze the latent space of various layers within ResNet-50 to gain insights into how classes are represented across different layers of the same model.
    Specifically, we examine the latent space of the final logit layer, the average pooling layer, and several convolution layers after different stages in the ResNet-50 architecture, as depicted in \Twotablesref{tab:layer}{tab:layer:summary}.

    Upon observation, we note that that number of overlapped classes from the average pooling layer decreases from $153$ to $56$ in the final logit layer, suggesting that logit layer fractures the overlapped regions. This is evident from the results as the number of fractured classes increases from $837$ in average pooling layer to $935$ for the logit layer. 

\subsection{Analysis of Latent Space of Different Training Distributions}

    It is well-established that the learned latent space of a neural network undergoes changes based on the distribution of the training data. 
    In order to assess these changes in the learned latent space with respect to training data distribution, we conduct an evaluation using ResNet-50 trained on different data distributions.

    Specifically, we compare the models trained on standard ImageNet1k samples, those trained on the standard ImageNet1k dataset \cite{russakovsky2015imagenet}, a stylized version of ImageNet-1k (Stylized ImageNet) \cite{geirhos2018}, and a combination of both, as illustrated in \Twotablesref{tab:training}{tab:training:summary}.
    Through this comparison, we can discern alterations in the representation of different classes in the latent space.
    Notably, training exclusively with Stylized ImageNet results in a more fractured distribution of samples in latent space for the non-stylized images.

    Furthermore, we observe a consistent trend where models exhibit better accuracy with higher $\mu_{k^*}$ and lower  $\gamma_{k^*}$ mirroring the pattern observed in \Tableref{tab:architectures:summary}. This suggests a correlation between improved model performance and specific characteristics of the {k*~distribution}s.

\subsection{Analysis of Latent Space of Adversarially Robust Models}

    \begin{table*}[!t]
        \centering
        \caption{
        Multi-category evaluation on samples transformed with image crop by using statistical metrics across object categories.
        }
        \label{tab:crop}
        \begin{tabular}{r|rrrrr|rrrrr|rrrrr}
        \toprule
        & \multicolumn{5}{c|}{\textbf{Airplane}}
        & \multicolumn{5}{c|}{\textbf{Bicycle}}
        & \multicolumn{5}{c}{\textbf{Car}}
        \\ \multirow{-2}{*}{\textbf{\shortstack[l]{Image Size ($s$)\\After Crop}}}
        & $\mu_{k^*}$ & $\sigma_{k^*}$ & $\gamma_{k^*}$ & Acc & Pat
        & $\mu_{k^*}$ & $\sigma_{k^*}$ & $\gamma_{k^*}$ & Acc & Pat
        & $\mu_{k^*}$ & $\sigma_{k^*}$ & $\gamma_{k^*}$ & Acc & Pat
        \\ \midrule

        20
        & 0.02 & 0.01 & 8.70 & 5.00 & \Fractured
        & 0.01 & 0.01 & 8.85 & 0.00 & \Fractured
        & 0.01 & 0.01 & 8.83 & 0.00 & \Fractured
        \\ 60
        & 0.06 & 0.06 & 6.05 & 80.00 & \Fractured
        & 0.02 & 0.02 & 8.56 & 48.75 & \Fractured
        & 0.04 & 0.05 & 7.03 & 36.25 & \Fractured
        \\ 100
        & 0.15 & 0.11 & 2.54 & 95.00 & \Fractured
        & 0.05 & 0.06 & 6.29 & 77.50 & \Fractured
        & 0.13 & 0.14 & 2.14 & 72.50 & \Fractured
        \\ 140
        & 0.36 & 0.22 & -0.05 & 97.50 & \Overlapped
        & 0.10 & 0.09 & 3.93 & 92.50 & \Fractured
        & 0.28 & 0.23 & 0.59 & 83.75 & \Fractured
        \\ 180
        & 0.47 & 0.24 & -0.48 & 98.75 & \Overlapped
        & 0.12 & 0.09 & 3.82 & 96.25 & \Fractured
        & 0.40 & 0.26 & 0.05 & 95.00 & \Overlapped
        \\ 220
        & 0.62 & 0.27 & -0.98 & 100.00 & \Clustered
        & 0.16 & 0.12 & 2.50 & 96.25 & \Fractured
        & 0.47 & 0.26 & -0.25 & 93.75 & \Overlapped

        \\ \bottomrule \toprule
        & \multicolumn{5}{c|}{\textbf{Dog}}
        & \multicolumn{5}{c|}{\textbf{Elephant}}
        & \multicolumn{5}{c}{\textbf{Truck}}
        \\ \multirow{-2}{*}{\textbf{\shortstack[l]{Image Size ($s$)\\After Crop}}}
        & $\mu_{k^*}$ & $\sigma_{k^*}$ & $\gamma_{k^*}$ & Acc & Pat
        & $\mu_{k^*}$ & $\sigma_{k^*}$ & $\gamma_{k^*}$ & Acc & Pat
        & $\mu_{k^*}$ & $\sigma_{k^*}$ & $\gamma_{k^*}$ & Acc & Pat
        \\ \midrule

        20
        & 0.02 & 0.01 & 8.85 & 18.75 & \Fractured
        & 0.02 & 0.01 & 8.78 & 16.25 & \Fractured
        & 0.01 & 0.01 & 8.85 & 7.50 & \Fractured
        \\ 60
        & 0.02 & 0.01 & 8.70 & 66.25 & \Fractured
        & 0.05 & 0.05 & 6.43 & 81.25 & \Fractured
        & 0.03 & 0.02 & 8.39 & 47.50 & \Fractured
        \\ 100
        & 0.04 & 0.03 & 7.81 & 82.50 & \Fractured
        & 0.20 & 0.16 & 1.23 & 88.75 & \Fractured
        & 0.08 & 0.09 & 3.87 & 78.75 & \Fractured
        \\ 140
        & 0.08 & 0.07 & 5.20 & 90.00 & \Fractured
        & 0.43 & 0.27 & -0.16 & 97.50 & \Overlapped
        & 0.18 & 0.18 & 1.40 & 92.50 & \Fractured
        \\ 180
        & 0.13 & 0.10 & 2.99 & 93.75 & \Fractured
        & 0.57 & 0.28 & -0.56 & 98.75 & \Clustered
        & 0.39 & 0.23 & -0.09 & 97.50 & \Overlapped
        \\ 220
        & 0.14 & 0.11 & 2.88 & 96.25 & \Fractured
        & 0.61 & 0.29 & -0.91 & 98.75 & \Clustered
        & 0.45 & 0.24 & -0.21 & 98.75 & \Overlapped

        \\ \bottomrule
        \end{tabular}
    \end{table*}

    \begin{table*}[!t]
        \centering
        \caption{
        Performance on samples transformed with image crop across varied visual patterns.
        }
        \label{tab:crop:summary}
        \begin{tabular}{r|rrrr|rrrr|rrrr|rrrH}
        \toprule
        & \multicolumn{4}{c|}{\textbf{\shortstack{Average of Classes with\\\PatternA~(\fractured)}}}
        & \multicolumn{4}{c|}{\textbf{\shortstack{Average of Classes with\\\PatternB~(\overlapped)}}}
        & \multicolumn{4}{c|}{\textbf{\shortstack{Average of Classes with\\\PatternC~(\clustered)}}}
        & \multicolumn{3}{c}{\textbf{\shortstack{Average of 1,000\\ImageNet1k Classes}}}
        \\ \multirow{-2}{*}{\textbf{\shortstack[l]{Image Size ($s$)\\After Crop}}}
        & $\mu_{k^*}$ & $\gamma_{k^*}$ & Acc & N
        & $\mu_{k^*}$ & $\gamma_{k^*}$ & Acc & N
        & $\mu_{k^*}$ & $\gamma_{k^*}$ & Acc & N
        & $\mu_{k^*}$ & $\gamma_{k^*}$ & Acc & N
        \\ \midrule

        20
        & 0.02 & 6.99 & 1.28 & 1,000
        & --- & --- & --- & 0
        & --- & --- & --- & 0
        & 0.02 & 6.99 & 1.28 & 1,000
        \\ 60
        & 0.02 & 6.91 & 15.84 & 1,000
        & --- & --- & --- & 0
        & --- & --- & --- & 0
        & 0.02 & 6.91 & 15.84 & 1,000
        \\ 100
        & 0.03 & 6.58 & 38.64 & 1,000
        & --- & --- & --- & 0
        & --- & --- & --- & 0
        & 0.03 & 6.58 & 38.64 & 1,000
        \\ 140
        & 0.05 & 6.05 & 54.24 & 997
        & 0.38 & 0.09 & 94.00 & 3
        & --- & --- & --- & 0
        & 0.05 & 6.03 & 54.36 & 1,000
        \\ 180
        & 0.06 & 5.53 & 62.29 & 992
        & 0.34 & 0.12 & 89.43 & 7
        & 0.60 & -0.85 & 98.00 & 1
        & 0.06 & 5.48 & 62.52 & 1,000
        \\ 220
        & 0.07 & 5.13 & 67.37 & 985
        & 0.36 & 0.11 & 93.45 & 11
        & 0.55 & -0.76 & 94.50 & 4
        & 0.08 & 5.05 & 67.77 & 1,000

        \\ \bottomrule
        \end{tabular}
    \end{table*}

    \begin{table*}[!t]
        \centering
        \caption{
            Multi-category evaluation on samples transformed with image rotation by using statistical metrics across object categories.
        }
        \label{tab:rotation}
        \begin{tabular}{r|rrrrr|rrrrr|rrrrr}
            \toprule
            & \multicolumn{5}{c|}{\textbf{Airplane}}
            & \multicolumn{5}{c|}{\textbf{Bicycle}}
            & \multicolumn{5}{c}{\textbf{Car}}
            \\ \multirow{-2}{*}{\textbf{\shortstack{Rotation angle \\($r^\degree$) (\CircleArrowright)}}}
            & $\mu_{k^*}$ & $\sigma_{k^*}$ & $\gamma_{k^*}$ & Acc & Pat
            & $\mu_{k^*}$ & $\sigma_{k^*}$ & $\gamma_{k^*}$ & Acc & Pat
            & $\mu_{k^*}$ & $\sigma_{k^*}$ & $\gamma_{k^*}$ & Acc & Pat
            \\ \midrule
            
            30
            & 0.10 & 0.10 & 3.54 & 48.75 & \Fractured
            & 0.06 & 0.06 & 5.74 & 71.25 & \Fractured
            & 0.10 & 0.11 & 3.34 & 63.75 & \Fractured
            \\ 60
            & 0.06 & 0.05 & 6.39 & 27.50 & \Fractured
            & 0.04 & 0.04 & 7.65 & 42.50 & \Fractured
            & 0.05 & 0.05 & 6.55 & 13.75 & \Fractured
            \\ 90
            & 0.10 & 0.10 & 3.60 & 50.00 & \Fractured
            & 0.08 & 0.08 & 4.53 & 85.00 & \Fractured
            & 0.07 & 0.06 & 5.64 & 28.75 & \Fractured
            \\ 120
            & 0.05 & 0.05 & 6.74 & 7.50 & \Fractured
            & 0.03 & 0.02 & 8.43 & 21.25 & \Fractured
            & 0.04 & 0.04 & 7.23 & 11.25 & \Fractured
            \\ 150
            & 0.05 & 0.04 & 7.23 & 12.50 & \Fractured
            & 0.02 & 0.02 & 8.55 & 20.00 & \Fractured
            & 0.04 & 0.05 & 6.88 & 15.00 & \Fractured
            \\ 180
            & 0.14 & 0.12 & 2.40 & 90.00 & \Fractured
            & 0.08 & 0.07 & 5.07 & 86.25 & \Fractured
            & 0.08 & 0.07 & 5.23 & 68.75 & \Fractured
            \\ 210
            & 0.05 & 0.04 & 7.41 & 11.25 & \Fractured
            & 0.03 & 0.02 & 8.38 & 21.25 & \Fractured
            & 0.04 & 0.06 & 6.32 & 17.50 & \Fractured
            \\ 240
            & 0.04 & 0.03 & 8.02 & 13.75 & \Fractured
            & 0.03 & 0.03 & 7.93 & 33.75 & \Fractured
            & 0.04 & 0.04 & 7.35 & 8.75 & \Fractured
            \\ 270
            & 0.11 & 0.08 & 4.29 & 51.25 & \Fractured
            & 0.07 & 0.07 & 5.50 & 85.00 & \Fractured
            & 0.07 & 0.07 & 5.18 & 26.25 & \Fractured
            \\ 300
            & 0.05 & 0.04 & 6.96 & 15.00 & \Fractured
            & 0.03 & 0.02 & 8.24 & 42.50 & \Fractured
            & 0.06 & 0.07 & 5.37 & 22.50 & \Fractured
            \\ 330
            & 0.09 & 0.08 & 4.37 & 45.00 & \Fractured
            & 0.08 & 0.07 & 5.00 & 63.75 & \Fractured
            & 0.12 & 0.11 & 2.85 & 58.75 & \Fractured

            \\ \bottomrule \toprule
            & \multicolumn{5}{c|}{\textbf{Dog}}
            & \multicolumn{5}{c|}{\textbf{Elephant}}
            & \multicolumn{5}{c}{\textbf{Truck}}
            \\ \multirow{-2}{*}{\textbf{\shortstack{Rotation angle \\($r^\degree$) (\CircleArrowright)}}}
            & $\mu_{k^*}$ & $\sigma_{k^*}$ & $\gamma_{k^*}$ & Acc & Pat
            & $\mu_{k^*}$ & $\sigma_{k^*}$ & $\gamma_{k^*}$ & Acc & Pat
            & $\mu_{k^*}$ & $\sigma_{k^*}$ & $\gamma_{k^*}$ & Acc & Pat
            \\ \midrule
            
            30
            & 0.09 & 0.09 & 4.30 & 65.00 & \Fractured
            & 0.14 & 0.14 & 2.03 & 46.25 & \Fractured
            & 0.09 & 0.09 & 3.86 & 67.50 & \Fractured
            \\ 60
            & 0.04 & 0.03 & 7.64 & 48.75 & \Fractured
            & 0.06 & 0.07 & 5.69 & 25.00 & \Fractured
            & 0.04 & 0.03 & 7.98 & 16.25 & \Fractured
            \\ 90
            & 0.05 & 0.05 & 6.39 & 77.50 & \Fractured
            & 0.12 & 0.12 & 2.88 & 65.00 & \Fractured
            & 0.05 & 0.04 & 7.14 & 57.50 & \Fractured
            \\ 120
            & 0.03 & 0.03 & 7.81 & 22.50 & \Fractured
            & 0.03 & 0.03 & 8.02 & 3.75 & \Fractured
            & 0.03 & 0.02 & 8.23 & 8.75 & \Fractured
            \\ 150
            & 0.03 & 0.03 & 8.17 & 25.00 & \Fractured
            & 0.04 & 0.03 & 7.82 & 3.75 & \Fractured
            & 0.04 & 0.03 & 8.07 & 7.50 & \Fractured
            \\ 180
            & 0.04 & 0.04 & 7.36 & 70.00 & \Fractured
            & 0.13 & 0.12 & 2.60 & 68.75 & \Fractured
            & 0.05 & 0.05 & 6.83 & 60.00 & \Fractured
            \\ 210
            & 0.03 & 0.04 & 7.65 & 21.25 & \Fractured
            & 0.04 & 0.04 & 7.38 & 3.75 & \Fractured
            & 0.04 & 0.03 & 7.81 & 8.75 & \Fractured
            \\ 240
            & 0.03 & 0.04 & 7.55 & 23.75 & \Fractured
            & 0.05 & 0.06 & 6.25 & 3.75 & \Fractured
            & 0.04 & 0.04 & 7.54 & 7.50 & \Fractured
            \\ 270
            & 0.04 & 0.04 & 7.48 & 77.50 & \Fractured
            & 0.12 & 0.12 & 2.54 & 61.25 & \Fractured
            & 0.05 & 0.04 & 7.46 & 56.25 & \Fractured
            \\ 300
            & 0.05 & 0.06 & 5.93 & 45.00 & \Fractured
            & 0.06 & 0.06 & 6.03 & 17.50 & \Fractured
            & 0.04 & 0.03 & 8.14 & 33.75 & \Fractured
            \\ 330
            & 0.08 & 0.07 & 5.08 & 68.75 & \Fractured
            & 0.14 & 0.14 & 2.07 & 47.50 & \Fractured
            & 0.07 & 0.08 & 4.72 & 57.50 & \Fractured
            
            \\ \bottomrule
        \end{tabular}
    \end{table*}

    Having explored the changes in representation space with variations in training distribution, it is essential to address models' susceptibility to adversarial attacks.
    To enhance robustness against such attacks, adversarially trained models have been proposed, incorporating adversarial samples in the training distribution \cite{pgd_AEs}.
    In order to evaluate the shifts in the learned latent space between adversarially trained models and their non-robust counterparts, we examine the latent space of different robust models alongside their non-robust counterparts, as presented in \Twotablesref{tab:adversarial}{tab:adversarial:summary}.

    Observations from the table indicate that adversarially trained models tend to exhibit a more fractured distribution of samples in latent space than their non-robust counterparts.
    This explains the current trade-off between accuracy and robustness in the image classification models studied by \citet{tsipras2019Robustness}.
    This suggests that, in an effort to achieve clustering of heterogeneous feature samples within a single cluster as supervised with the pre-defined class label, models often compromise on learning robust features.
    Additionally, we observe a consistent trend where accuracy is associated with higher $\mu_{k^*}$ and lower $\gamma_{k^*}$, explaining why adversarially trained models may demonstrate reduced performance on clean samples compared to their non-robust counterparts.

\subsection{Analysis of Latent Space of Different Input Transformations}
	
    \begin{table*}[!t]
    	\centering
    	\caption{
    		Performance on samples transformed with image rotation across varied visual patterns. 
    	}
    	\label{tab:rotation:summary}
    	\begin{tabular}{r|rrrr|rrrr|rrrr|rrrH}
    		\toprule
    		& \multicolumn{4}{c|}{\textbf{\shortstack{Average of Classes with\\\PatternA~(\fractured)}}}
    		& \multicolumn{4}{c|}{\textbf{\shortstack{Average of Classes with\\\PatternB~(\overlapped)}}}
    		& \multicolumn{4}{c|}{\textbf{\shortstack{Average of Classes with\\\PatternC~(\clustered)}}}
    		& \multicolumn{3}{c}{\textbf{\shortstack{Average of 1,000\\ImageNet1k Classes}}}
    		\\ \multirow{-2}{*}{\textbf{\shortstack{Rotation angle\\ ($r^\degree$) (\CircleArrowright)}}}
    		& $\mu_{k^*}$ & $\gamma_{k^*}$ & Acc & N
    		& $\mu_{k^*}$ & $\gamma_{k^*}$ & Acc & N
    		& $\mu_{k^*}$ & $\gamma_{k^*}$ & Acc & N
    		& $\mu_{k^*}$ & $\gamma_{k^*}$ & Acc & N
    		\\ \midrule
    		
    		30
    		& 0.06 & 5.51 & 53.26 & 983
    		& 0.35 & 0.22 & 86.53 & 15
    		& 0.58 & -0.96 & 92.00 & 2
    		& 0.07 & 5.42 & 53.84 & 1,000
    		\\ 60
    		& 0.05 & 6.05 & 40.57 & 993
    		& 0.38 & 0.05 & 88.33 & 6
    		& 0.50 & -0.93 & 86.00 & 1
    		& 0.05 & 6.00 & 40.90 & 1,000
    		\\ 90
    		& 0.05 & 5.94 & 50.46 & 988
    		& 0.36 & 0.24 & 92.73 & 11
    		& 0.54 & -1.15 & 90.00 & 1
    		& 0.06 & 5.87 & 50.97 & 1,000
    		\\ 120
    		& 0.04 & 6.38 & 28.87 & 999
    		& 0.42 & -0.41 & 88.00 & 1
    		& --- & --- & --- & 0
    		& 0.04 & 6.37 & 28.93 & 1,000
    		\\ 150
    		& 0.04 & 6.39 & 29.07 & 997
    		& 0.33 & 0.30 & 85.00 & 2
    		& 0.48 & -0.63 & 90.00 & 1
    		& 0.04 & 6.37 & 29.24 & 1,000
    		\\ 180
    		& 0.05 & 5.95 & 51.19 & 991
    		& 0.37 & 0.18 & 92.25 & 8
    		& 0.53 & -0.77 & 90.00 & 1
    		& 0.05 & 5.90 & 51.55 & 1,000
    		\\ 210
    		& 0.04 & 6.38 & 28.61 & 999
    		& 0.43 & -0.46 & 88.00 & 1
    		& --- & --- & --- & 0
    		& 0.04 & 6.38 & 28.67 & 1,000
    		\\ 240
    		& 0.04 & 6.37 & 29.24 & 999
    		& 0.43 & -0.40 & 88.00 & 1
    		& --- & --- & --- & 0
    		& 0.04 & 6.36 & 29.30 & 1,000
    		\\ 270
    		& 0.05 & 5.95 & 49.92 & 989
    		& 0.35 & 0.22 & 90.80 & 10
    		& 0.51 & -0.92 & 92.00 & 1
    		& 0.06 & 5.88 & 50.37 & 1,000
    		\\ 300
    		& 0.05 & 6.05 & 39.58 & 994
    		& 0.35 & 0.28 & 88.40 & 5
    		& 0.49 & -0.70 & 88.00 & 1
    		& 0.05 & 6.01 & 39.87 & 1,000
    		\\ 330
    		& 0.06 & 5.51 & 53.39 & 978
    		& 0.36 & 0.14 & 87.68 & 19
    		& 0.58 & -0.81 & 96.00 & 3
    		& 0.07 & 5.39 & 54.17 & 1,000
    		
    		\\ \bottomrule
    	\end{tabular}
    \end{table*}
    
    \begin{table*}[!t]
    	\centering
    	\caption{
    		Multi-category evaluation on samples transformed with Gaussian noise by using statistical metrics across object categories.
    	}
    	\label{tab:noise}
    	\begin{tabular}{r|rrrrr|rrrrr|rrrrr}
    		\toprule
    		& \multicolumn{5}{c|}{\textbf{Airplane}}
    		& \multicolumn{5}{c|}{\textbf{Bicycle}}
    		& \multicolumn{5}{c}{\textbf{Car}}
    		\\ \multirow{-2}{*}{\textbf{\shortstack{Strength ($\alpha$) of\\Gaussian Noise}}}
    		& $\mu_{k^*}$ & $\sigma_{k^*}$ & $\gamma_{k^*}$ & Acc & Pat
    		& $\mu_{k^*}$ & $\sigma_{k^*}$ & $\gamma_{k^*}$ & Acc & Pat
    		& $\mu_{k^*}$ & $\sigma_{k^*}$ & $\gamma_{k^*}$ & Acc & Pat
    		\\ \midrule
    		
    		0.05
    		& 0.39 & 0.23 & -0.10 & 100.00 & \Overlapped
    		& 0.14 & 0.11 & 3.01 & 93.75 & \Fractured
    		& 0.38 & 0.25 & 0.07 & 87.50 & \Overlapped
    		\\ 0.06
    		& 0.32 & 0.20 & 0.20 & 100.00 & \Overlapped
    		& 0.13 & 0.10 & 3.21 & 93.75 & \Fractured
    		& 0.36 & 0.24 & 0.21 & 87.50 & \Overlapped
    		\\ 0.08
    		& 0.34 & 0.22 & 0.08 & 97.50 & \Overlapped
    		& 0.12 & 0.09 & 3.57 & 96.25 & \Fractured
    		& 0.34 & 0.24 & 0.36 & 83.75 & \Overlapped
    		\\ 0.12
    		& 0.20 & 0.16 & 1.46 & 95.00 & \Fractured
    		& 0.10 & 0.07 & 4.92 & 91.25 & \Fractured
    		& 0.25 & 0.21 & 0.86 & 85.00 & \Fractured
    		\\ 0.18
    		& 0.13 & 0.12 & 2.60 & 87.50 & \Fractured
    		& 0.12 & 0.12 & 2.77 & 83.75 & \Fractured
    		& 0.24 & 0.20 & 0.77 & 77.50 & \Fractured
    		\\ 0.26
    		& 0.07 & 0.07 & 5.32 & 63.75 & \Fractured
    		& 0.07 & 0.09 & 4.11 & 60.00 & \Fractured
    		& 0.13 & 0.12 & 2.55 & 61.25 & \Fractured
    		\\ 0.38
    		& 0.03 & 0.04 & 7.65 & 15.00 & \Fractured
    		& 0.02 & 0.02 & 8.56 & 20.00 & \Fractured
    		& 0.07 & 0.08 & 4.67 & 27.50 & \Fractured
    		\\ 0.70
    		& 0.02 & 0.02 & 8.59 & 0.00 & \Fractured
    		& 0.01 & 0.00 & 8.86 & 1.25 & \Fractured
    		& 0.02 & 0.01 & 8.67 & 2.50 & \Fractured
    		\\ 1.00
    		& 0.02 & 0.01 & 8.77 & 0.00 & \Fractured
    		& 0.02 & 0.00 & 8.86 & 0.00 & \Fractured
    		& 0.02 & 0.01 & 8.84 & 0.00 & \Fractured

    		\\ \bottomrule \toprule
    		& \multicolumn{5}{c|}{\textbf{Dog}}
    		& \multicolumn{5}{c|}{\textbf{Elephant}}
    		& \multicolumn{5}{c}{\textbf{Truck}}
    		\\ \multirow{-2}{*}{\textbf{\shortstack{Strength ($\alpha$) of\\Gaussian Noise}}}
    		& $\mu_{k^*}$ & $\sigma_{k^*}$ & $\gamma_{k^*}$ & Acc & Pat
    		& $\mu_{k^*}$ & $\sigma_{k^*}$ & $\gamma_{k^*}$ & Acc & Pat
    		& $\mu_{k^*}$ & $\sigma_{k^*}$ & $\gamma_{k^*}$ & Acc & Pat
    		\\ \midrule
    		
    		0.05
    		& 0.09 & 0.08 & 4.51 & 91.25 & \Fractured
    		& 0.51 & 0.29 & -0.37 & 97.50 & \Overlapped
    		& 0.32 & 0.22 & 0.35 & 96.25 & \Overlapped
    		\\ 0.06
    		& 0.08 & 0.08 & 4.77 & 87.50 & \Fractured
    		& 0.42 & 0.28 & 0.03 & 97.50 & \Overlapped
    		& 0.32 & 0.22 & 0.28 & 97.50 & \Overlapped
    		\\ 0.08
    		& 0.06 & 0.05 & 6.25 & 88.75 & \Fractured
    		& 0.39 & 0.27 & 0.05 & 96.25 & \Overlapped
    		& 0.31 & 0.23 & 0.47 & 97.50 & \Overlapped
    		\\ 0.12
    		& 0.04 & 0.04 & 7.41 & 81.25 & \Fractured
    		& 0.22 & 0.20 & 0.97 & 93.75 & \Fractured
    		& 0.22 & 0.19 & 1.01 & 97.50 & \Fractured
    		\\ 0.18
    		& 0.03 & 0.03 & 8.19 & 68.75 & \Fractured
    		& 0.15 & 0.15 & 2.03 & 92.50 & \Fractured
    		& 0.11 & 0.11 & 2.96 & 93.75 & \Fractured
    		\\ 0.26
    		& 0.02 & 0.02 & 8.64 & 42.50 & \Fractured
    		& 0.06 & 0.08 & 5.07 & 77.50 & \Fractured
    		& 0.07 & 0.08 & 4.71 & 67.50 & \Fractured
    		\\ 0.38
    		& 0.02 & 0.01 & 8.80 & 20.00 & \Fractured
    		& 0.02 & 0.02 & 8.43 & 48.75 & \Fractured
    		& 0.02 & 0.02 & 8.57 & 21.25 & \Fractured
    		\\ 0.70
    		& 0.02 & 0.01 & 8.83 & 0.00 & \Fractured
    		& 0.01 & 0.01 & 8.86 & 1.25 & \Fractured
    		& 0.02 & 0.01 & 8.72 & 0.00 & \Fractured
    		\\ 1.00
    		& 0.01 & 0.01 & 8.86 & 0.00 & \Fractured
    		& 0.01 & 0.00 & 8.86 & 0.00 & \Fractured
    		& 0.02 & 0.01 & 8.75 & 0.00 & \Fractured
    		
    		\\ \bottomrule
    	\end{tabular}
    \end{table*}
    
    \begin{table*}[!t]
    	\centering
    	\caption{
    		Performance on samples transformed with Gaussian noise across varied visual patterns.
    	}
    	\label{tab:noise:summary}
    	\begin{tabular}{r|rrrr|rrrr|rrrr|rrrH}
    		\toprule
    		& \multicolumn{4}{c|}{\textbf{\shortstack{Average of Classes with\\\PatternA~(\fractured)}}}
    		& \multicolumn{4}{c|}{\textbf{\shortstack{Average of Classes with\\\PatternB~(\overlapped)}}}
    		& \multicolumn{4}{c|}{\textbf{\shortstack{Average of Classes with\\\PatternC~(\clustered)}}}
    		& \multicolumn{3}{c}{\textbf{\shortstack{Average of 1,000\\ImageNet1k Classes}}}
    		\\ \multirow{-2}{*}{\textbf{\shortstack{Strength ($\alpha$) of \\Gaussian Noise}}}
    		& $\mu_{k^*}$ & $\gamma_{k^*}$ & Acc & N
    		& $\mu_{k^*}$ & $\gamma_{k^*}$ & Acc & N
    		& $\mu_{k^*}$ & $\gamma_{k^*}$ & Acc & N
    		& $\mu_{k^*}$ & $\gamma_{k^*}$ & Acc & N
    		\\ \midrule
    		
    		0.05
    		& 0.09 & 4.75 & 72.26 & 956
    		& 0.35 & 0.20 & 93.06 & 36
    		& 0.58 & -0.83 & 96.25 & 8
    		& 0.10 & 4.55 & 73.20 & 1,000
    		\\ 0.06
    		& 0.09 & 4.84 & 71.57 & 961
    		& 0.35 & 0.18 & 92.73 & 33
    		& 0.58 & -0.83 & 96.33 & 6
    		& 0.10 & 4.66 & 72.41 & 1,000
    		\\ 0.08
    		& 0.08 & 4.96 & 69.81 & 972
    		& 0.37 & 0.12 & 93.00 & 24
    		& 0.59 & -0.84 & 95.50 & 4
    		& 0.09 & 4.82 & 70.47 & 1,000
    		\\ 0.12
    		& 0.07 & 5.27 & 65.31 & 984
    		& 0.38 & 0.06 & 93.43 & 14
    		& 0.60 & -0.96 & 93.00 & 2
    		& 0.08 & 5.19 & 65.76 & 1,000
    		\\ 0.18
    		& 0.05 & 5.78 & 56.24 & 991
    		& 0.35 & 0.20 & 88.25 & 8
    		& 0.49 & -0.74 & 88.00 & 1
    		& 0.06 & 5.73 & 56.53 & 1,000
    		\\ 0.26
    		& 0.04 & 6.30 & 42.35 & 998
    		& 0.42 & -0.11 & 89.00 & 2
    		& --- & --- & --- & 0
    		& 0.04 & 6.28 & 42.44 & 1,000
    		\\ 0.38
    		& 0.03 & 6.73 & 24.49 & 1,000
    		& --- & --- & --- & 0
    		& --- & --- & --- & 0
    		& 0.03 & 6.73 & 24.49 & 1,000
    		\\ 0.70
    		& 0.02 & 6.96 & 4.78 & 1,000
    		& --- & --- & --- & 0
    		& --- & --- & --- & 0
    		& 0.02 & 6.96 & 4.78 & 1,000
    		\\ 1.00
    		& 0.02 & 6.98 & 1.45 & 1,000
    		& --- & --- & --- & 0
    		& --- & --- & --- & 0
    		& 0.02 & 6.98 & 1.45 & 1,000
    		
    		\\ \bottomrule
    	\end{tabular}
    \end{table*}

    \begin{table*}[!t]
        \centering
        \caption{
            Multi-category evaluation of samples transformed with image stylization on models trained with different 
            distributions by using statistical metrics across object categories.
        }
        \label{tab:stylised}
        \resizebox{\textwidth}{!}{
            \begin{tabular}{l|rrrrr|rrrrr|rrrrr}
                \toprule
                & \multicolumn{5}{c|}{\textbf{Airplane}}
                & \multicolumn{5}{c|}{\textbf{Bicycle}}
                & \multicolumn{5}{c}{\textbf{Car}}
                \\ \textbf{Training Distribution}
                & $\mu_{k^*}$ & $\sigma_{k^*}$ & $\gamma_{k^*}$ & Acc & Pat
                & $\mu_{k^*}$ & $\sigma_{k^*}$ & $\gamma_{k^*}$ & Acc & Pat
                & $\mu_{k^*}$ & $\sigma_{k^*}$ & $\gamma_{k^*}$ & Acc & Pat
                \\ \midrule
                
                ImageNet1k \cite{russakovsky2015imagenet}
                & 0.02 & 0.01 & 6.94 & 18.00 & \Fractured
                & 0.03 & 0.02 & 6.84 & 46.00 & \Fractured
                & 0.04 & 0.03 & 6.42 & 30.00 & \Fractured
                \\ Stylised ImageNet \cite{geirhos2018}
                & 0.55 & 0.27 & -0.40 & 94.00 & \Overlapped
                & 0.33 & 0.20 & 0.40 & 94.00 & \Overlapped
                & 0.18 & 0.15 & 1.85 & 88.00 & \Fractured
                \\ ImageNet1k \cite{russakovsky2015imagenet} + Stylised \cite{geirhos2018}
                & 0.51 & 0.26 & -0.33 & 94.00 & \Overlapped
                & 0.35 & 0.22 & 0.32 & 96.00 & \Overlapped
                & 0.15 & 0.12 & 2.54 & 96.00 & \Fractured
                
                \\ \bottomrule \toprule
                & \multicolumn{5}{c|}{\textbf{Dog}}
                & \multicolumn{5}{c|}{\textbf{Elephant}}
                & \multicolumn{5}{c}{\textbf{Truck}}
                \\ \textbf{Training Distribution}
                & $\mu_{k^*}$ & $\sigma_{k^*}$ & $\gamma_{k^*}$ & Acc & Pat
                & $\mu_{k^*}$ & $\sigma_{k^*}$ & $\gamma_{k^*}$ & Acc & Pat
                & $\mu_{k^*}$ & $\sigma_{k^*}$ & $\gamma_{k^*}$ & Acc & Pat
                \\ \midrule
                
                ImageNet1k \cite{russakovsky2015imagenet}
                & 0.02 & 0.01 & 6.95 & 34.00 & \Fractured
                & 0.02 & 0.01 & 6.98 & 46.00 & \Fractured
                & 0.02 & 0.01 & 6.91 & 32.00 & \Fractured
                \\ Stylised \cite{geirhos2018}
                & 0.16 & 0.11 & 2.85 & 94.00 & \Fractured
                & 0.44 & 0.24 & 0.21 & 98.00 & \Overlapped
                & 0.09 & 0.09 & 4.16 & 84.00 & \Fractured
                \\ ImageNet1k \cite{russakovsky2015imagenet} + Stylised \cite{geirhos2018}
                & 0.14 & 0.09 & 3.69 & 96.00 & \Fractured
                & 0.44 & 0.21 & -0.26 & 96.00 & \Overlapped
                & 0.11 & 0.09 & 3.68 & 82.00 & \Fractured
                
                \\ \bottomrule
        \end{tabular}}
    \end{table*}

    \begin{table*}[!t]
        \centering
        \caption{
            Performance by samples transformed with image stylization on models trained with different distributions across varied visual patterns. 
        }
        \label{tab:stylised:summary}
        \resizebox{\textwidth}{!}{
            \begin{tabular}{l|rrrr|rrrr|rrrr|rrrH}
                \toprule
                & \multicolumn{4}{c|}{\textbf{\shortstack{Average of Classes with\\\PatternA~(\fractured)}}}
                & \multicolumn{4}{c|}{\textbf{\shortstack{Average of Classes with\\\PatternB~(\overlapped)}}}
                & \multicolumn{4}{c|}{\textbf{\shortstack{Average of Classes with\\\PatternC~(\clustered)}}}
                & \multicolumn{3}{c}{\textbf{\shortstack{Average of 1,000\\ImageNet1k Classes}}}
                \\ \textbf{Training Distribution}
                & $\mu_{k^*}$ & $\gamma_{k^*}$ & Acc & N
                & $\mu_{k^*}$ & $\gamma_{k^*}$ & Acc & N
                & $\mu_{k^*}$ & $\gamma_{k^*}$ & Acc & N
                & $\mu_{k^*}$ & $\gamma_{k^*}$ & Acc & N
                \\ \midrule
                
                ImageNet1k \cite{russakovsky2015imagenet}
                & 0.02 & 6.91 & 20.18 & 1,000
                & --- & --- & --- & 0
                & --- & --- & --- & 0
                & 0.02 & 6.91 & 20.18 & 1,000
                \\ Stylised \cite{geirhos2018}
                & 0.04 & 6.25 & 54.04 & 999
                & 0.29 & 0.45 & 86.00 & 1
                & --- & --- & --- & 0
                & 0.04 & 6.24 & 54.07 & 1,000
                \\ ImageNet1k \cite{russakovsky2015imagenet} + Stylised \cite{geirhos2018}
                & 0.04 & 6.41 & 47.97 & 1,000
                & --- & --- & --- & 0
                & --- & --- & --- & 0
                & 0.04 & 6.41 & 47.97 & 1,000
                
                \\ \bottomrule
        \end{tabular}}
    \end{table*}
    
    The learned latent space of neural networks is known to be highly susceptible to slight perturbations in input samples, making it brittle.
    We analyze the latent space of ResNet-50 using samples modified with various input transformations to comprehend the distribution of such transformed samples.
    Specifically, we apply the five transformations described below to the input sample and individually evaluate their impact.
    Each sample is transformed once using the transformation to estimate their distribution in the latent space.

    \begin{description}[style=sameline, leftmargin=*, itemsep=0.5em]

    \item[Image Crop:]
    The input sample is center-cropped to a size $s$ in this transformation. 
    The reduced image is then resized to the standard $256\times 256$ dimensions and processed by the neural network.
    The results of this transformation are presented in \Twotablesref{tab:crop}{tab:crop:summary}.
    Notably, as more pixels are cropped from the samples, the distribution of transformed samples becomes more fractured in the latent space.
    This suggests that the model relies on detecting multiple features within the image, and missing information can lead the model to perceive the sample as entirely different from the original.

    \item[Image Rotation:]
    In this transformation, the input sample is rotated counter-clockwise (\CircleArrowright) by an angle $r^\degree$, which is processed by the neural network.
    Results of this transformation are presented in \Twotablesref{tab:rotation}{tab:rotation:summary}.
    Our examination of the {k*~distribution} for the rotated samples reveals that transformed samples are highly fractured in the latent space implying that rotated samples are interpreted differently from non-rotated samples, i.e. samples with $0^\degree$ rotation that has less fracturing (see \Twotablesref{tab:architectures}{tab:architectures:summary}).
    Further these fracturing is also not separated by the neural network as we observe a significant degradation in performance as measured in Accuracy (Acc).

    \item[Gaussian Noise]:
    In this transformation, Gaussian noise is added to the input sample.
    Mathematically, if $x$ is the input sample, $z \sim \gN(\mu, \alpha^2)$ is sampled Gaussian noise, and $\alpha$ is the strength of Gaussian Noise, transformed image $x'$ can be written as, $x' = x + (\alpha \times z)$,
    which is then processed by the neural network, and the results of this transformation are presented in \Twotablesref{tab:noise}{tab:noise:summary}.

    Similar to the other two input transformations, this also induces fracturing of the distribution of transformed samples in the latent space.
    Additionally, it is noteworthy that as the strength of the Gaussian noise gradually increases, more classes become fractured, suggesting a gradual breakdown in the features identified by the model.
    
    \item[Image Stylization:]
    In this transformation, the input sample is deprived of its original texture and replaced with a random painting style \cite{geirhos2018}.
    The results of this transformation on models trained on the standard ImageNet-1k \cite{russakovsky2015imagenet} dataset, a stylized version Stylized ImageNet \cite{geirhos2018}, and a combination of both are displayed in \Twotablesref{tab:stylised}{tab:stylised:summary}.
    Similar to other transformations, stylization in the images induces the fracturing of transformed samples.
    This further underscores the model's sensitivity to variations in input samples, leading to distinct representations for stylized samples.

    \begin{table*}[!t]
        \centering
        \caption{
            Multi-category evaluation of adversarial samples created with PGD Attack by using statistical metrics across object categories.
        }
        \label{tab:attack}
        \begin{tabular}{ll|rrrHr|rrrHr|rrrHr}
            \toprule
            && \multicolumn{5}{c|}{\textbf{Airplane}}
            & \multicolumn{5}{c|}{\textbf{Bicycle}}
            & \multicolumn{5}{c}{\textbf{Car}}
            \\ \textbf{Architecture} & \textvareps
            & $\mu_{k^*}$ & $\sigma_{k^*}$ & $\gamma_{k^*}$ & Acc & Pat
            & $\mu_{k^*}$ & $\sigma_{k^*}$ & $\gamma_{k^*}$ & Acc & Pat
            & $\mu_{k^*}$ & $\sigma_{k^*}$ & $\gamma_{k^*}$ & Acc & Pat
            \\ \midrule
            
            \multirow{3}{*}{\shortstack[l]{ResNet-50\\Adversarially\\Trained \cite{dong2020Benchmarking} }} & 2/255
            & 0.33 & 0.22 & -0.08 & 100.00 & \Overlapped
            & 0.10 & 0.08 & 4.29 & 96.25 & \Fractured
            & 0.26 & 0.19 & 0.74 & 86.25 & \Fractured
            \\  & 4/255
            & 0.31 & 0.22 & 0.03 & 97.50 & \Overlapped
            & 0.09 & 0.08 & 4.22 & 85.00 & \Fractured
            & 0.17 & 0.13 & 2.00 & 73.75 & \Fractured
            \\  & 8/255
            & 0.26 & 0.20 & 0.34 & 83.75 & \Overlapped
            & 0.06 & 0.06 & 5.65 & 55.00 & \Fractured
            & 0.10 & 0.07 & 4.97 & 55.00 & \Fractured
            
            \\ \midrule \multirow{3}{*}{\shortstack[l]{ResNet-50\\Standard\\Trained \cite{he2016deep}}} & 2/255
            & 0.04 & 0.03 & 7.69 & 0.00 & \Fractured
            & 0.02 & 0.02 & 8.61 & 0.00 & \Fractured
            & 0.18 & 0.15 & 1.22 & 0.00 & \Fractured
            \\ & 4/255
            & 0.06 & 0.06 & 5.83 & 0.00 & \Fractured
            & 0.02 & 0.02 & 8.60 & 0.00 & \Fractured
            & 0.29 & 0.22 & 0.10 & 0.00 & \Overlapped
            \\ & 8/255
            & 0.05 & 0.04 & 6.94 & 0.00 & \Fractured
            & 0.03 & 0.02 & 8.37 & 0.00 & \Fractured
            & 0.36 & 0.27 & -0.30 & 0.00 & \Overlapped
            
            \\ \bottomrule \toprule
            && \multicolumn{5}{c|}{\textbf{Dog}}
            & \multicolumn{5}{c|}{\textbf{Elephant}}
            & \multicolumn{5}{c}{\textbf{Truck}}
            \\ \textbf{Architecture} & \textvareps
            & $\mu_{k^*}$ & $\sigma_{k^*}$ & $\gamma_{k^*}$ & Acc & Pat
            & $\mu_{k^*}$ & $\sigma_{k^*}$ & $\gamma_{k^*}$ & Acc & Pat
            & $\mu_{k^*}$ & $\sigma_{k^*}$ & $\gamma_{k^*}$ & Acc & Pat
            \\ \midrule
            
            \multirow{3}{*}{\shortstack[l]{ResNet-50\\Adversarially\\Trained \cite{dong2020Benchmarking} }} & 2/255
            & 0.04 & 0.04 & 7.43 & 33.75 & \Fractured
            & 0.19 & 0.17 & 1.23 & 90.00 & \Fractured
            & 0.17 & 0.14 & 1.56 & 85.00 & \Fractured
            \\  & 4/255
            & 0.03 & 0.03 & 7.96 & 12.50 & \Fractured
            & 0.10 & 0.10 & 3.29 & 76.25 & \Fractured
            & 0.12 & 0.12 & 2.51 & 68.75 & \Fractured
            \\  & 8/255
            & 0.02 & 0.02 & 8.57 & 2.50 & \Fractured
            & 0.06 & 0.05 & 6.26 & 35.00 & \Fractured
            & 0.07 & 0.08 & 4.80 & 41.25 & \Fractured
            \\ \midrule \multirow{3}{*}{\shortstack[l]{ResNet-50\\Standard\\Trained \cite{he2016deep}}} & 2/255
            & 0.02 & 0.01 & 8.79 & 0.00 & \Fractured
            & 0.04 & 0.04 & 7.38 & 0.00 & \Fractured
            & 0.03 & 0.02 & 8.35 & 0.00 & \Fractured
            \\ & 4/255
            & 0.02 & 0.02 & 8.60 & 0.00 & \Fractured
            & 0.05 & 0.05 & 6.37 & 0.00 & \Fractured
            & 0.04 & 0.04 & 7.54 & 0.00 & \Fractured
            \\ & 8/255
            & 0.04 & 0.04 & 7.42 & 0.00 & \Fractured
            & 0.05 & 0.05 & 6.61 & 0.00 & \Fractured
            & 0.04 & 0.04 & 7.53 & 0.00 & \Fractured
            
            \\ \bottomrule
        \end{tabular}
    \end{table*}
    
    \begin{table*}[!t]
        \centering
        \caption{
            Performance on adversarial samples created with PGD Attack across varied visual patterns.
        }
        \label{tab:attack:summary}
        \begin{tabular}{ll|rrrr|rrrr|rrrr|rrrH}
            \toprule
            && \multicolumn{4}{c|}{\textbf{\shortstack{Average of Classes with\\\PatternA~(\fractured)}}}
            & \multicolumn{4}{c|}{\textbf{\shortstack{Average of Classes with\\\PatternB~(\overlapped)}}}
            & \multicolumn{4}{c|}{\textbf{\shortstack{Average of Classes with\\\PatternC~(\clustered)}}}
            & \multicolumn{3}{c}{\textbf{\shortstack{Average of 1,000\\ImageNet1k Classes}}}
            \\ \textbf{Architecture} & \textvareps
            & $\mu_{k^*}$ & $\gamma_{k^*}$ & Acc & N
            & $\mu_{k^*}$ & $\gamma_{k^*}$ & Acc & N
            & $\mu_{k^*}$ & $\gamma_{k^*}$ & Acc & N
            & $\mu_{k^*}$ & $\gamma_{k^*}$ & Acc & N
            \\ \midrule
            
            \multirow{3}{*}{\shortstack[l]{ResNet-50\\Adversarially\\Trained \cite{dong2020Benchmarking} }} & 2/255
            & 0.05 & 6.14 & 50.61 & 983
            & 0.34 & 0.05 & 90.43 & 14
            & 0.57 & -1.18 & 91.33 & 3
            & 0.05 & 6.03 & 51.29 & 1,000
            \\  & 4/255
            & 0.04 & 6.38 & 36.18 & 990
            & 0.32 & 0.14 & 85.50 & 8
            & 0.57 & -1.17 & 88.00 & 2
            & 0.04 & 6.31 & 36.68 & 1,000
            \\  & 8/255
            & 0.03 & 6.64 & 16.45 & 996
            & 0.30 & 0.24 & 72.67 & 3
            & 0.50 & -1.00 & 90.00 & 1
            & 0.03 & 6.62 & 16.69 & 1,000
            \\ \midrule \multirow{3}{*}{\shortstack[l]{ResNet-50\\Standard\\Trained \cite{he2016deep}}} & 2/255
            & 0.05 & 5.82 & 0.00 & 980
            & 0.32 & 0.01 & 0.00 & 15
            & 0.50 & -0.80 & 0.00 & 5
            & 0.06 & 5.70 & 0.00 & 1,000
            \\ & 4/255
            & 0.05 & 5.93 & 0.00 & 986
            & 0.33 & -0.01 & 0.00 & 11
            & 0.50 & -0.70 & 0.00 & 3
            & 0.05 & 5.85 & 0.00 & 1,000
            \\ & 8/255
            & 0.04 & 6.38 & 0.03 & 996
            & 0.31 & 0.25 & 0.00 & 4
            & --- & --- & --- & 0
            & 0.04 & 6.36 & 0.03 & 1,000

            \\ \bottomrule
        \end{tabular}
    \end{table*}

    \begin{table*}[!t]
        \centering
        \caption{
        Multi-category evaluation using different distance metrics by using statistical metrics across object categories.
        }
        \label{tab:distance}
        \begin{tabular}{l|rrrHr|rrrHr|rrrHr}
        \toprule
        & \multicolumn{5}{c|}{\textbf{Airplane}}
        & \multicolumn{5}{c|}{\textbf{Bicycle}}
        & \multicolumn{5}{c}{\textbf{Car}}
        \\ \textbf{Distance Metric}
        & $\mu_{k^*}$ & $\sigma_{k^*}$ & $\gamma_{k^*}$ & Acc & Pat
        & $\mu_{k^*}$ & $\sigma_{k^*}$ & $\gamma_{k^*}$ & Acc & Pat
        & $\mu_{k^*}$ & $\sigma_{k^*}$ & $\gamma_{k^*}$ & Acc & Pat
        \\ \midrule

        Euclidean ($l_2$ norm)
        & 0.57 & 0.25 & -0.84 & 100.00 & \Clustered
        & 0.15 & 0.11 & 2.60 & 97.50 & \Fractured
        & 0.45 & 0.26 & -0.17 & 93.75 & \Overlapped
        \\ CityBlock ($l_1$ norm)
        & 0.55 & 0.25 & -0.71 & 100.00 & \Clustered
        & 0.14 & 0.10 & 2.99 & 97.50 & \Fractured
        & 0.41 & 0.24 & -0.05 & 93.75 & \Overlapped
        \\ Max Norm ($l_\infty$ norm)
        & 0.69 & 0.31 & -1.02 & 100.00 & \Clustered
        & 0.31 & 0.28 & 0.74 & 97.50 & \Overlapped
        & 0.42 & 0.26 & -0.26 & 93.75 & \Overlapped
        \\ Cosine
        & 0.65 & 0.29 & -0.96 & 100.00 & \Clustered
        & 0.23 & 0.16 & 1.29 & 97.50 & \Fractured
        & 0.45 & 0.28 & -0.05 & 93.75 & \Overlapped

        \\ \bottomrule \toprule
        & \multicolumn{5}{c|}{\textbf{Dog}}
        & \multicolumn{5}{c|}{\textbf{Elephant}}
        & \multicolumn{5}{c}{\textbf{Truck}}
        \\ \textbf{Distance Metric}
        & $\mu_{k^*}$ & $\sigma_{k^*}$ & $\gamma_{k^*}$ & Acc & Pat
        & $\mu_{k^*}$ & $\sigma_{k^*}$ & $\gamma_{k^*}$ & Acc & Pat
        & $\mu_{k^*}$ & $\sigma_{k^*}$ & $\gamma_{k^*}$ & Acc & Pat
        \\ \midrule

        Euclidean ($l_2$ norm)
        & 0.14 & 0.11 & 2.88 & 95.00 & \Fractured
        & 0.62 & 0.29 & -0.99 & 98.75 & \Clustered
        & 0.44 & 0.25 & -0.20 & 98.75 & \Overlapped
        \\ CityBlock ($l_1$ norm)
        & 0.13 & 0.10 & 3.12 & 95.00 & \Fractured
        & 0.56 & 0.27 & -0.79 & 98.75 & \Overlapped
        & 0.39 & 0.23 & 0.05 & 98.75 & \Overlapped
        \\ Max Norm ($l_\infty$ norm)
        & 0.10 & 0.07 & 4.74 & 95.00 & \Fractured
        & 0.63 & 0.28 & -1.09 & 98.75 & \Clustered
        & 0.53 & 0.30 & -0.37 & 98.75 & \Overlapped
        \\ Cosine
        & 0.17 & 0.12 & 2.11 & 95.00 & \Fractured
        & 0.57 & 0.29 & -0.46 & 98.75 & \Clustered
        & 0.52 & 0.28 & -0.43 & 98.75 & \Overlapped

        \\ \bottomrule
        \end{tabular}
    \end{table*}

    \item[Adversarial Perturbation:]
    To measure the distribution of adversarial samples, we perturb the input sample with adversarial perturbations optimized by a PGD attack \cite{pgd_AEs} with varying strength $\epsilon$ of adversarial perturbation.
    For a given input sample $x$ and a neural network $f$ such that $f(x)$ is the label of sample $x$ predicted by the neural network, adversarial perturbation $\delta$ can be defined as, $f(x) \neq f(x+\delta)$, where,
    the adversarial perturbation $\delta$ can be further optimized as,
    
    \begin{aequation}
        & \underset{\delta}{\text{minimize}} & & f(x+\delta)
        & \text{subject to} & & \Vert \delta \Vert_p \leq \epsilon \\
    \end{aequation}

    The results of adversarial samples on robust and non-robust models are presented in \Twotablesref{tab:attack}{tab:attack:summary}.
    Like other transformations, adversarial perturbations cause also fragmentation in the distribution of samples in the latent space.
    Although models trained with adversarial techniques exhibit increased robustness and improved transferability as demonstrated by \citet{kotyan2022Transferabilitya}, they are not completely robust. 
    This is indicated by a gradual rise in the fragmentation with stronger perturbations, making a evident trade-off between robustness and accuracy as highlighted by \cite{tsipras2019Robustness,raghunathan2020understanding}. 

    \end{description}

    In summary, these findings highlight the neural network's sensitivity to variations in input samples, showcasing that a model perceives transformed samples as substantially different from their original counterparts.
    The results across various input transformations, including adversarial perturbations, indicate that current models struggle to cluster transformed samples together, interpreting them as distinct from the original sample and each other.
    Observations also indicate that white-box attacks such as PGD exploit this struggle by further fragmenting the distribution of samples into smaller fractures. 

\subsection{Effect of Different Distance Metrics on {k*~Distribution}}

    To evaluate the sensitivity of the nearest neighbor method to different distance metrics and understand their impact on {k*~distribution} and values, we compute the {k*~distribution} using Euclidean ($l_2$ norm), City Block ($l_1$ norm), Max Norm ($l_\infty$ norm), and Cosine Distances.
    The results from \Tableref{tab:distance} reveal the responsiveness of the {k*~distribution} in terms of metrics such as $\mu_{k^*}$, $\sigma_{k^*}$, and $\gamma_{k^*}$.
    Despite variations, substantial agreement exists on the sample distribution classification across different distance metrics.
    Having said that, this trend shifts when collectively assessing performance, as shown in \Tableref{tab:distance:summary}.
    Here, we observe minimal sensitivity in metric values like $\mu_{k^*}$ and $\gamma_{k^*}$ to choose distance metrics.
    In other words, the selection of distance metrics does impact the number of classes classified into multiple patterns.

    \begin{table*}[!t]
        \centering
        \caption{
        Performance using Different Distance Metrics across varied visual patterns.
        }
        \label{tab:distance:summary}
        \begin{tabular}{l|rrHr|rrHr|rrHr|rrHH}
        \toprule
        & \multicolumn{4}{c|}{\textbf{\shortstack{Average of\\ Classes with\\\PatternA\\~(\fractured)}}}
        & \multicolumn{4}{c|}{\textbf{\shortstack{Average of\\ Classes with\\\PatternB\\~(\overlapped)}}}
        & \multicolumn{4}{c|}{\textbf{\shortstack{Average of\\ Classes with\\\PatternC\\~(\clustered)}}}
        & \multicolumn{3}{c}{\textbf{\shortstack{Average of\\ 1,000\\ImageNet1k\\ Classes}}}
        \\ \textbf{Distance Metric}
        & $\mu_{k^*}$ & $\gamma_{k^*}$ & Acc & N
        & $\mu_{k^*}$ & $\gamma_{k^*}$ & Acc & N
        & $\mu_{k^*}$ & $\gamma_{k^*}$ & Acc & N
        & $\mu_{k^*}$ & $\gamma_{k^*}$ & Acc & N
        \\ \midrule

        Euclidean ($l_2$ norm)
        & 0.10 & 4.51 & 74.88 & 935
        & 0.37 & 0.14 & 93.96 & 56
        & 0.61 & -1.04 & 97.11 & 9
        & 0.12 & 4.21 & 76.15 & 1,000
        \\ CityBlock ($l_1$ norm)
        & 0.09 & 4.59 & 75.16 & 949
        & 0.37 & 0.15 & 93.95 & 42
        & 0.58 & -0.88 & 97.11 & 9
        & 0.11 & 4.35 & 76.15 & 1,000
        \\ Max Norm ($l_\infty$ norm)
        & 0.10 & 4.27 & 74.20 & 901
        & 0.39 & 0.10 & 93.46 & 82
        & 0.63 & -1.12 & 95.88 & 17
        & 0.14 & 3.83 & 76.15 & 1,000
        \\ Cosine
        & 0.10 & 4.30 & 74.49 & 915
        & 0.38 & 0.07 & 93.53 & 73
        & 0.62 & -1.14 & 96.83 & 12
        & 0.13 & 3.93 & 76.15 & 1,000

        \\ \bottomrule
        \end{tabular}
    \end{table*}

    \begin{table*}[!t]
        \centering
        \caption{
        Performance using different tasks and architectures across varied visual patterns
        }
        \label{tab:beyond:summary}
        \resizebox{\textwidth}{!}{
        \begin{tabular}{ll|rrrr|rrrr|rrrr|rrrr}
        \toprule
        && \multicolumn{4}{c|}{\textbf{\shortstack{Average of Classes with\\\PatternA~(\fractured)}}}
        & \multicolumn{4}{c|}{\textbf{\shortstack{Average of Classes with\\\PatternB~(\overlapped)}}}
        & \multicolumn{4}{c|}{\textbf{\shortstack{Average of Classes with\\\PatternC~(\clustered)}}}
        & \multicolumn{4}{c}{\textbf{\shortstack{Average on\\ Entire Dataset}}}
        \\ \textbf{Task} & \textbf{Architecture}
        & $\mu_{k^*}$ & $\gamma_{k^*}$ & Acc & N
        & $\mu_{k^*}$ & $\gamma_{k^*}$ & Acc & N
        & $\mu_{k^*}$ & $\gamma_{k^*}$ & Acc & N
        & $\mu_{k^*}$ & $\gamma_{k^*}$ & Acc & N
        \\ \midrule
        \multirow{4}{*}{\shortstack[l]{Intent\\Classification\\ on \cite{fitzgerald2022massive} (Text)}}  &
        DeBERTa V3 \cite{kubis2023back}
        & 0.18 & 2.68 & 64.64 & 24
        & 0.38 & -0.05 & 81.12 & 14
        & 0.65 & -1.29 & 83.73 & 21
        & 0.40 & 0.62 & 75.34 & 59
        \\ & XLM-RoBERTa \cite{kubis2023back}
        & 0.21 & 2.04 & 79.26 & 22
        & 0.40 & 0.07 & 64.00 & 8
        & 0.69 & -1.60 & 90.04 & 29
        & 0.47 & -0.02 & 82.49 & 59
        \\ & BERT \cite{kubis2023back}
        & 0.22 & 2.18 & 64.81 & 15
        & 0.40 & -0.09 & 87.29 & 8
        & 0.71 & -1.75 & 92.38 & 36
        & 0.54 & -0.53 & 84.68 & 59
        \\ & Multilingual-MiniLM \cite{kubis2023back}
        & 0.20 & 2.17 & 80.22 & 15
        & 0.42 & -0.14 & 82.87 & 11
        & 0.71 & -1.69 & 91.33 & 33
        & 0.53 & -0.42 & 86.93 & 59
        \\ \midrule
        \multirow{2}{*}{\shortstack[l]{Keyword Spotting \\ on \cite{speechcommandsv2} (Audio)}} &
         AST \cite{2023moonseok}
        & 0.26 & 1.36 & 70.00 & 1
        & 0.43 & 0.00 & 58.83 & 3
        & 0.77 & -1.89 & 95.18 & 32
        & 0.73 & -1.65 & 91.45 & 36
        \\ & Wav2Vec2-Conformer-L \cite{2023juliensimon}
        & 0.22 & 1.82 & 92.00 & 1
        & 0.38 & 0.01 & 40.00 & 2
        & 0.92 & -3.81 & 98.04 & 33
        & 0.87 & -3.44 & 94.65 & 36

        \\ \bottomrule
        \end{tabular}}
    \end{table*}

\subsection{Going Beyond Image Classification}

    Exploring beyond the domain of computer vision, we choose to apply the {k*~distribution} to models trained for tasks other than image classification.
    Specifically, we evaluate models trained for classifying intent from MASSIVE dataset \cite{fitzgerald2022massive} in the domain of natural language processing,
    and models trained for keyword spotting from the Speech Commands dataset \cite{speechcommandsv2} in the domain of speech processing as shown in \Tableref{tab:beyond:summary}.
    We note the wider applicability of {k*~distribution} to multiple domains and observe a similar trend of higher $\mu_{k^*}$ and lower $\gamma_{k^*}$ for better accuracy.
    Interestingly, we observe that the models trained for intent classification and keyword spotting were less fractured than image classification, suggesting that neural networks' feature and label association for such tasks can be further improved.

\section{Conclusion}

    In this article, we introduce the {k*~distribution}, a methodology grounded in the local neighborhood, to assess the distribution of samples in the learned latent space of neural networks.
    Our experimental findings indicate that the distribution can be primarily categorized into three distinct patterns:
    \PatternA~representing \fractured~distribution of samples in latent space: Identified by a positively skewed {k*~distribution}.
    \PatternB~representing \overlapped~distribution of samples in latent space: Characterized by a nearly uniform {k*~distribution}.
    \PatternC~representing \clustered~distribution of samples in latent space: Indicated by a negatively skewed {k*~distribution}.
    Using {k*~distribution}, we analyzed the learned latent space of different models, with different training data distributions, and of different layers.
    Further, the learned latent space was tested against various input transformations to understand how the input transformations affect the learned latent space and are associated with the training distribution.
    We hope this methodology and analysis will help understand the neural networks' latent spaces and improve the distributions of samples in latent space.

\section*{Acknowledgments}
    \noindent
    This work was supported by JSPS Grant-in-Aid for Challenging Exploratory Research - Grant Number JP22K19814, JST Strategic Basic Research Promotion Program (AIP Accelerated Research) - Grant Number JP22584686, JSPS Research on Academic Transformation Areas (A) - Grant Number JP22H05194.

\small{
\bibliographystyle{IEEEtranN}
\bibliography{TNNLS-2023-P-26404-R2}
}

\clearpage
\onecolumn
\appendix
\section*{Extra Visualizations}
    Here, we provide the various visualizations of the latent space for the different cases, we investigated in the main article.
    We provide visualizations for the {k*~distribution}, t-SNE, Isomap, PCA and UMAP of all the classes of \SubImageNet~dataset ($1,280$ samples).
    Further, to visualize the local neighbor space, we also provide the neighbor distribution of all the classes of \SubImageNet~dataset.

    \clearpage
    \begin{figure*}[p]
        \centering
        \includegraphics[width=\linewidth]{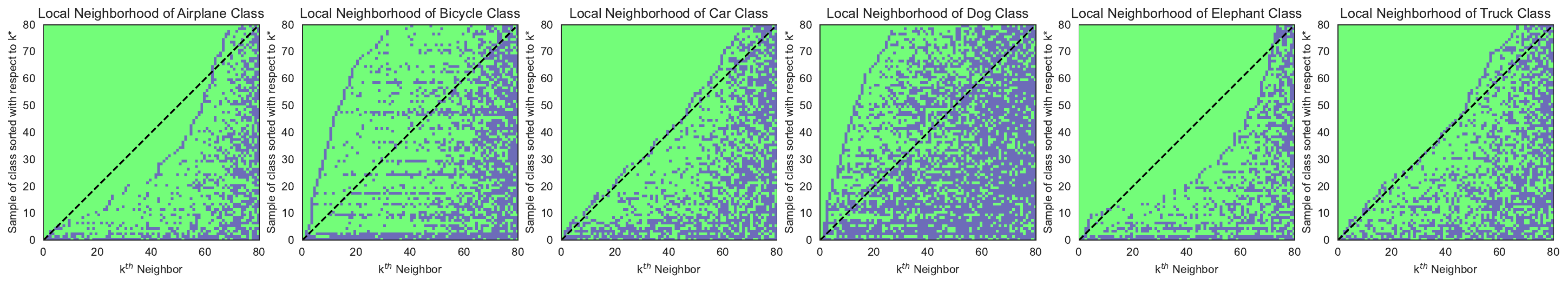} \\
        \includegraphics[width=\linewidth]{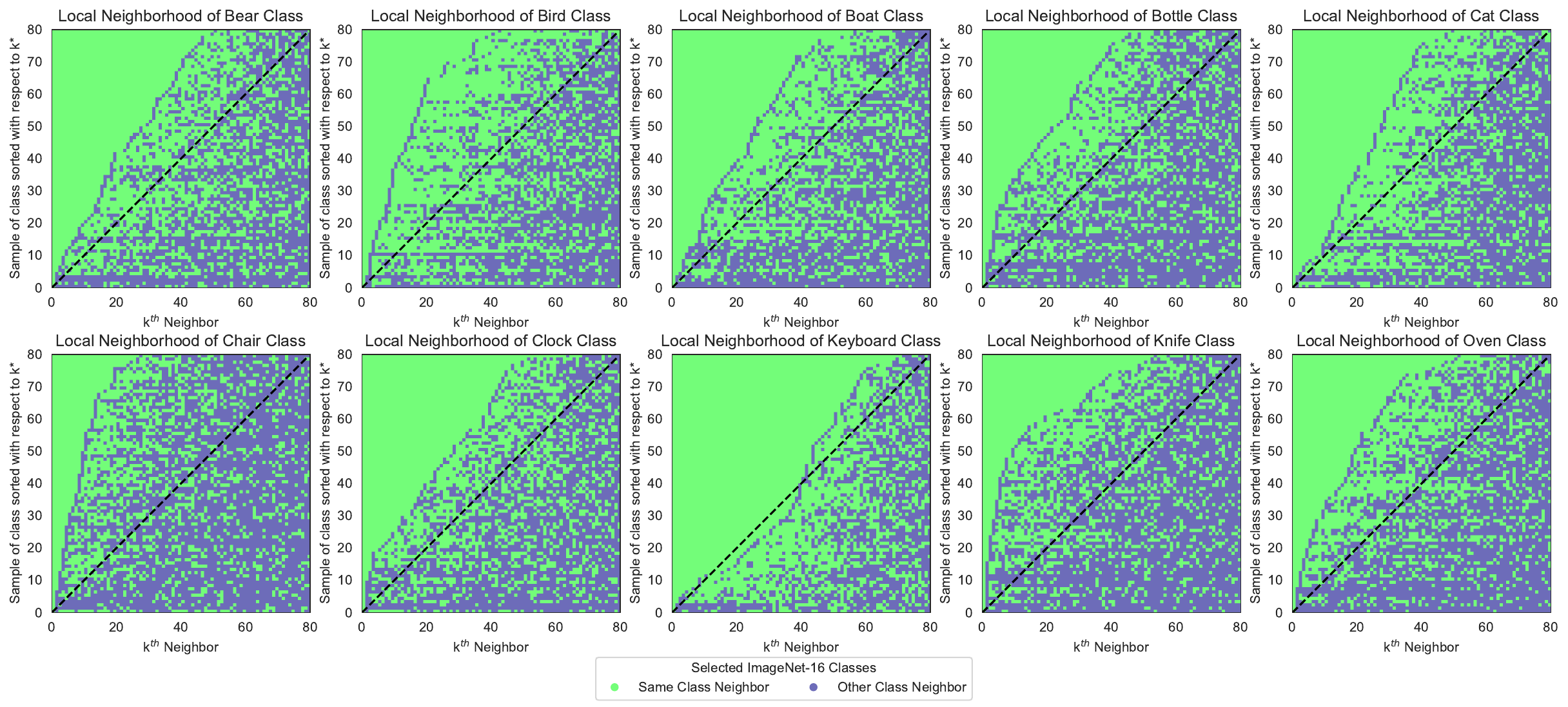}
        \caption{
        We visualize the neighbor distribution of all samples of a class for ResNet-50 architecture \cite{he2016deep} (see \Tableref{tab:architectures}).
        The green color represents that the neighbor to the sample belongs to the same class as the testing sample, while the gray color represents that the neighbor belongs to a different class compared to the testing sample.
        A \fractured~distribution of samples will have different class neighbors above the diagonal (black dashed line);
        An \overlapped~distribution of samples will first different class neighbors around the diagonal, and;
        A \clustered~distribution of samples will have different class neighbors below the diagonal.
        }
        \vspace{0.5em}
        \includegraphics[width=0.49\linewidth]{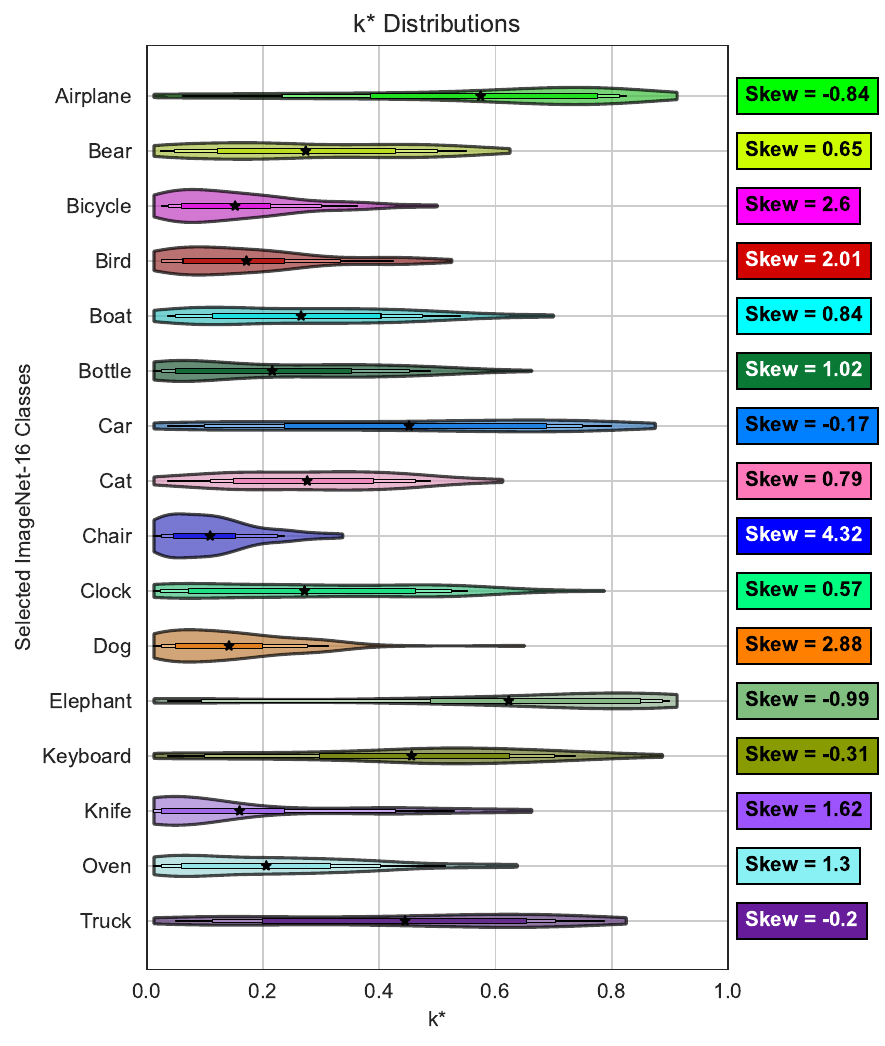}
        \rulesep
        \includegraphics[width=0.49\linewidth]{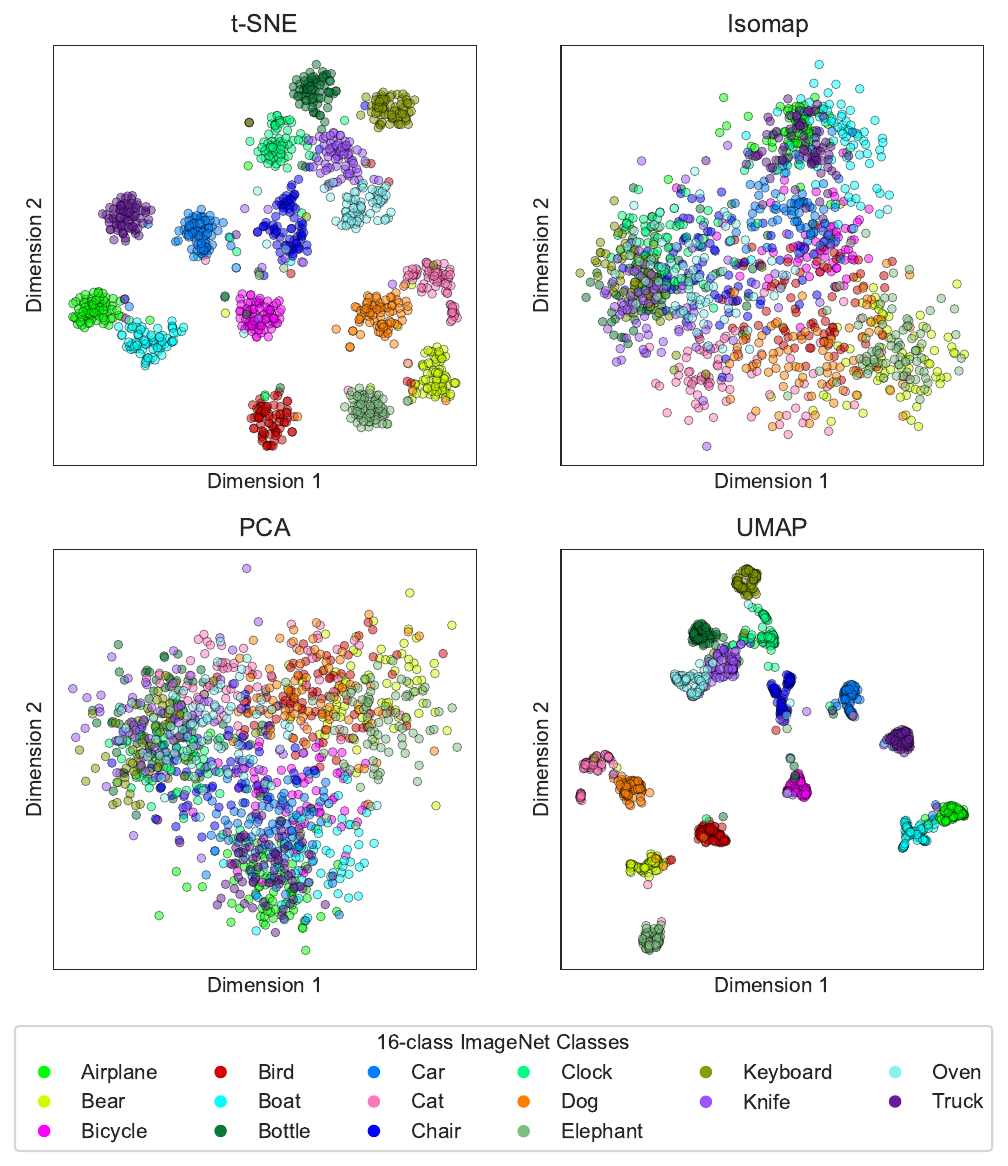}
        \caption{
               Visualization of the distribution of samples in latent space using, \figleft~{k*~distribution}, and \figright~Dimensionality Reduction techniques like t-SNE \figtopleft, Isomap \figtopright, PCA \figbottomleft, and UMAP \figbottomright~ of all classes of 16-class-ImageNet for the Logit Layer of ResNet-50 Architecture \cite{he2016deep} (see \Tableref{tab:architectures}).
        }
    \end{figure*}

    \clearpage
    \begin{figure*}[p]
        \centering
        \includegraphics[width=\linewidth]{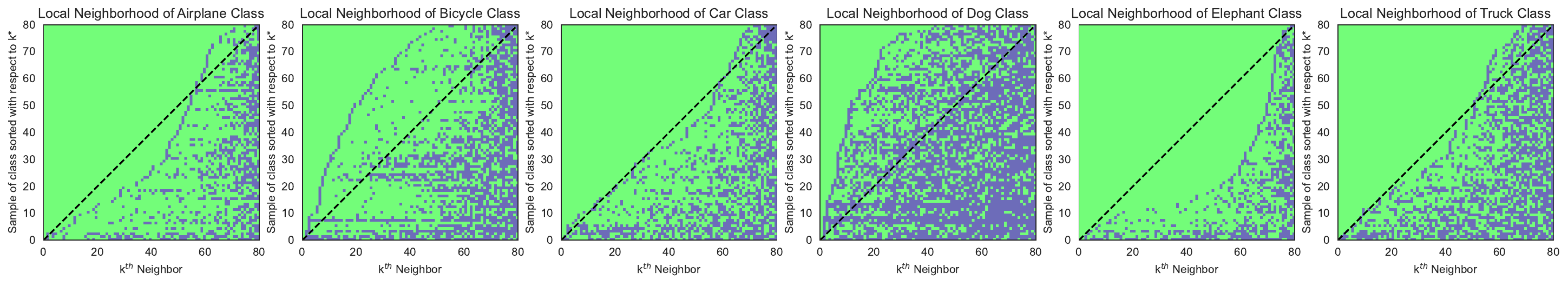} \\
        \includegraphics[width=\linewidth]{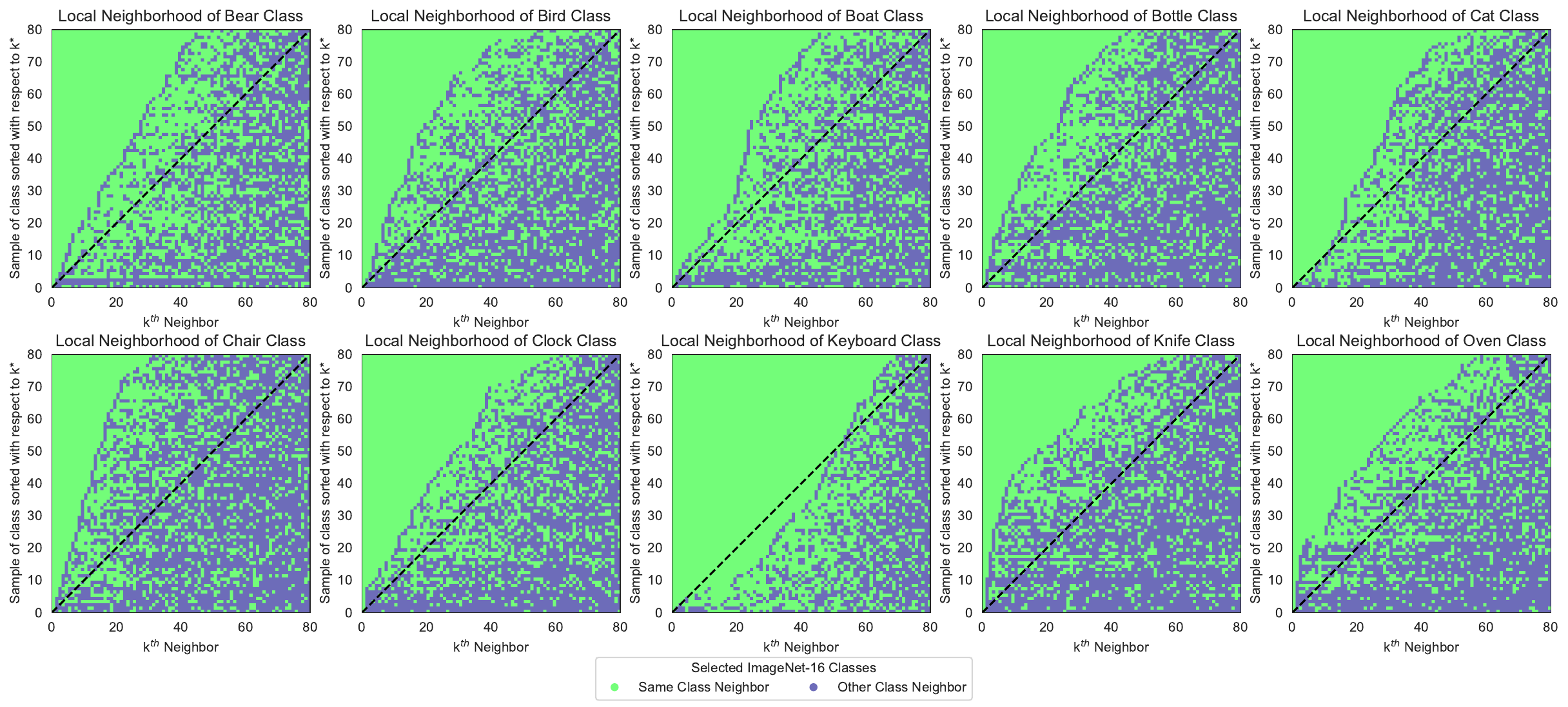}
        \caption{
        We visualize the neighbor distribution of all samples of a class for ResNeXt-101 architecture \cite{xie2017aggregated} (see \Tableref{tab:architectures}).
        The green color represents that the neighbor to the sample belongs to the same class as the testing sample, while the gray color represents that the neighbor belongs to a different class compared to the testing sample.
        A \fractured~distribution of samples will have different class neighbors above the diagonal (black dashed line);
        An \overlapped~distribution of samples will first different class neighbors around the diagonal, and;
        A \clustered~distribution of samples will have different class neighbors below the diagonal.
        }
        \vspace{0.5em}
        \includegraphics[width=0.49\linewidth]{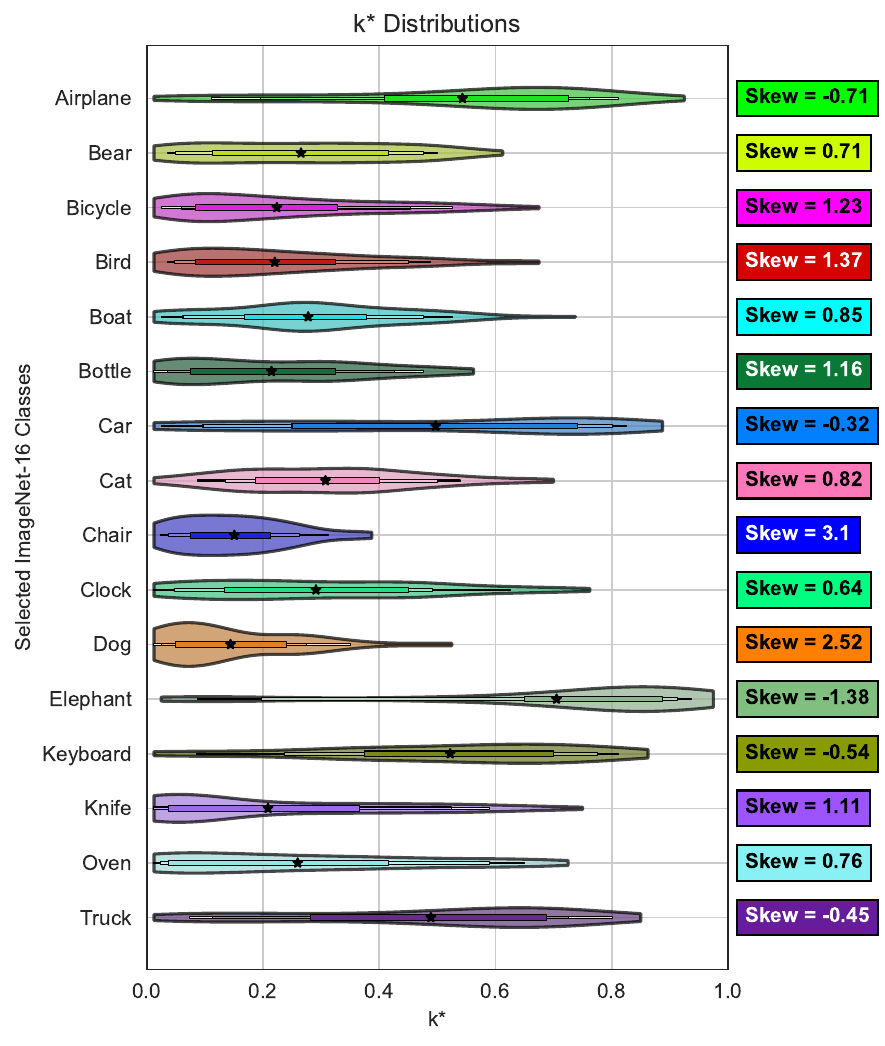}
        \rulesep
        \includegraphics[width=0.49\linewidth]{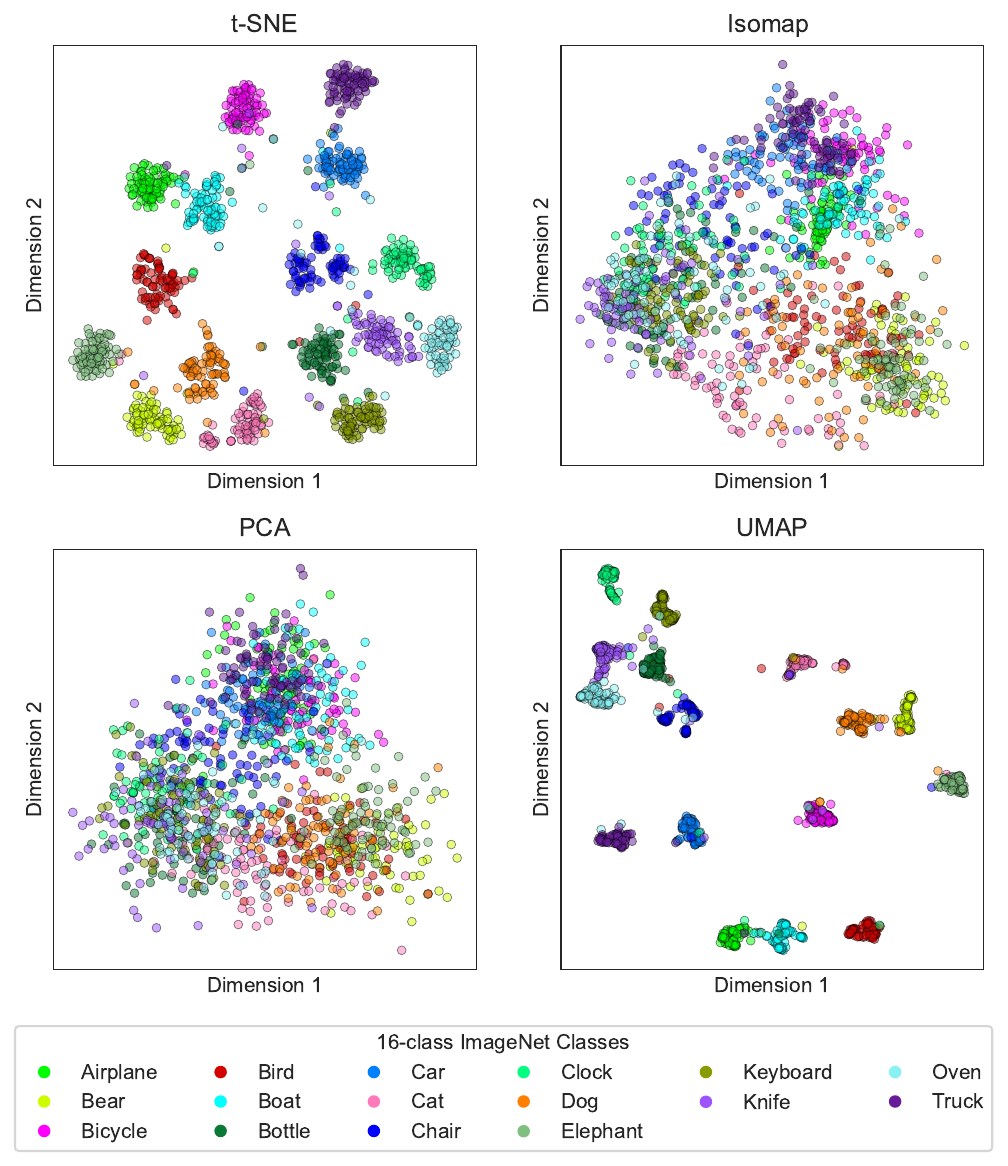}
        \caption{
               Visualization of the distribution of samples in latent space using, \figleft~{k*~distribution}, and \figright~Dimensionality Reduction techniques like t-SNE \figtopleft, Isomap \figtopright, PCA \figbottomleft, and UMAP \figbottomright~ of all classes of 16-class-ImageNet for the Logit Layer of ResNeXt-101 Architecture \cite{xie2017aggregated} (see \Tableref{tab:architectures}).
        }
    \end{figure*}

    \clearpage
    \begin{figure*}[p]
        \centering
        \includegraphics[width=\linewidth]{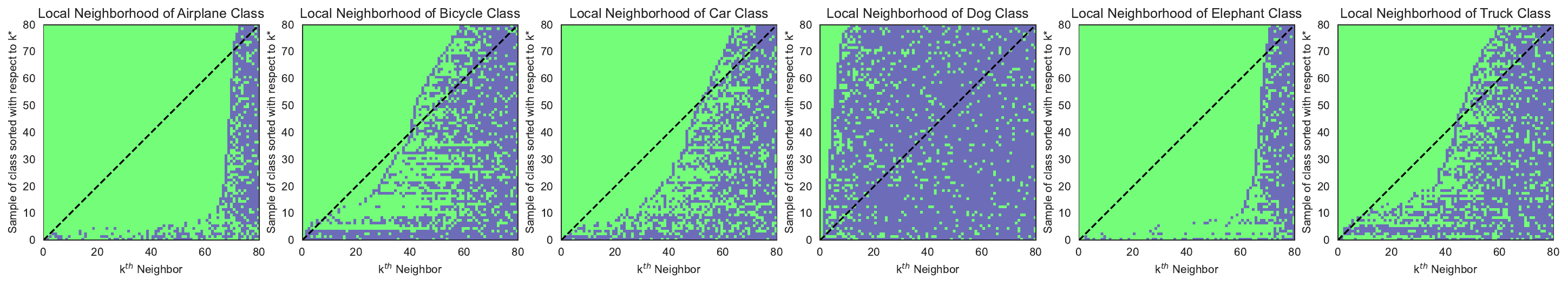} \\
        \includegraphics[width=\linewidth]{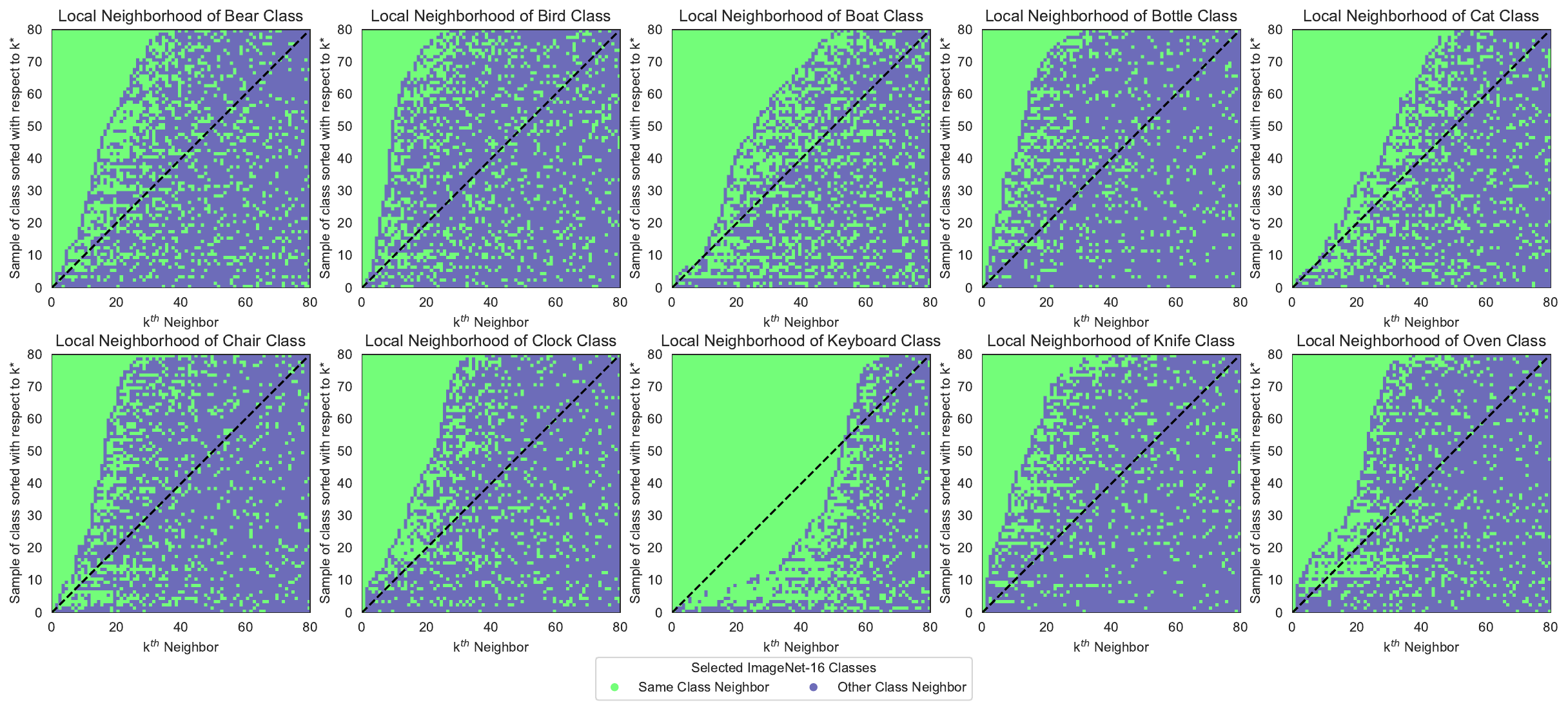}
        \caption{
        We visualize the neighbor distribution of all samples of a class for EfficientNet-B0 \cite{tan2019efficientnet} (see \Tableref{tab:architectures}).
        The green color represents that the neighbor to the sample belongs to the same class as the testing sample, while the gray color represents that the neighbor belongs to a different class compared to the testing sample.
        A \fractured~distribution of samples will have different class neighbors above the diagonal (black dashed line);
        An \overlapped~distribution of samples will first different class neighbors around the diagonal, and;
        A \clustered~distribution of samples will have different class neighbors below the diagonal.
        }
        \vspace{0.5em}
        \includegraphics[width=0.49\linewidth]{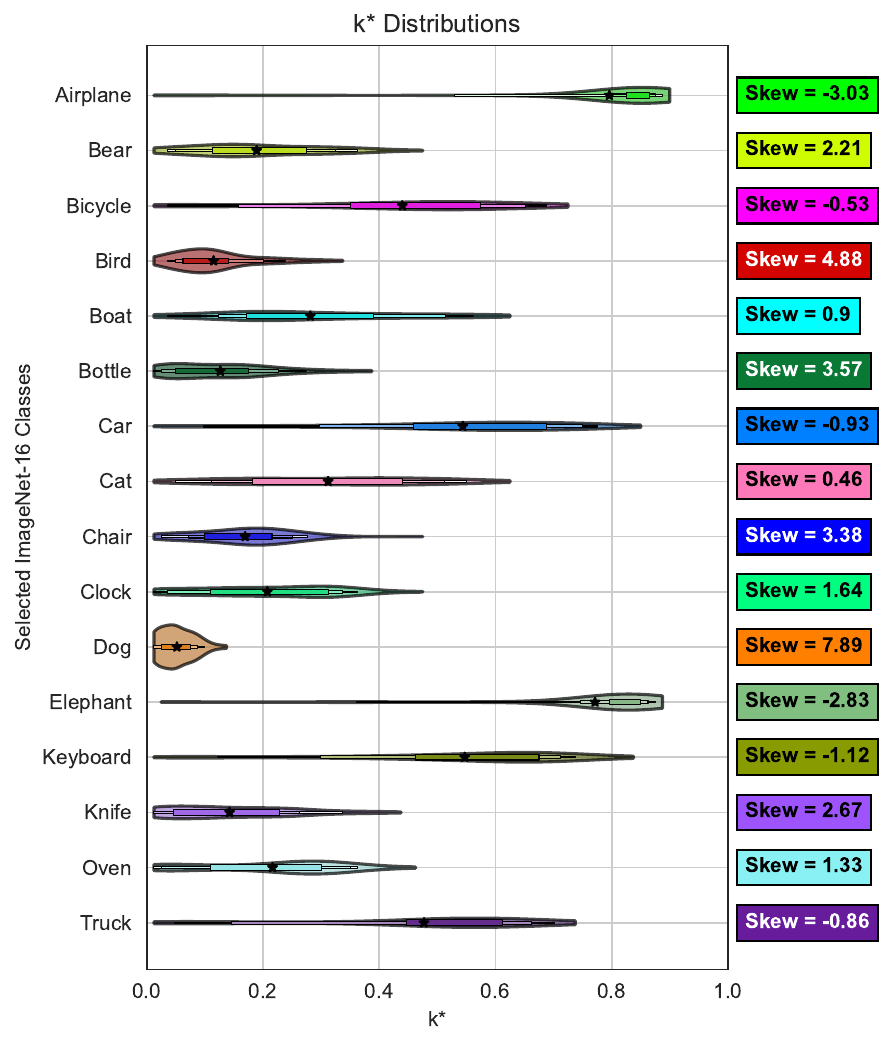}
        \rulesep
        \includegraphics[width=0.49\linewidth]{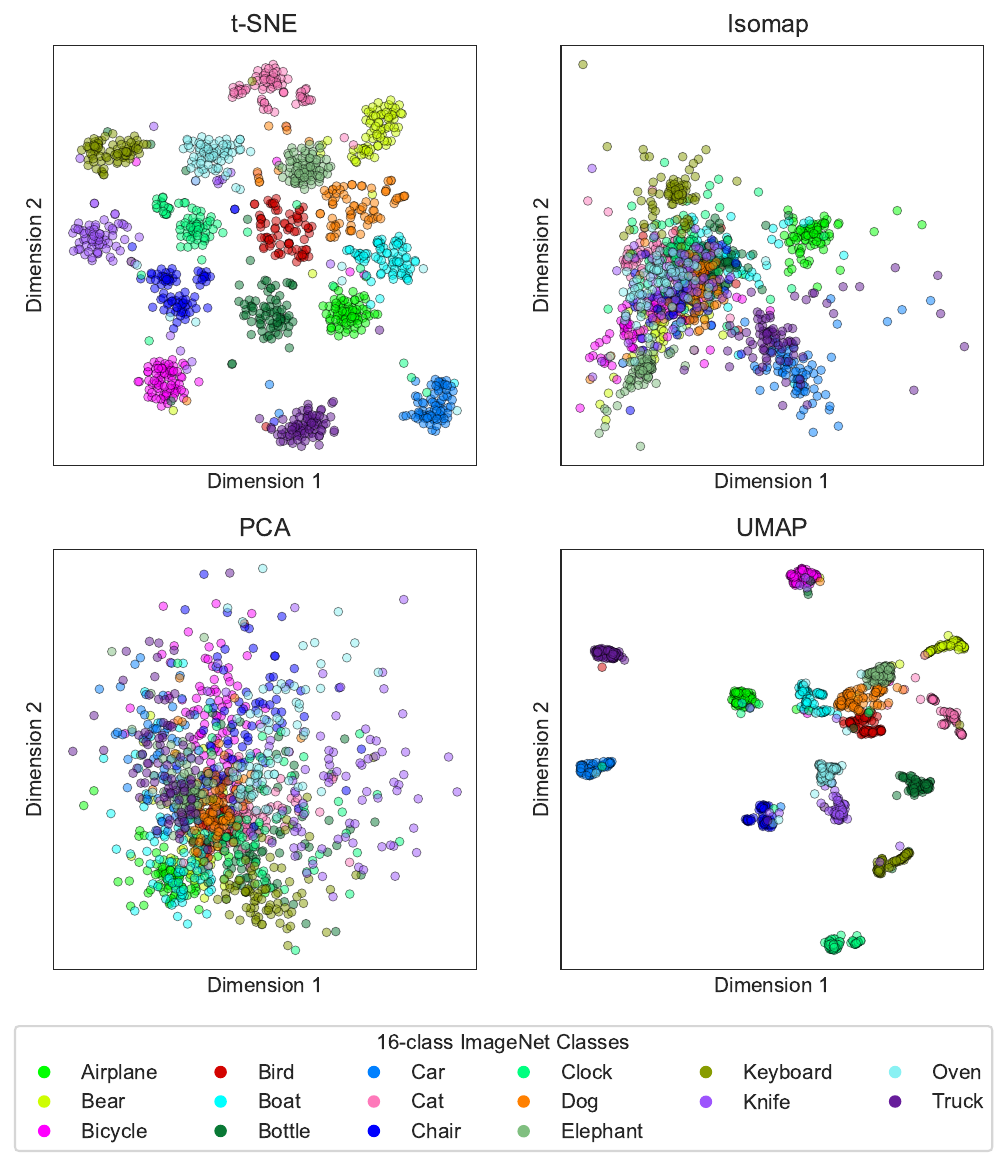}
        \caption{
               Visualization of the distribution of samples in latent space using, \figleft~{k*~distribution}, and \figright~Dimensionality Reduction techniques like t-SNE \figtopleft, Isomap \figtopright, PCA \figbottomleft, and UMAP \figbottomright~ of all classes of 16-class-ImageNet for the Logit Layer of EfficientNet-B0 Architecture \cite{tan2019efficientnet} (see \Tableref{tab:architectures}).
        }
    \end{figure*}

    \clearpage
    \begin{figure*}[p]
        \centering
        \includegraphics[width=\linewidth]{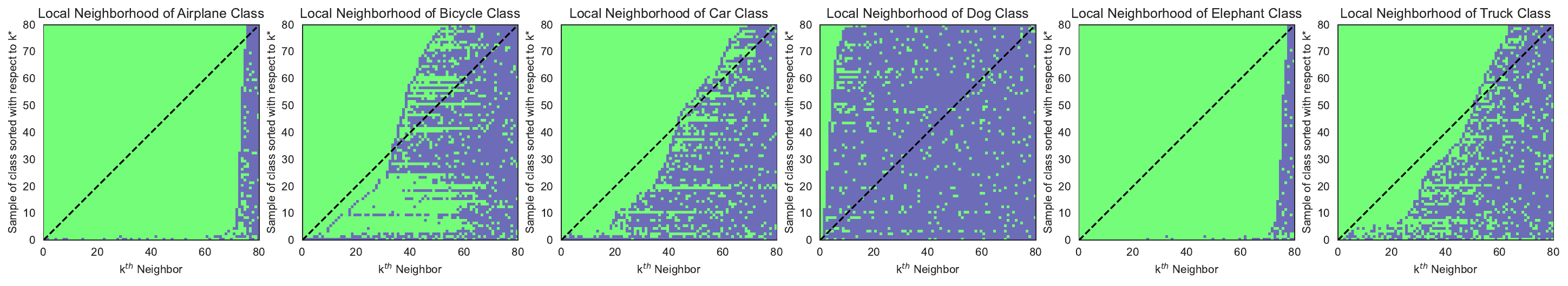} \\
        \includegraphics[width=\linewidth]{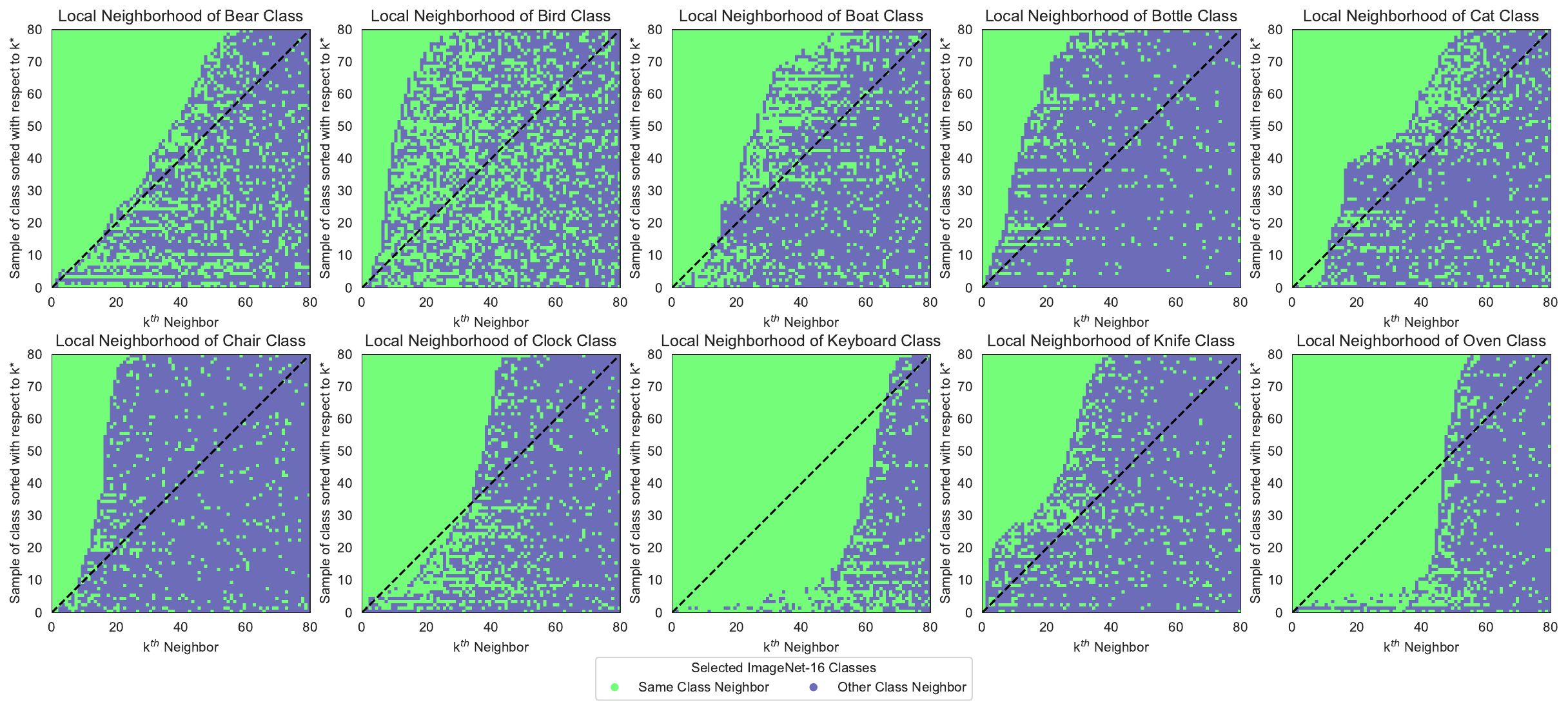}
        \caption{
        We visualize the neighbor distribution of all samples of a class for ViT-B \cite{dosovitskiy2020image} (see \Tableref{tab:architectures}).
        The green color represents that the neighbor to the sample belongs to the same class as the testing sample, while the gray color represents that the neighbor belongs to a different class compared to the testing sample.
        A \fractured~distribution of samples will have different class neighbors above the diagonal (black dashed line);
        An \overlapped~distribution of samples will first different class neighbors around the diagonal, and;
        A \clustered~distribution of samples will have different class neighbors below the diagonal.
        }
        \vspace{0.5em}
        \includegraphics[width=0.49\linewidth]{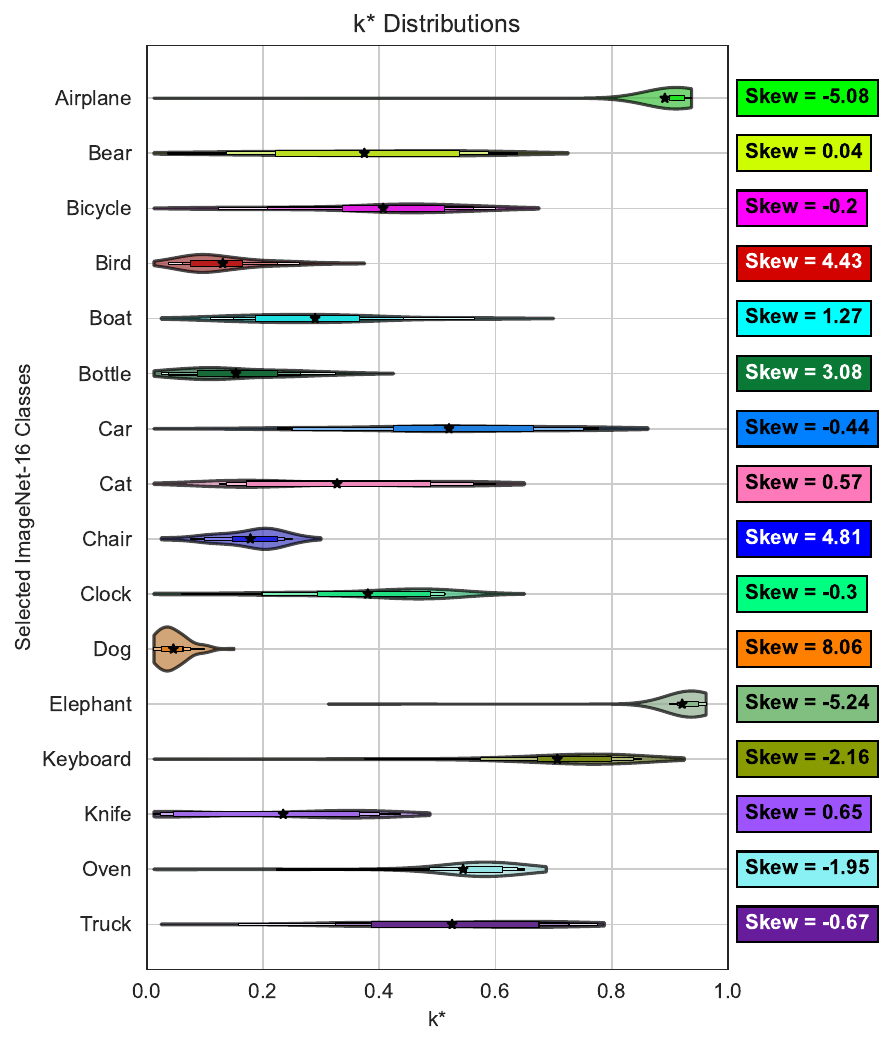}
        \rulesep
        \includegraphics[width=0.49\linewidth]{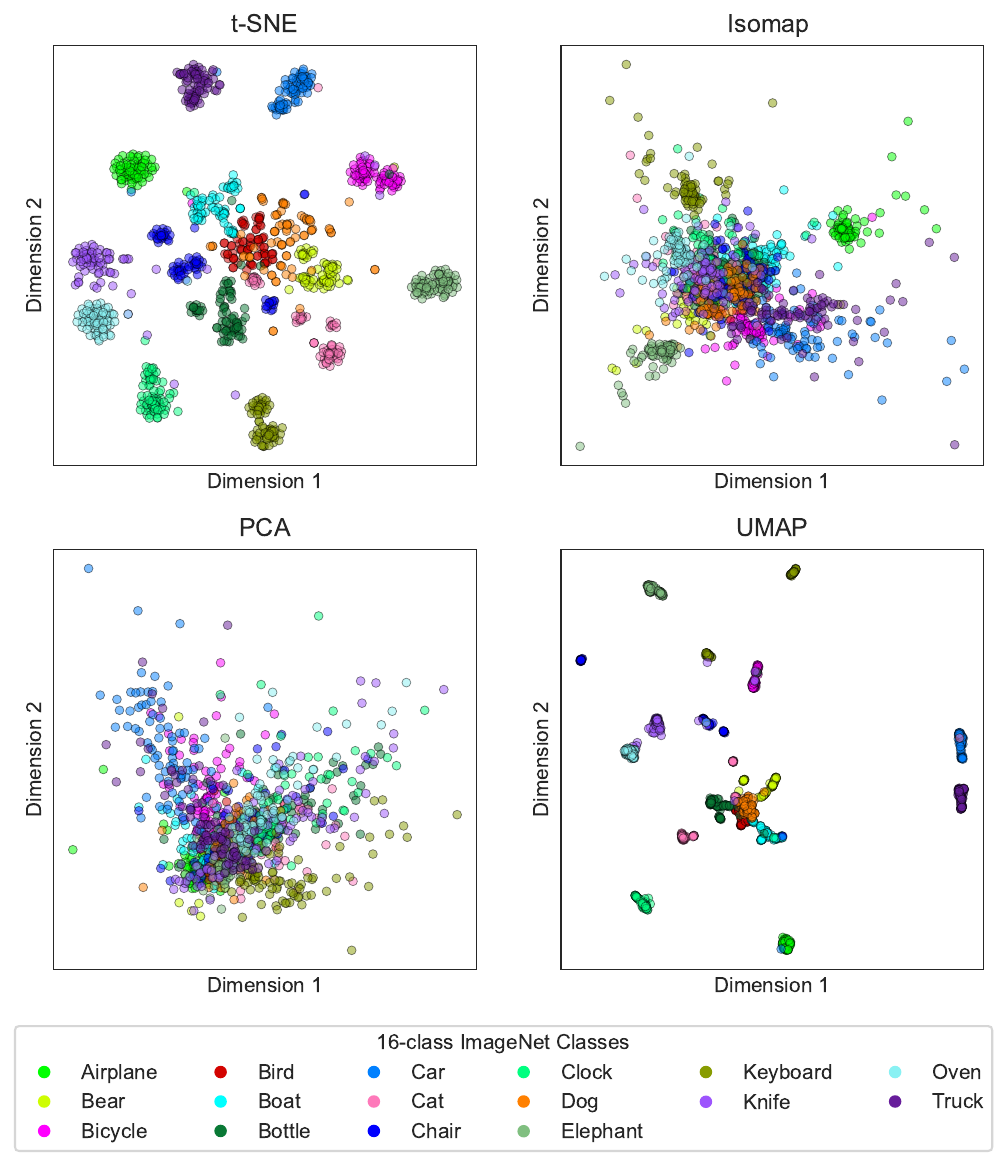}
        \caption{
               Visualization of the distribution of samples in latent space using, \figleft~{k*~distribution}, and \figright~Dimensionality Reduction techniques like t-SNE \figtopleft, Isomap \figtopright, PCA \figbottomleft, and UMAP \figbottomright~ of all classes of 16-class-ImageNet for the Logit Layer of ViT-B Architecture \cite{dosovitskiy2020image} (see \Tableref{tab:architectures}).
        }
    \end{figure*}

    \clearpage
    \begin{figure*}[p]
        \centering
        \includegraphics[width=\linewidth]{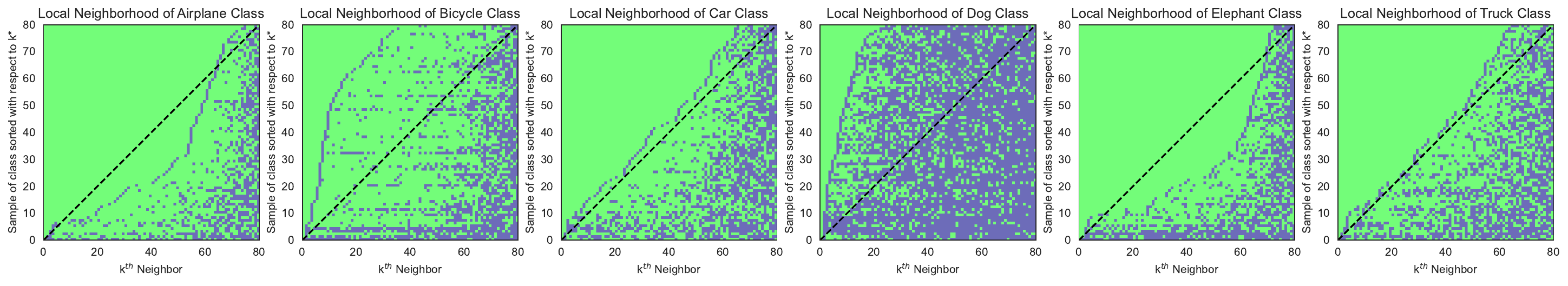} \\
        \includegraphics[width=\linewidth]{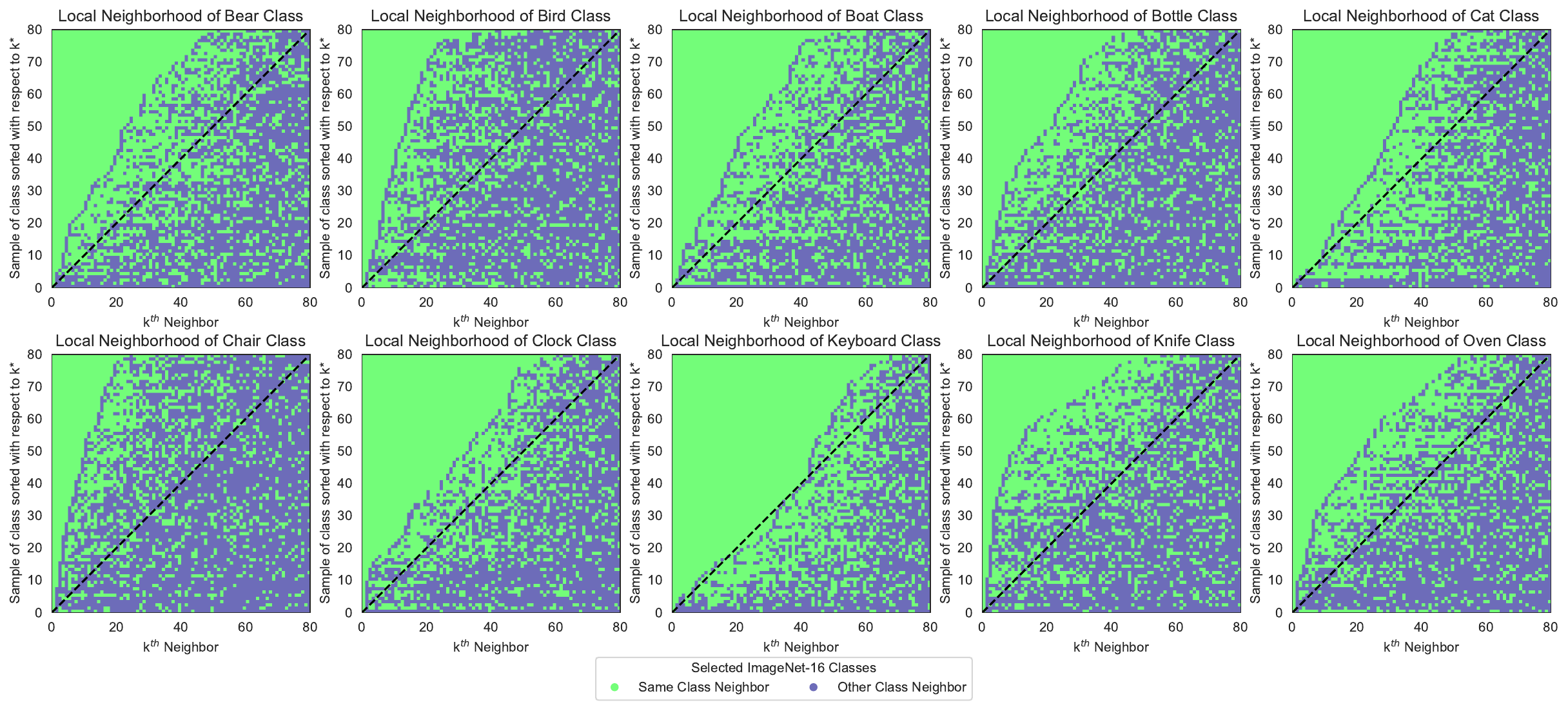}
        \caption{
        We visualize the neighbor distribution of all samples of a class for the Average Pooling Layer of ResNet-50 \cite{he2016deep} (see \Tableref{tab:layer}).
        The green color represents that the neighbor to the sample belongs to the same class as the testing sample, while the gray color represents that the neighbor belongs to a different class compared to the testing sample.
        A \fractured~distribution of samples will have different class neighbors above the diagonal (black dashed line);
        An \overlapped~distribution of samples will first different class neighbors around the diagonal, and;
        A \clustered~distribution of samples will have different class neighbors below the diagonal.
        }
        \vspace{0.5em}
        \includegraphics[width=0.49\linewidth]{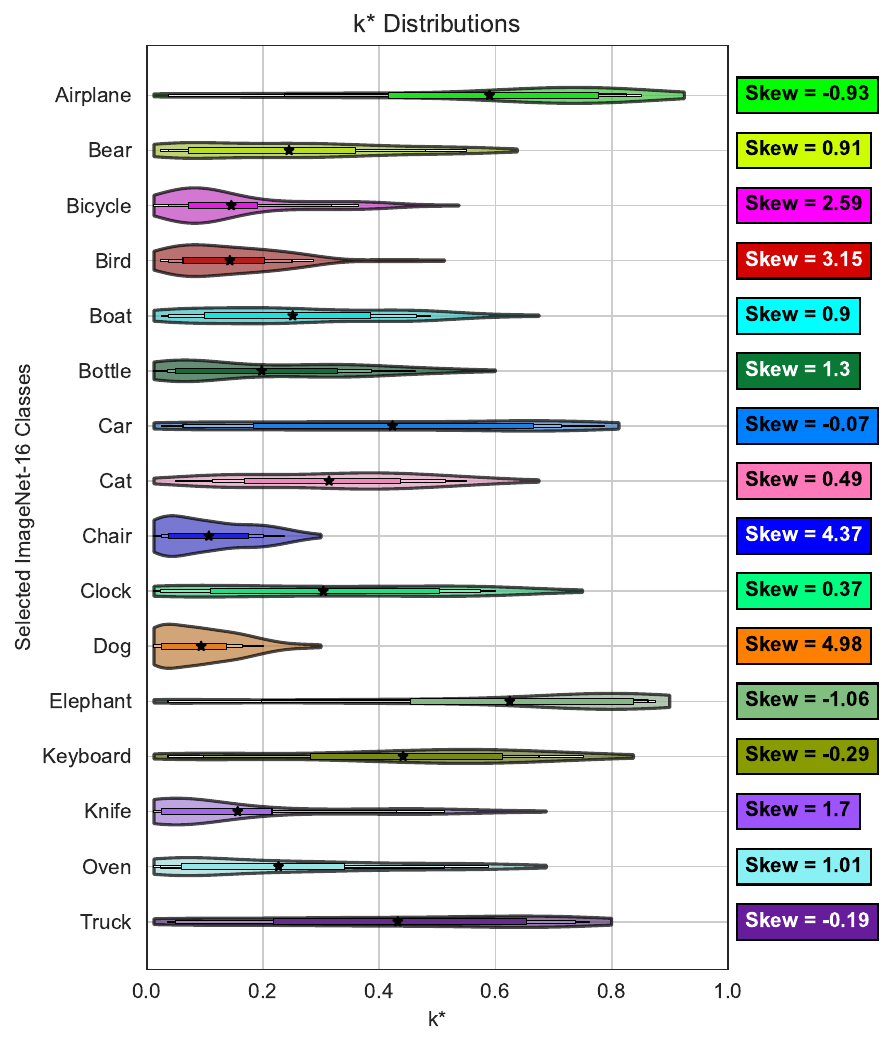}
        \rulesep
        \includegraphics[width=0.49\linewidth]{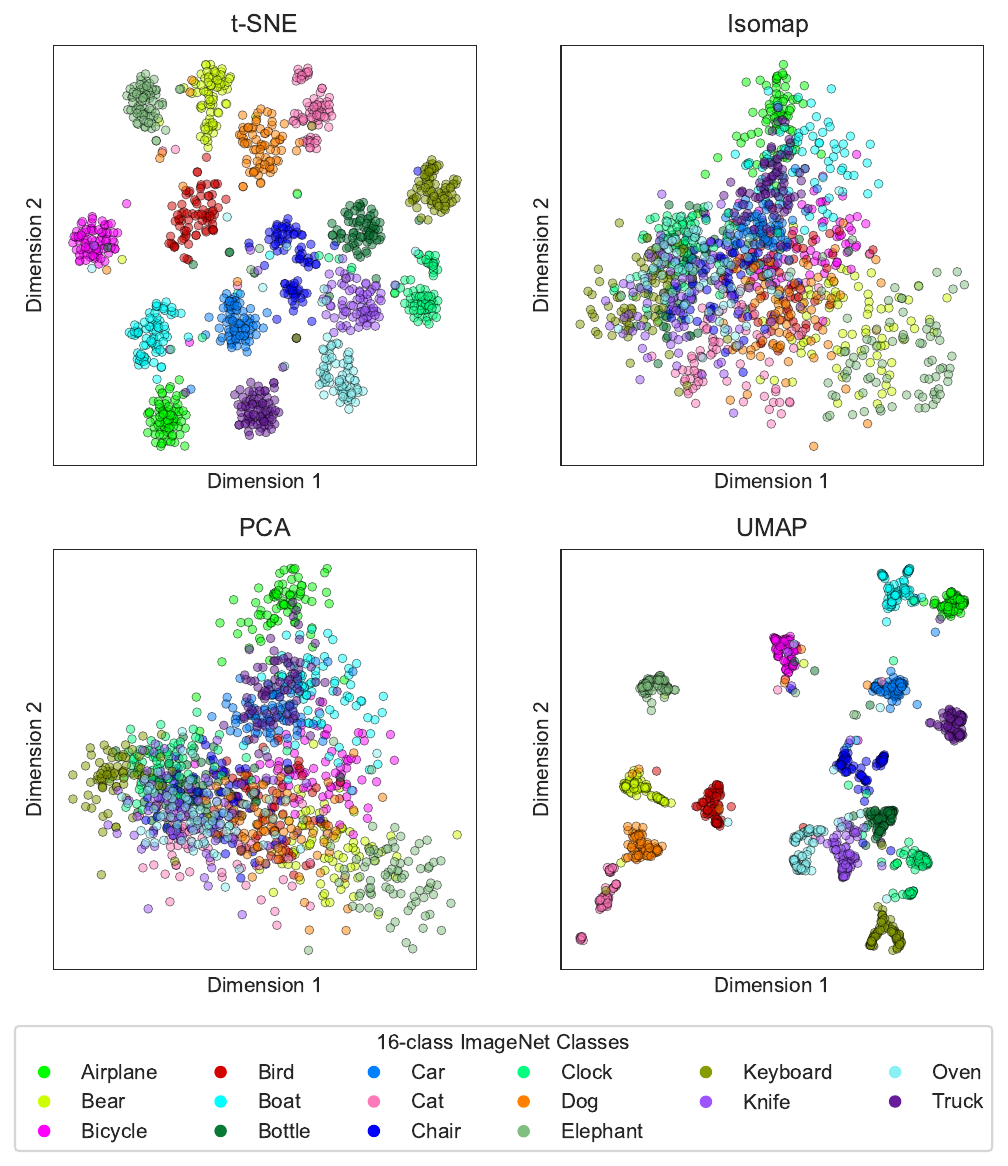}
        \caption{
               Visualization of the distribution of samples in latent space using, \figleft~{k*~distribution}, and \figright~Dimensionality Reduction techniques like t-SNE \figtopleft, Isomap \figtopright, PCA \figbottomleft, and UMAP \figbottomright~ of all classes of 16-class-ImageNet for the Average Pooling Layer of ResNet-50 \cite{he2016deep} (see \Tableref{tab:layer}).
        }
    \end{figure*}

    \clearpage
    \begin{figure*}[p]
        \centering
        \includegraphics[width=\linewidth]{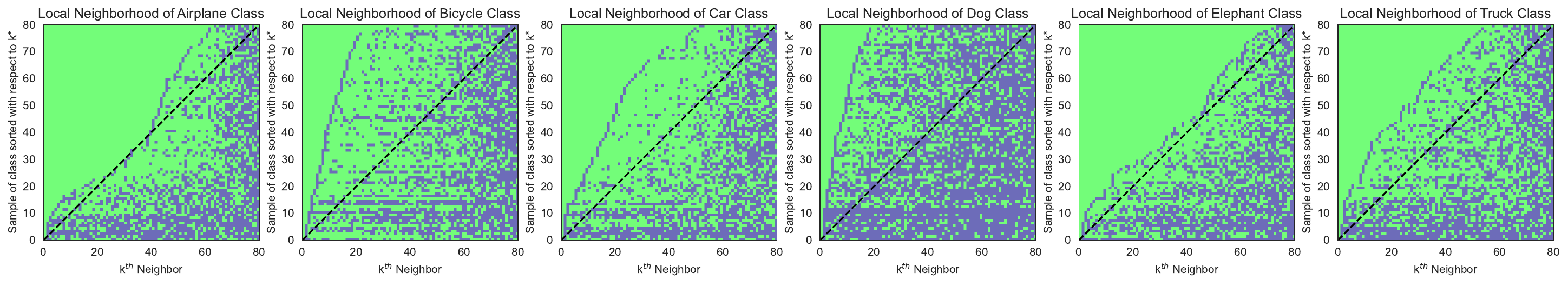} \\
        \includegraphics[width=\linewidth]{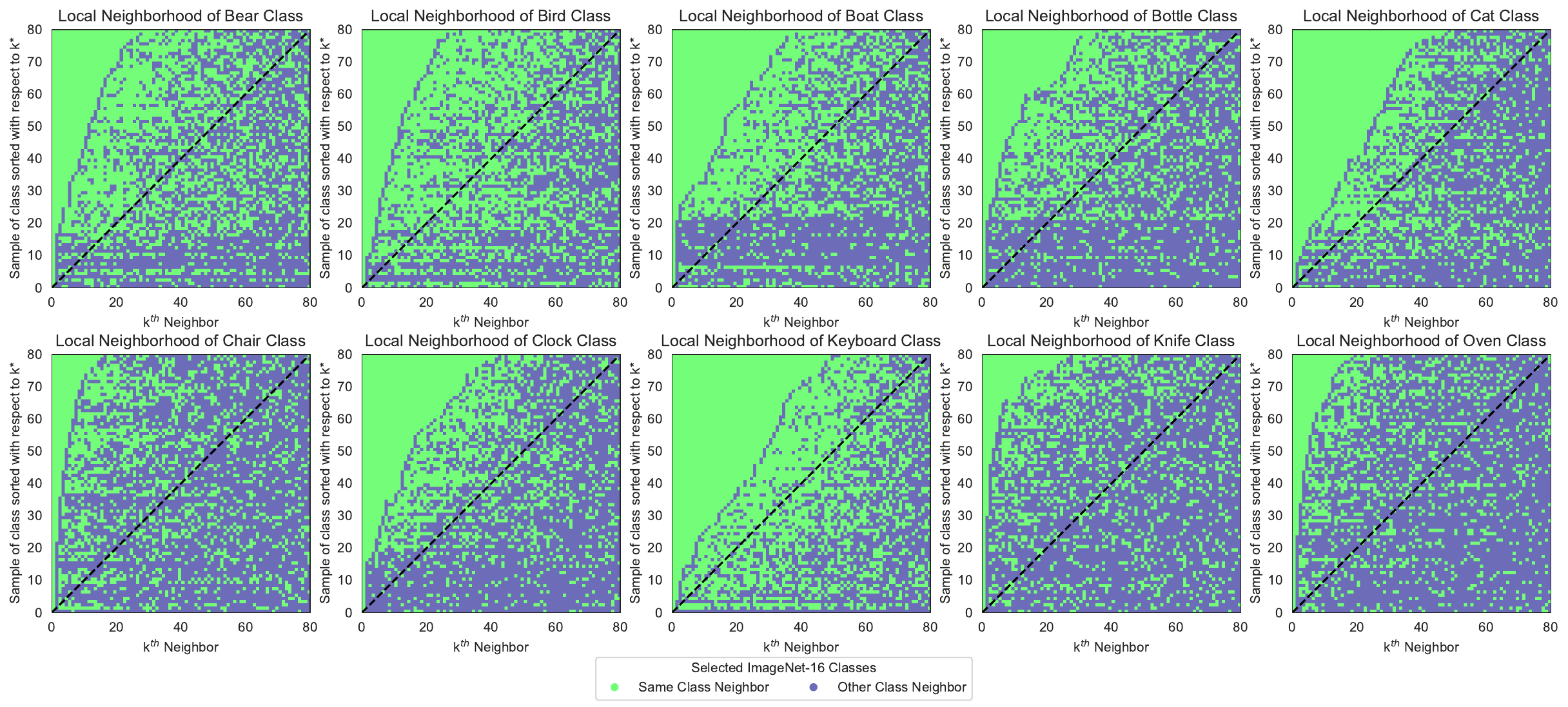}
        \caption{
        We visualize the neighbor distribution of all samples of a class for ResNet-50 trained on Stylized ImageNet \cite{geirhos2018} (see \Tableref{tab:training}).
        The green color represents that the neighbor to the sample belongs to the same class as the testing sample, while the gray color represents that the neighbor belongs to a different class compared to the testing sample.
        A \fractured~distribution of samples will have different class neighbors above the diagonal (black dashed line);
        An \overlapped~distribution of samples will first different class neighbors around the diagonal, and;
        A \clustered~distribution of samples will have different class neighbors below the diagonal.
        }
        \vspace{0.5em}
        \includegraphics[width=0.49\linewidth]{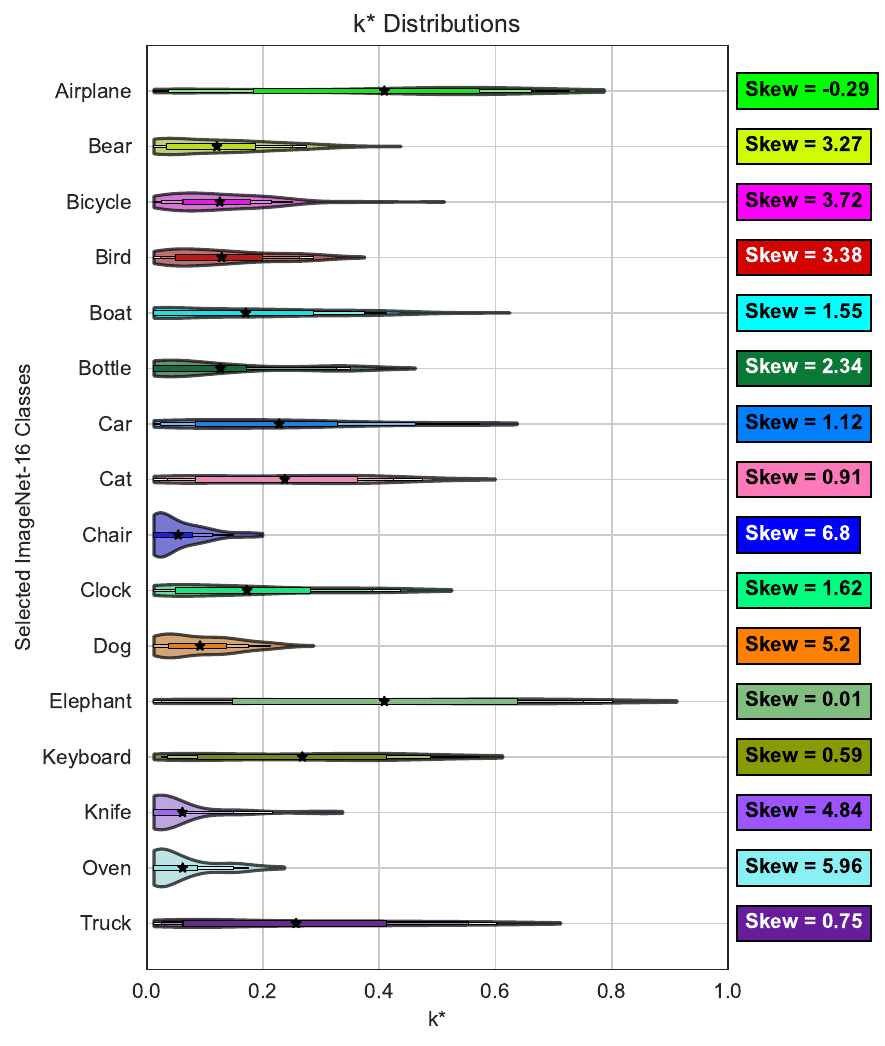}
        \rulesep
        \includegraphics[width=0.49\linewidth]{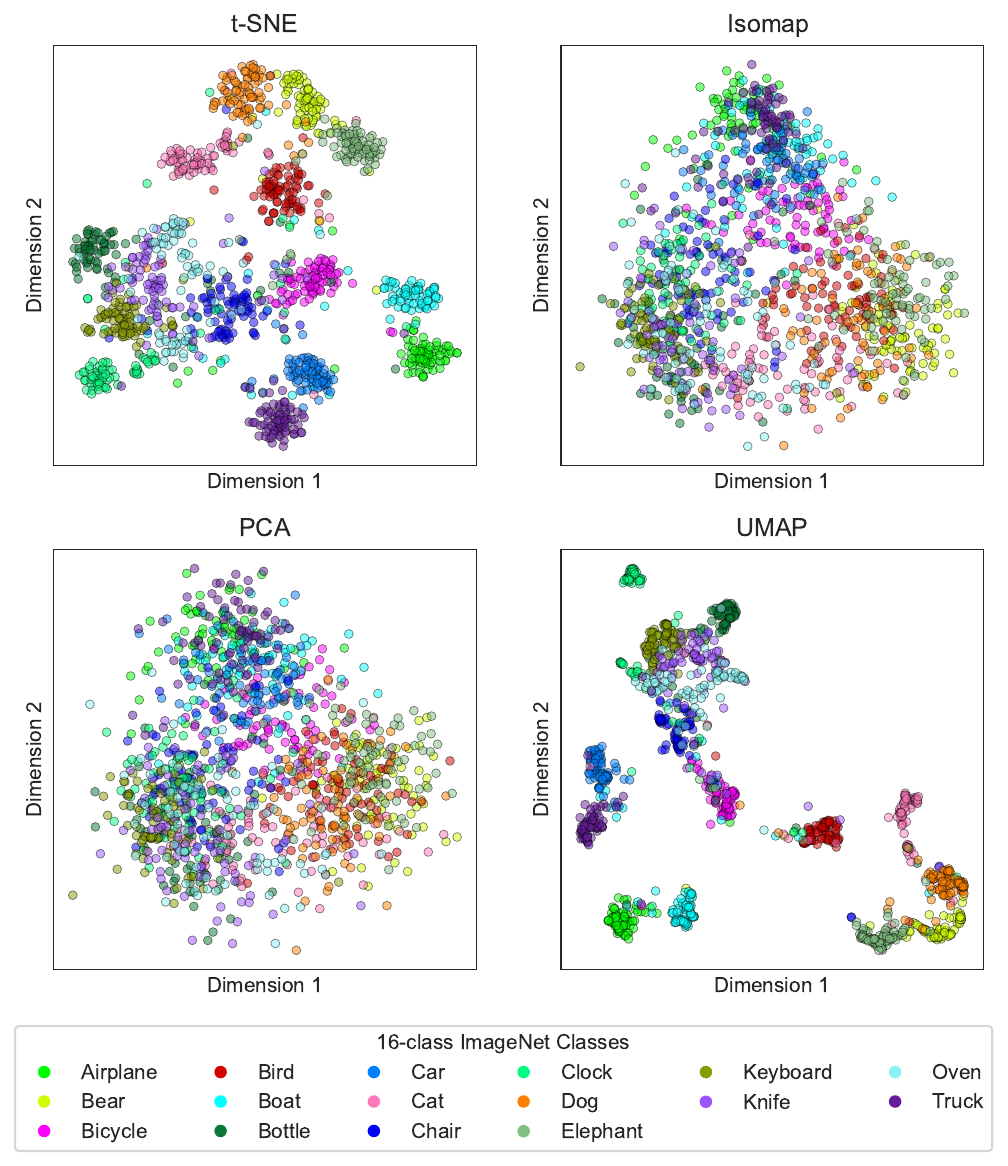}
        \caption{
               Visualization of the distribution of samples in latent space using, \figleft~{k*~distribution}, and \figright~Dimensionality Reduction techniques like t-SNE \figtopleft, Isomap \figtopright, PCA \figbottomleft, and UMAP \figbottomright~ of all classes of 16-class-ImageNet for the Logit Layer of ResNet-50 trained on Stylized ImageNet \cite{geirhos2018} (see \Tableref{tab:training}).
        }
    \end{figure*}

    \clearpage
    \begin{figure*}[p]
        \centering
        \includegraphics[width=\linewidth]{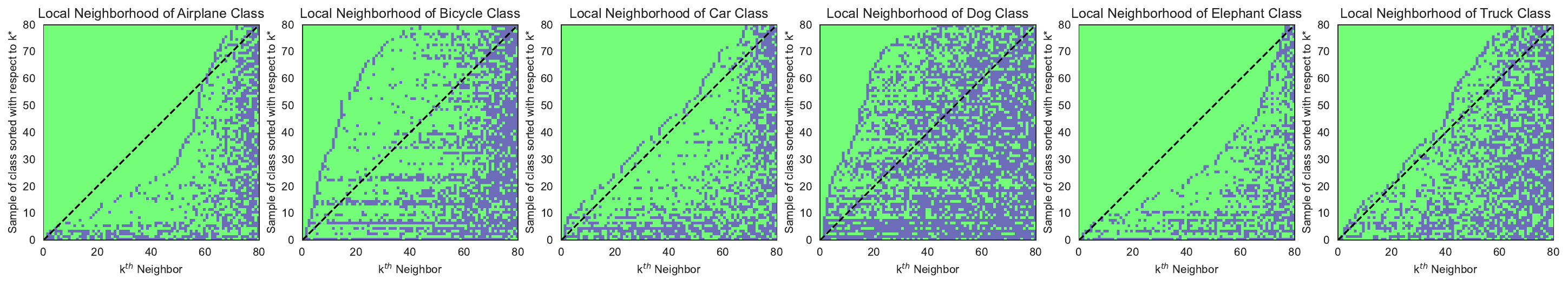} \\
        \includegraphics[width=\linewidth]{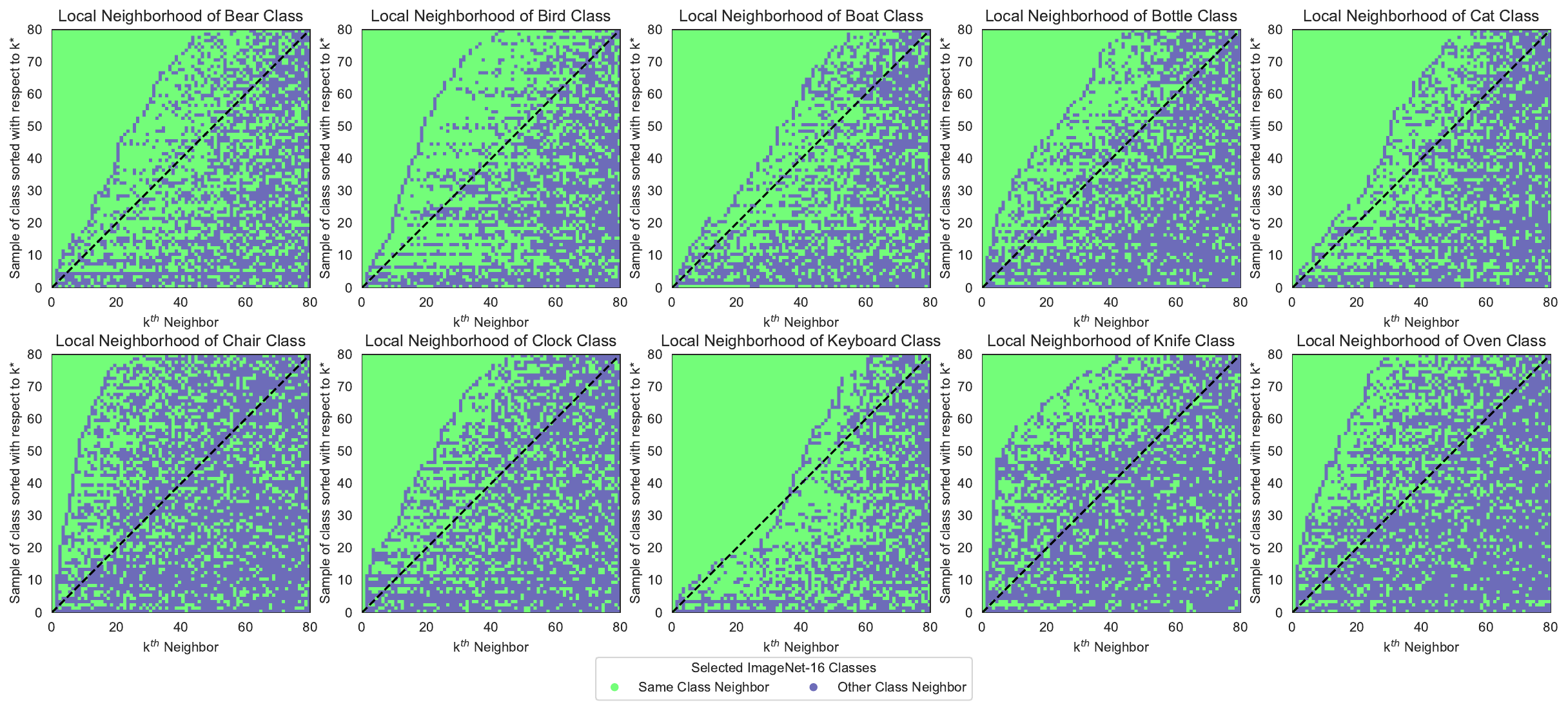}
        \caption{
        We visualize the neighbor distribution of all samples of a class for ResNet-50 trained on ImageNet-1K and Stylized ImageNet \cite{geirhos2018} (see \Tableref{tab:architectures}).
        The green color represents that the neighbor to the sample belongs to the same class as the testing sample, while the gray color represents that the neighbor belongs to a different class compared to the testing sample.
        A \fractured~distribution of samples will have different class neighbors above the diagonal (black dashed line);
        An \overlapped~distribution of samples will first different class neighbors around the diagonal, and;
        A \clustered~distribution of samples will have different class neighbors below the diagonal.
        }
        \vspace{0.5em}
        \includegraphics[width=0.49\linewidth]{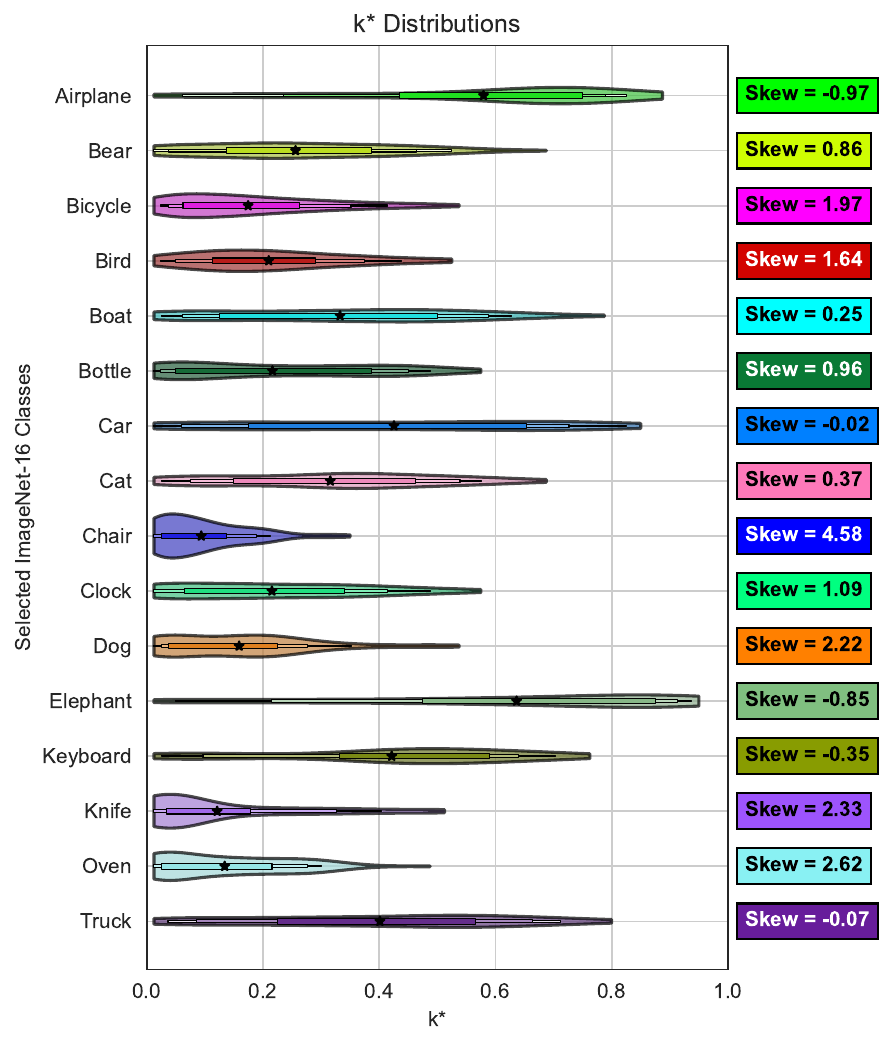}
        \rulesep
        \includegraphics[width=0.49\linewidth]{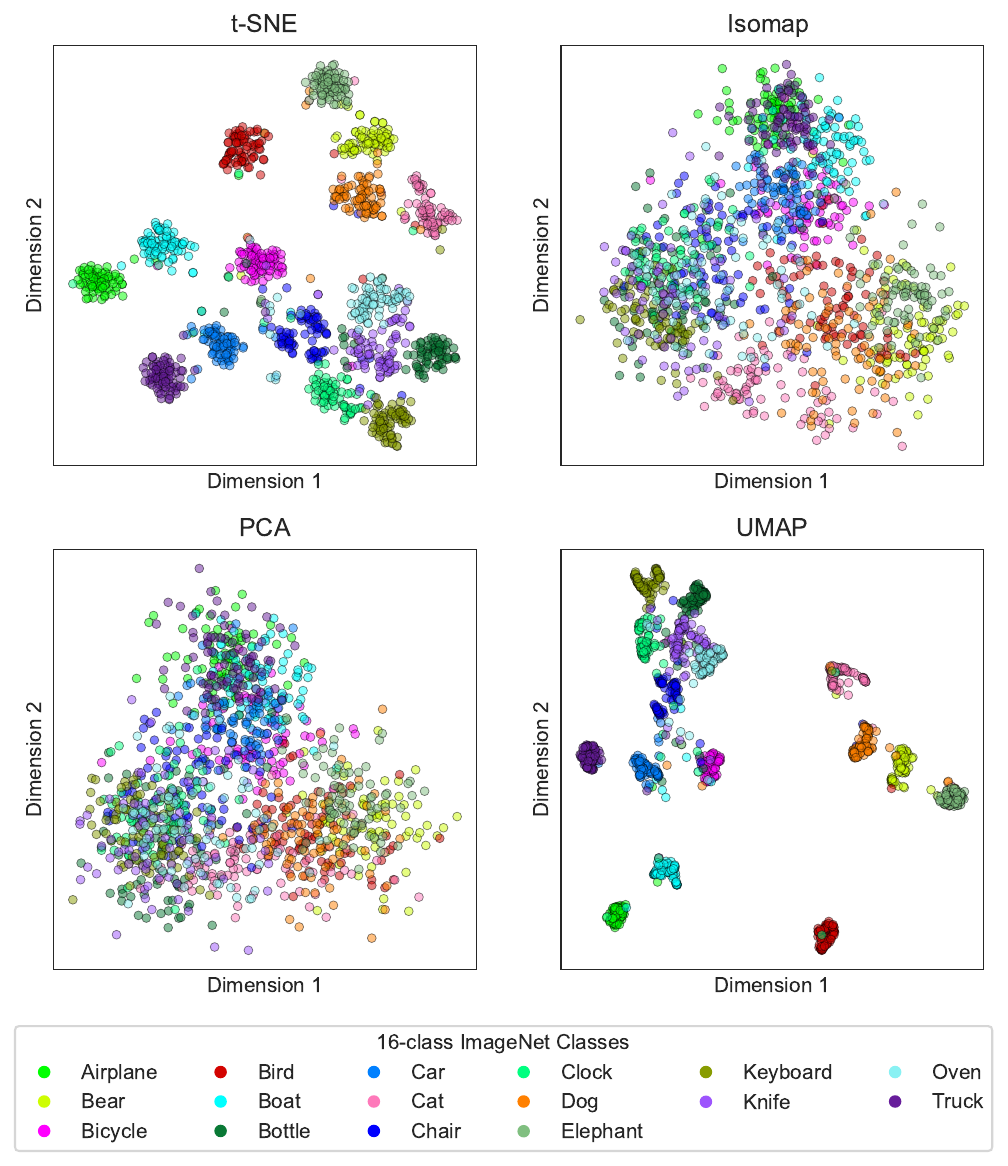}
        \caption{
               Visualization of the distribution of samples in latent space using, \figleft~{k*~distribution}, and \figright~Dimensionality Reduction techniques like t-SNE \figtopleft, Isomap \figtopright, PCA \figbottomleft, and UMAP \figbottomright~ of all classes of 16-class-ImageNet for the Logit Layer of ResNet-50 trained on ImageNet-1K and Stylized ImageNet \cite{geirhos2018} (see \Tableref{tab:training}).
        }
    \end{figure*}

    \clearpage
    \begin{figure*}[p]
        \centering
        \includegraphics[width=\linewidth]{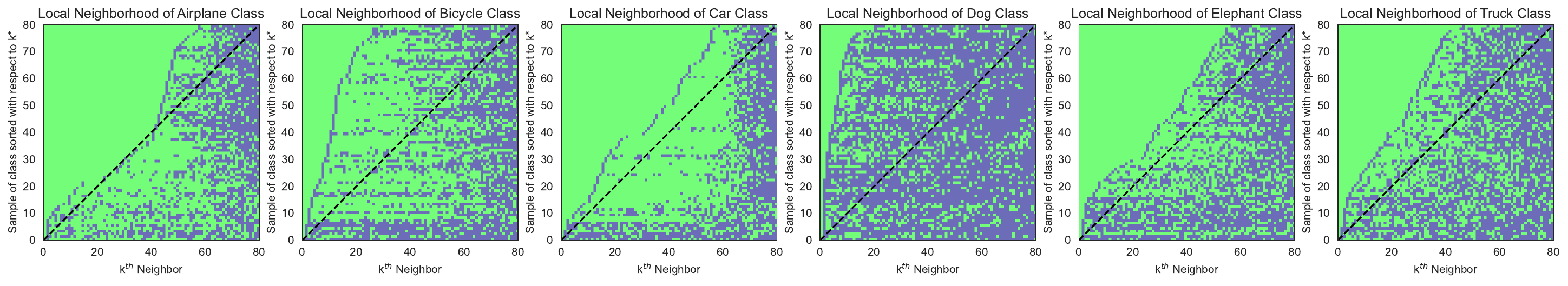} \\
        \includegraphics[width=\linewidth]{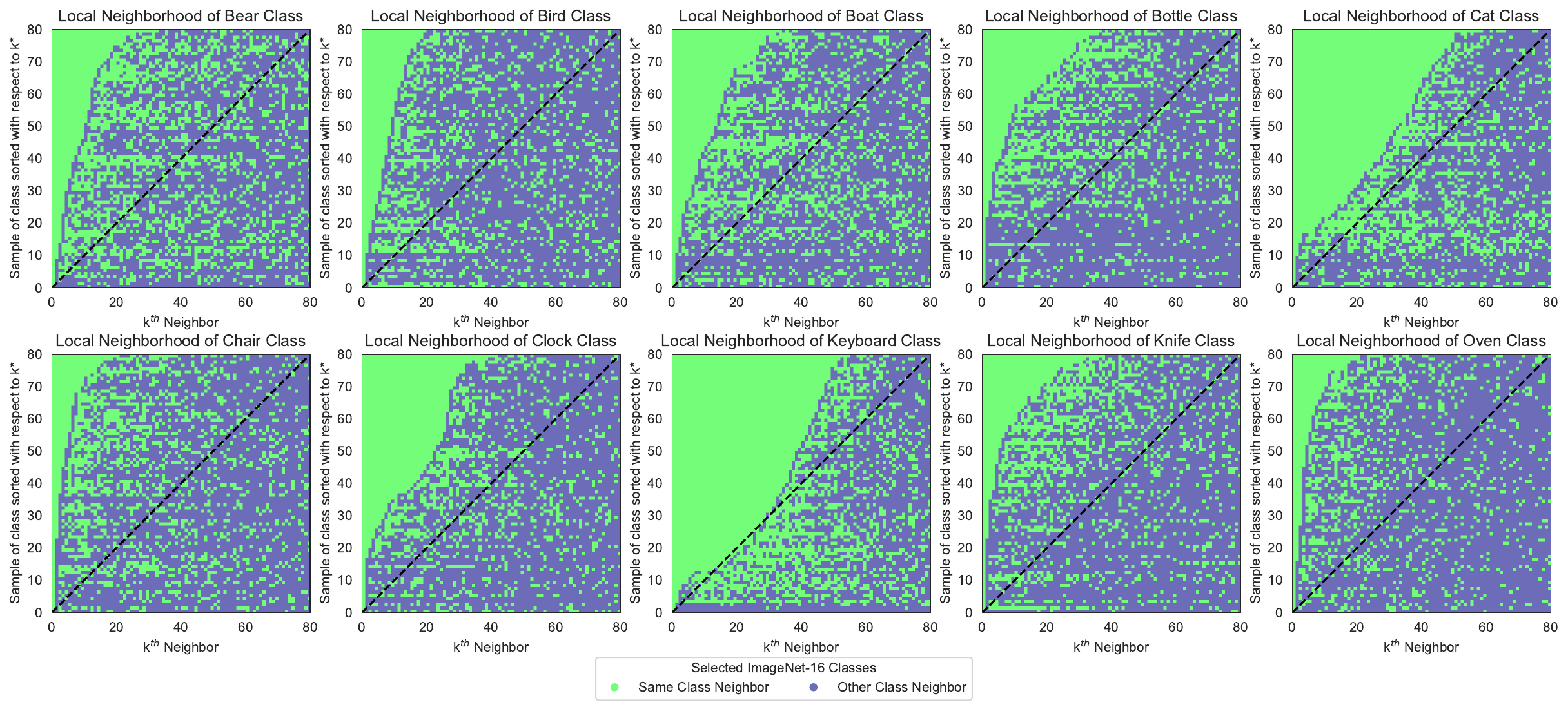}
        \caption{
        We visualize the neighbor distribution of all samples of a class for Adversarially Trained ResNet-50 \cite{dong2020Benchmarking} (see \Tableref{tab:adversarial}).
        The green color represents that the neighbor to the sample belongs to the same class as the testing sample, while the gray color represents that the neighbor belongs to a different class compared to the testing sample.
        A \fractured~distribution of samples will have different class neighbors above the diagonal (black dashed line);
        An \overlapped~distribution of samples will first different class neighbors around the diagonal, and;
        A \clustered~distribution of samples will have different class neighbors below the diagonal.
        }
        \vspace{0.5em}
        \includegraphics[width=0.49\linewidth]{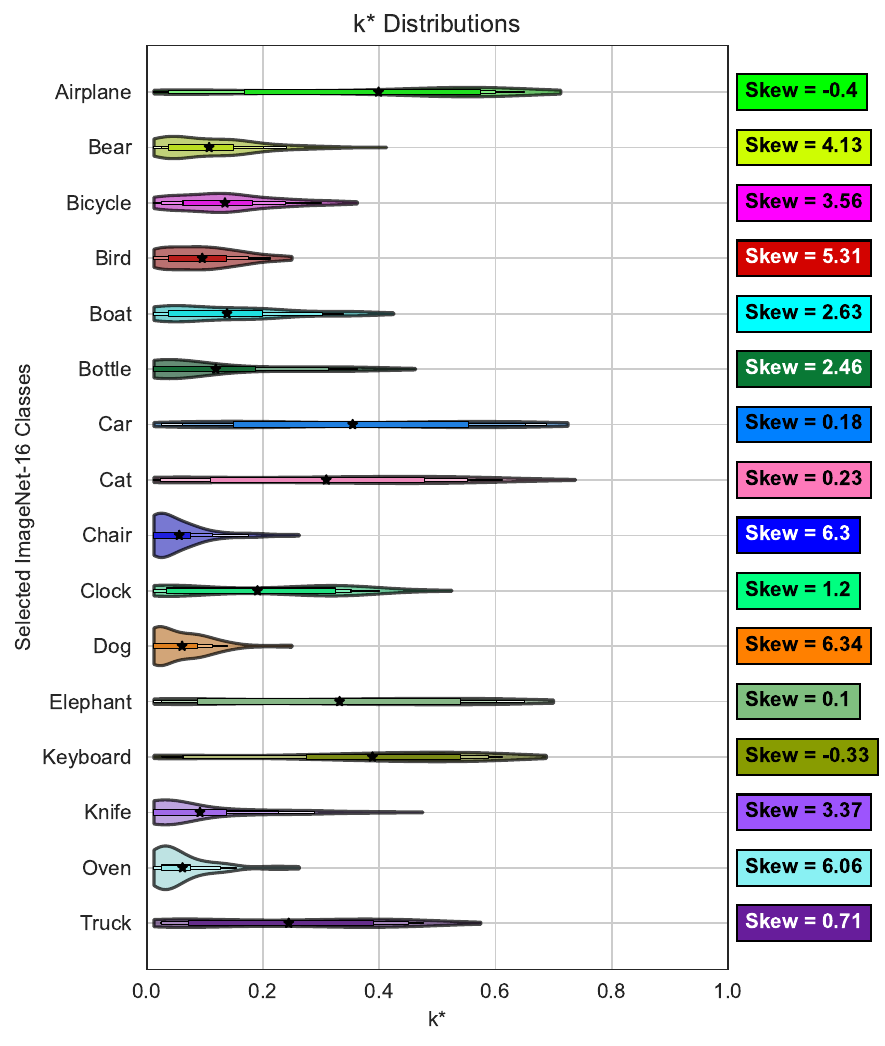}
        \rulesep
        \includegraphics[width=0.49\linewidth]{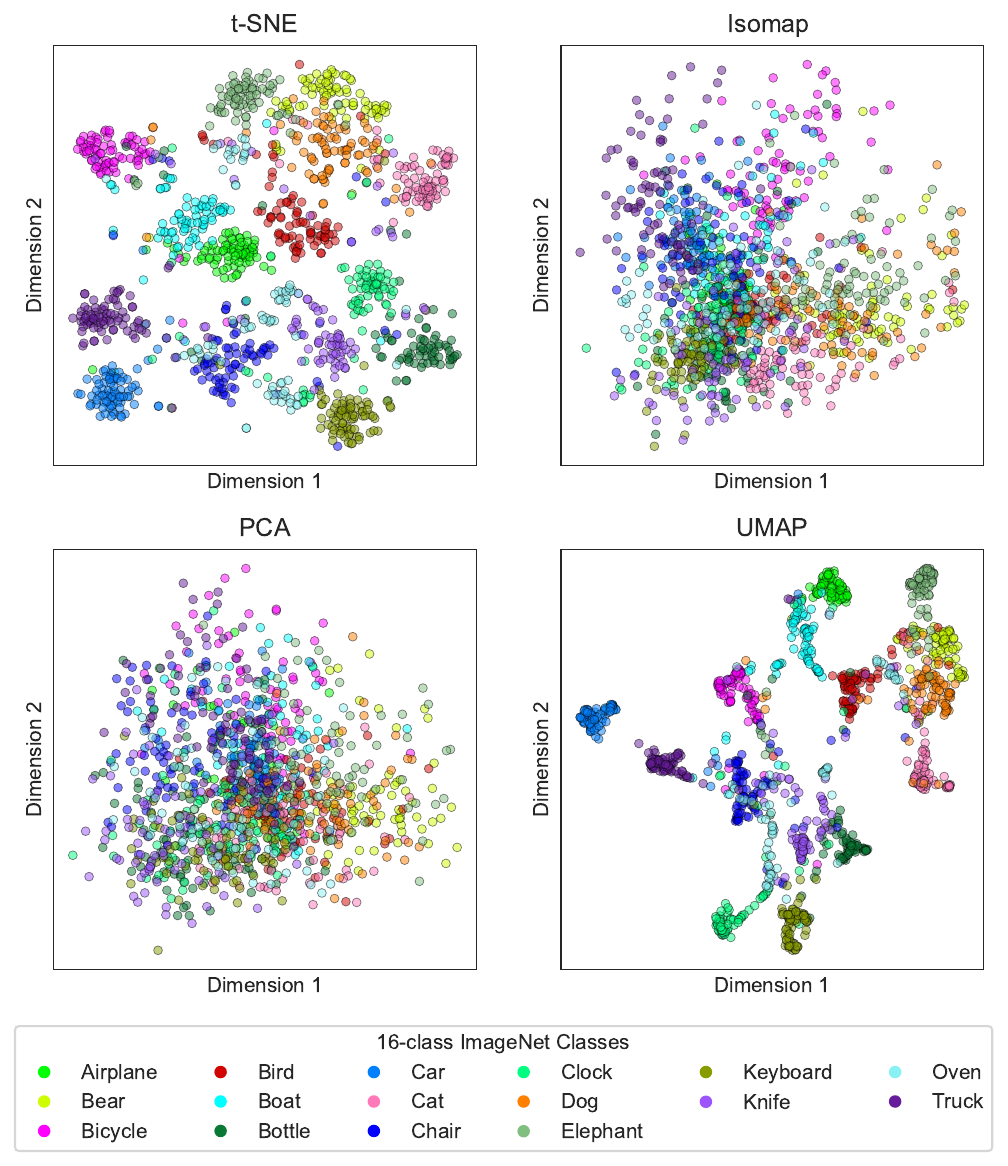}
        \caption{
               Visualization of the distribution of samples in latent space using, \figleft~{k*~distribution}, and \figright~Dimensionality Reduction techniques like t-SNE \figtopleft, Isomap \figtopright, PCA \figbottomleft, and UMAP \figbottomright~ of all classes of 16-class-ImageNet for the Logit Layer of Adversarially Trained ResNet-50 \cite{dong2020Benchmarking} (see \Tableref{tab:adversarial}).
        }
    \end{figure*}

    \clearpage
    \begin{figure*}[p]
        \centering
        \includegraphics[width=\linewidth]{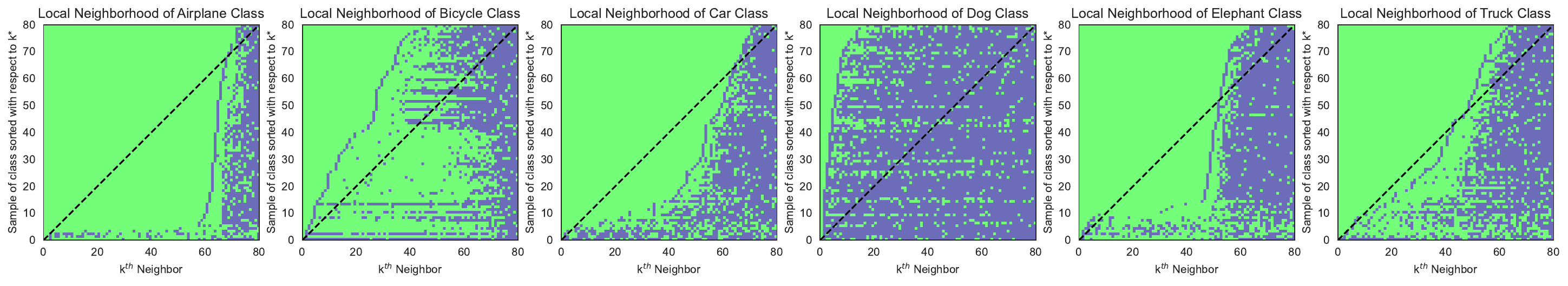} \\
        \includegraphics[width=\linewidth]{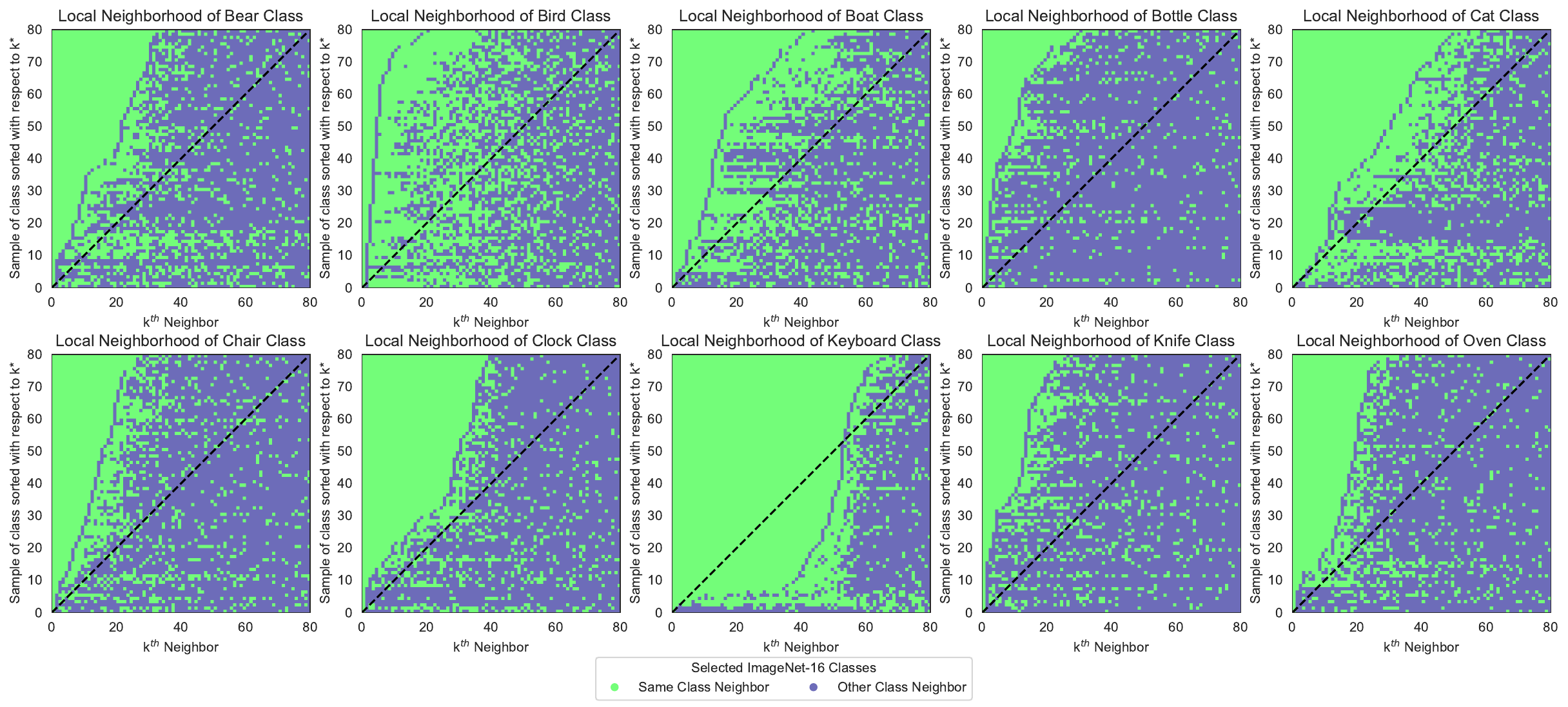}
        \caption{
        We visualize the neighbor distribution of all samples of a class for Adversarially Trained ViT-B \cite{dong2020Benchmarking} (see \Tableref{tab:adversarial}).
        The green color represents that the neighbor to the sample belongs to the same class as the testing sample, while the gray color represents that the neighbor belongs to a different class compared to the testing sample.
        A \fractured~distribution of samples will have different class neighbors above the diagonal (black dashed line);
        An \overlapped~distribution of samples will first different class neighbors around the diagonal, and;
        A \clustered~distribution of samples will have different class neighbors below the diagonal.
        }
        \vspace{0.5em}
        \includegraphics[width=0.49\linewidth]{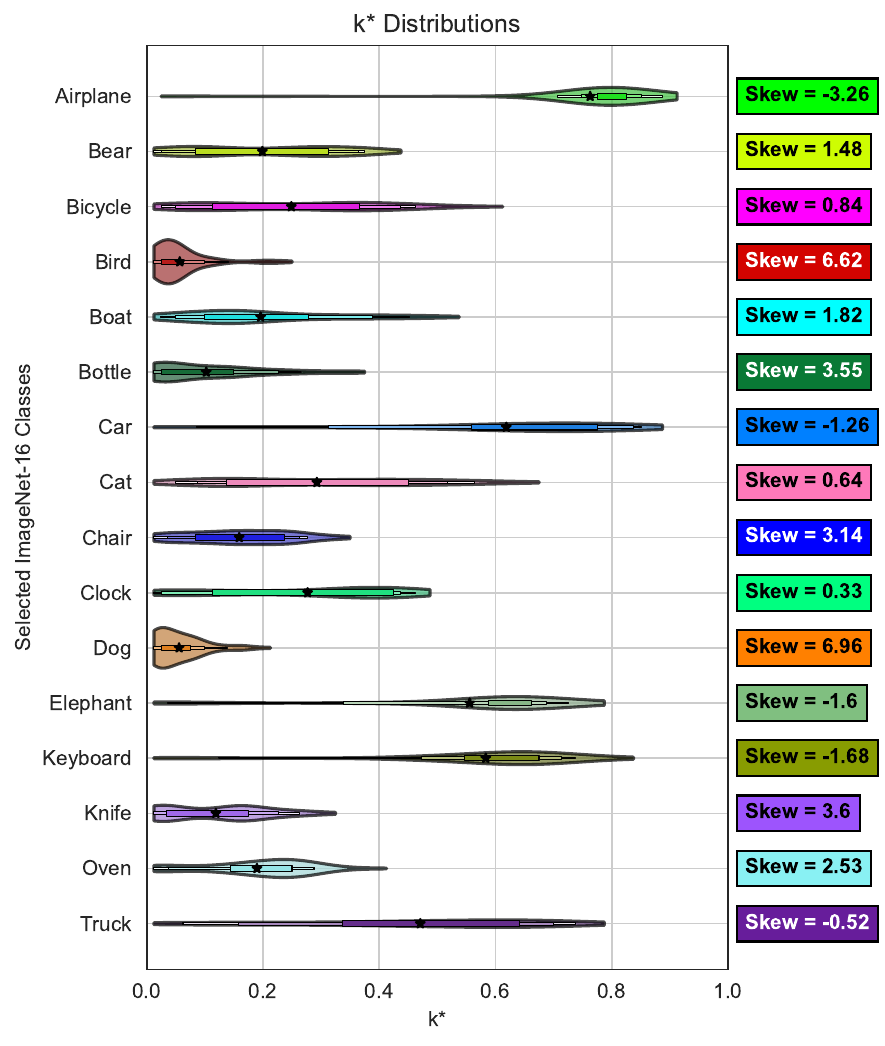}
        \rulesep
        \includegraphics[width=0.49\linewidth]{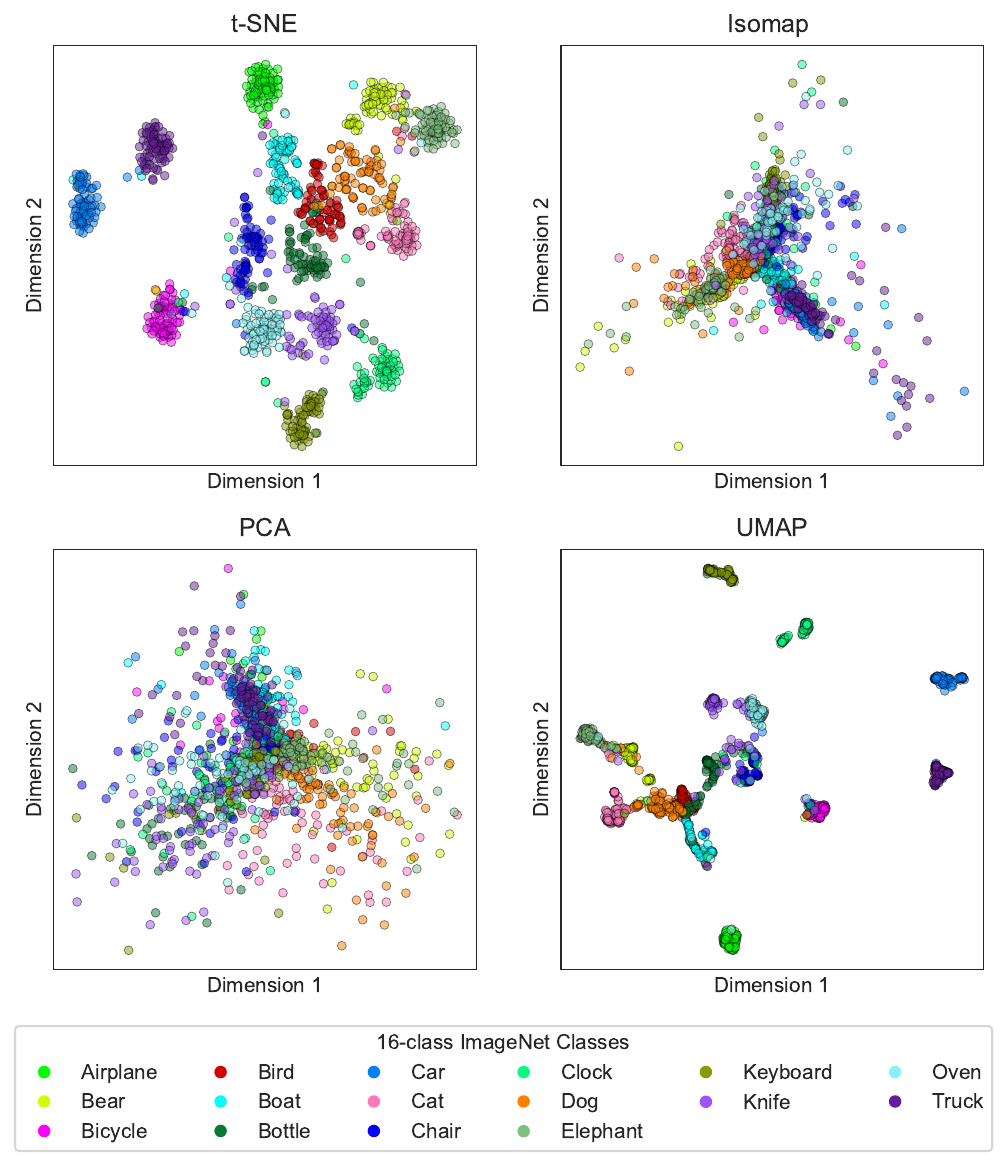}
        \caption{
               Visualization of the distribution of samples in latent space using, \figleft~{k*~distribution}, and \figright~Dimensionality Reduction techniques like t-SNE \figtopleft, Isomap \figtopright, PCA \figbottomleft, and UMAP \figbottomright~ of all classes of 16-class-ImageNet for the Logit Layer of Adversarially Trained ViT-B \cite{dong2020Benchmarking} (see \Tableref{tab:adversarial}).
        }
    \end{figure*}

    \clearpage
    \begin{figure*}[p]
        \centering
        \includegraphics[width=\linewidth]{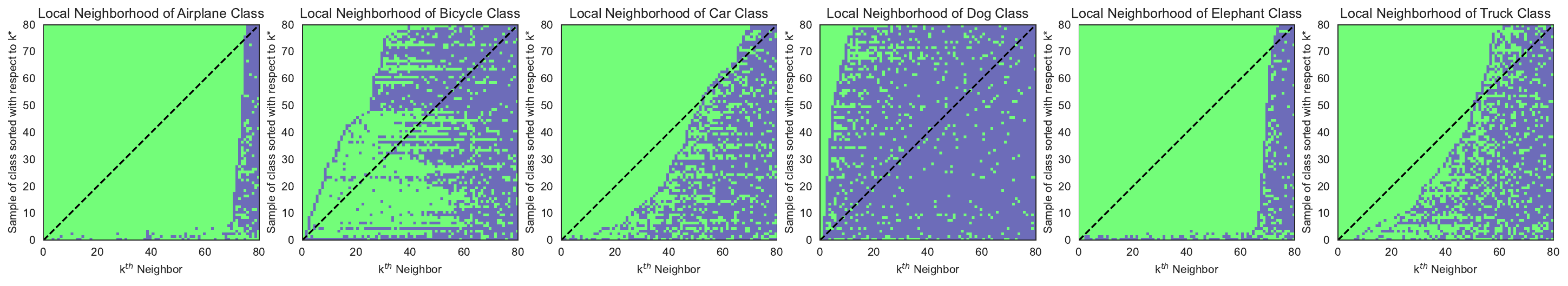} \\
        \includegraphics[width=\linewidth]{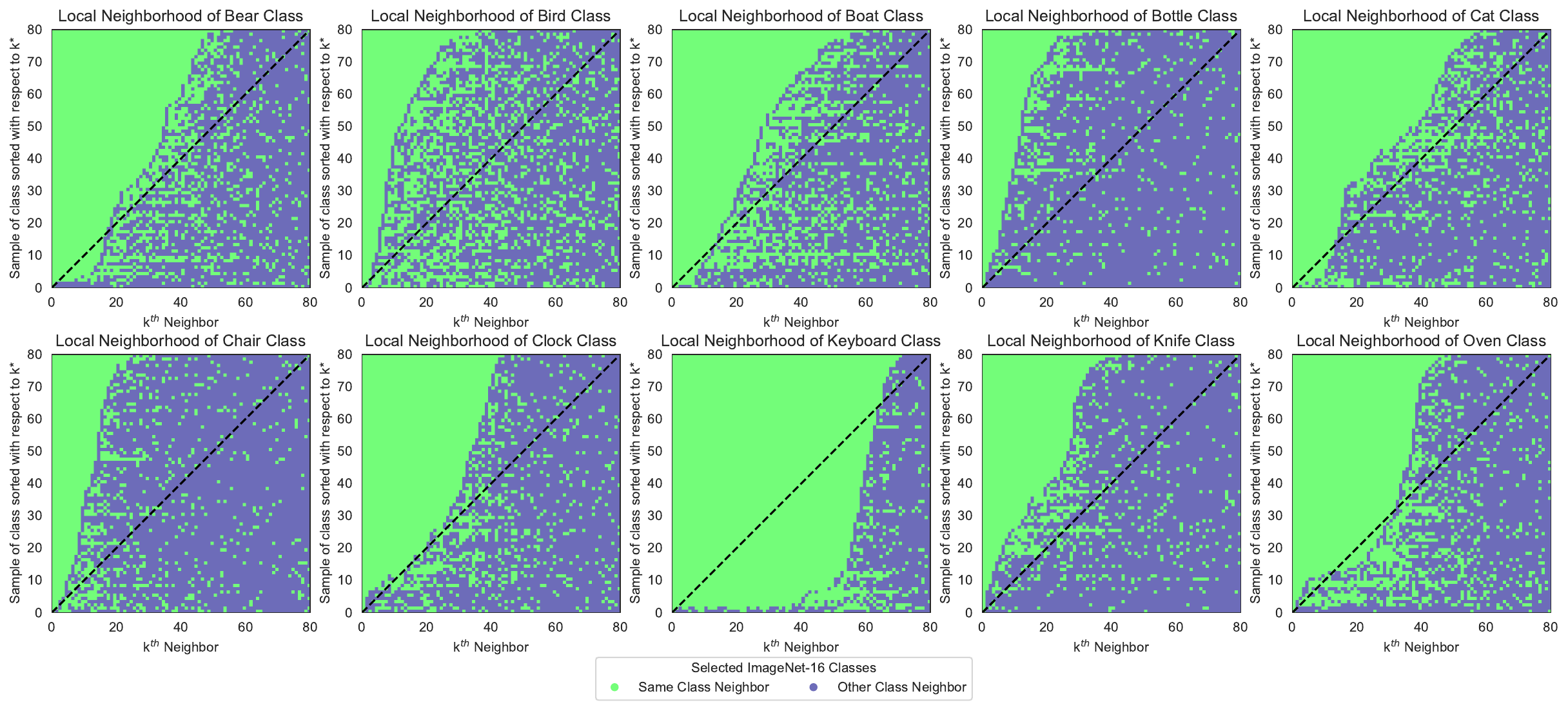}
        \caption{
        We visualize the neighbor distribution of all samples of a class for Standard Trained WideResNet-50 \cite{zagoruyko2017Widea} (see \Tableref{tab:adversarial}).
        The green color represents that the neighbor to the sample belongs to the same class as the testing sample, while the gray color represents that the neighbor belongs to a different class compared to the testing sample.
        A \fractured~distribution of samples will have different class neighbors above the diagonal (black dashed line);
        An \overlapped~distribution of samples will first different class neighbors around the diagonal, and;
        A \clustered~distribution of samples will have different class neighbors below the diagonal.
        }
        \vspace{0.5em}
        \includegraphics[width=0.49\linewidth]{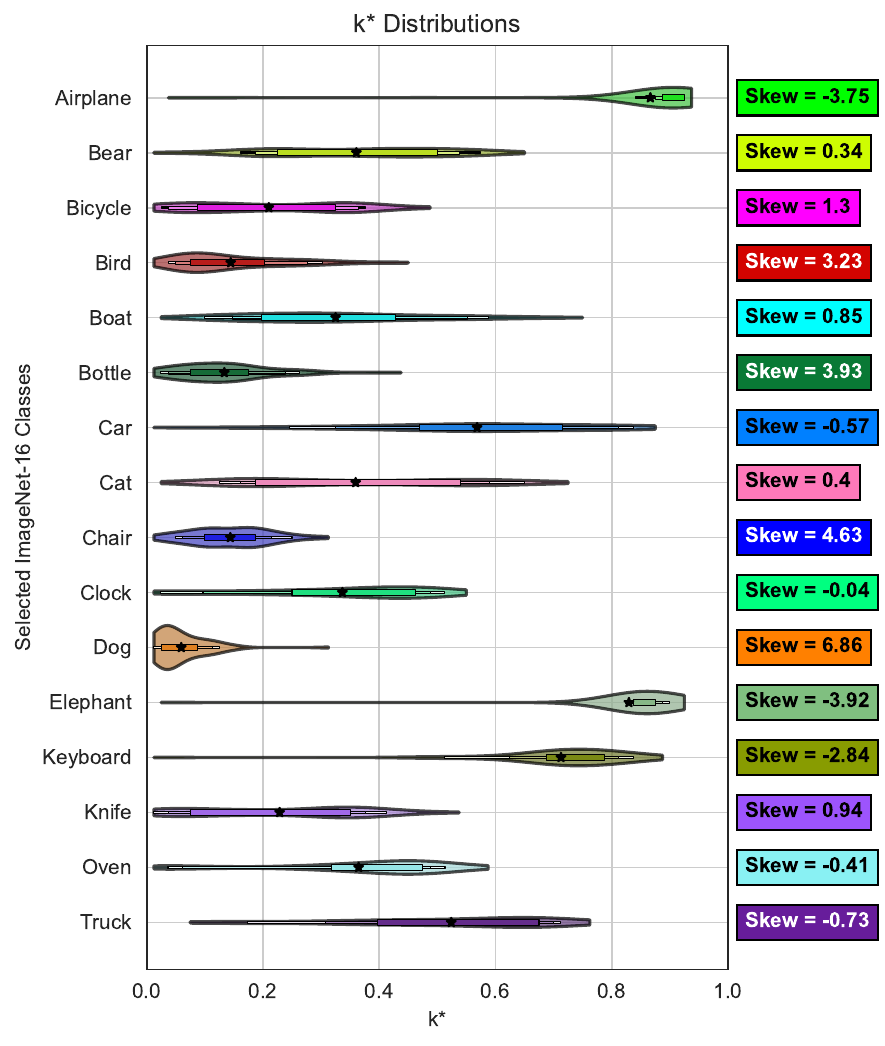}
        \rulesep
        \includegraphics[width=0.49\linewidth]{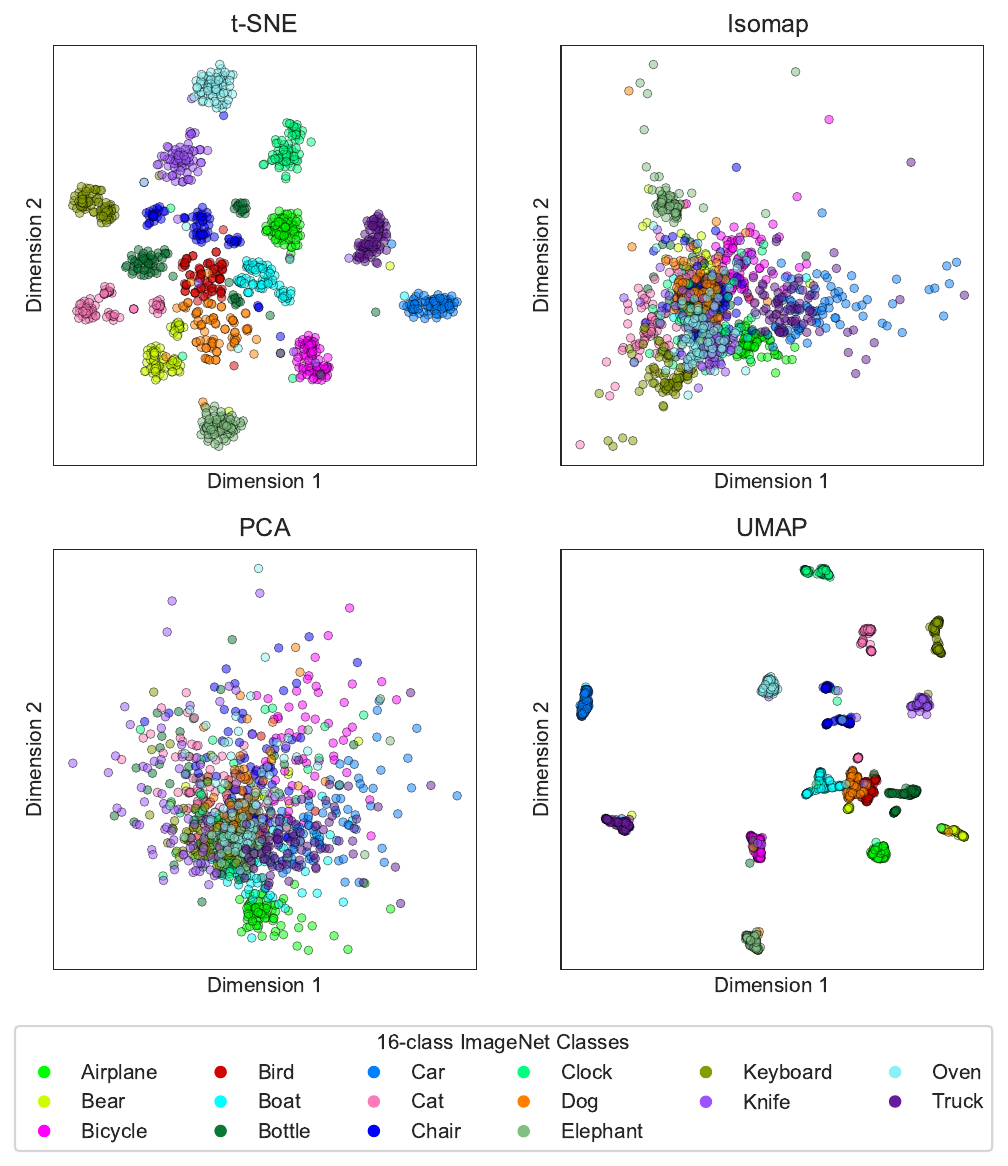}
        \caption{
               Visualization of the distribution of samples in latent space using, \figleft~{k*~distribution}, and \figright~Dimensionality Reduction techniques like t-SNE \figtopleft, Isomap \figtopright, PCA \figbottomleft, and UMAP \figbottomright~ of all classes of 16-class-ImageNet for the Logit Layer of Standard Trained WideResNet-50 \cite{zagoruyko2017Widea} (see \Tableref{tab:adversarial}).
        }
    \end{figure*}

    \clearpage
    \begin{figure*}[p]
        \centering
        \includegraphics[width=\linewidth]{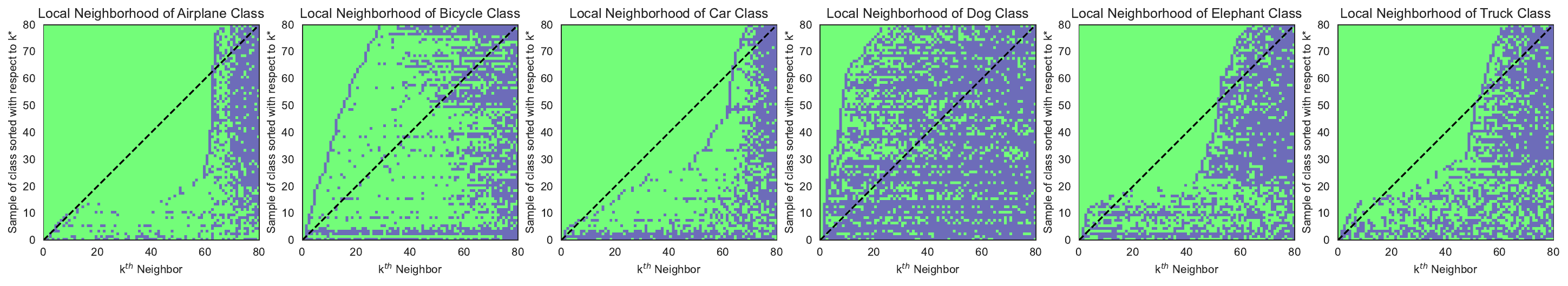} \\
        \includegraphics[width=\linewidth]{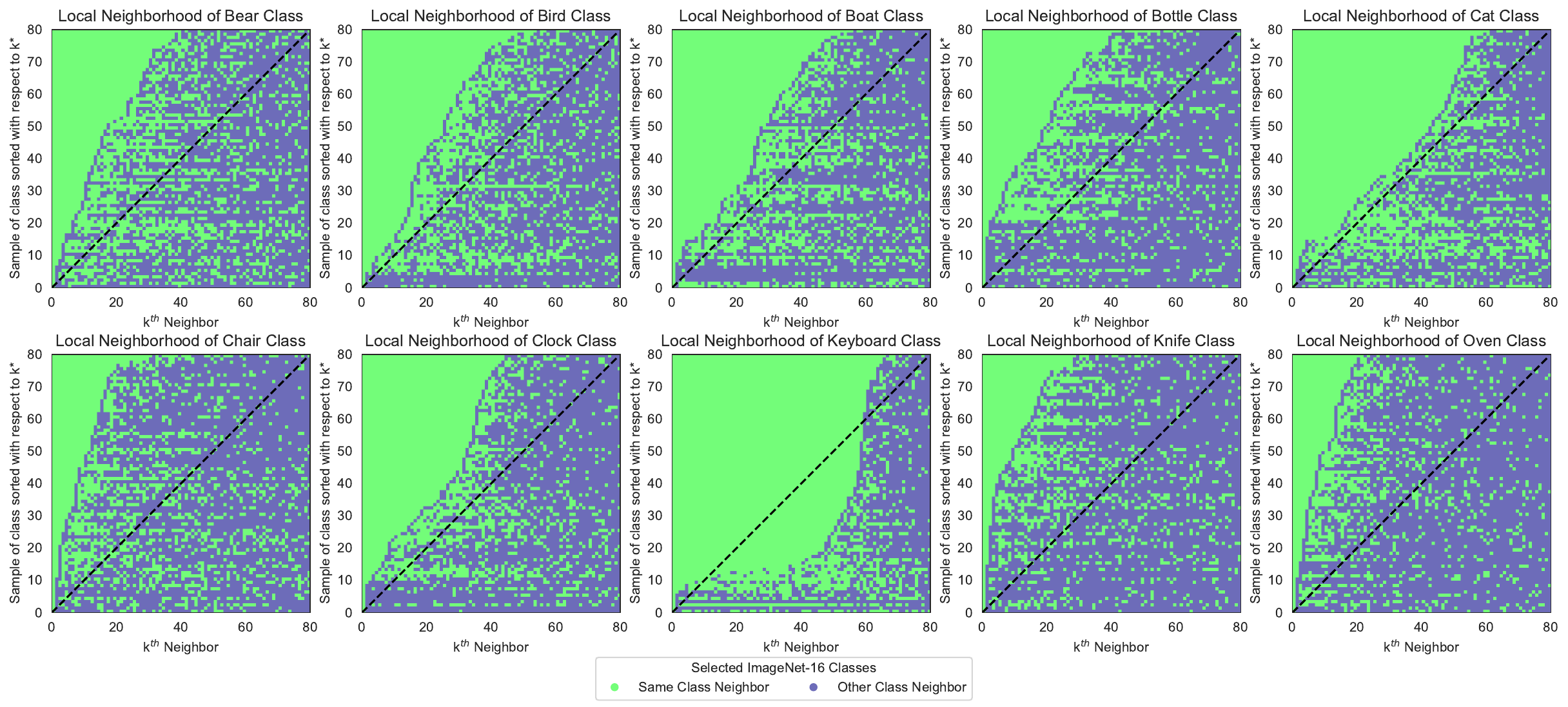}
        \caption{
        We visualize the neighbor distribution of all samples of a class for Adversarially Trained WideResNet-50 \cite{dong2020Benchmarking} (see \Tableref{tab:adversarial}).
        The green color represents that the neighbor to the sample belongs to the same class as the testing sample, while the gray color represents that the neighbor belongs to a different class compared to the testing sample.
        A \fractured~distribution of samples will have different class neighbors above the diagonal (black dashed line);
        An \overlapped~distribution of samples will first different class neighbors around the diagonal, and;
        A \clustered~distribution of samples will have different class neighbors below the diagonal.
        }
        \vspace{0.5em}
        \includegraphics[width=0.49\linewidth]{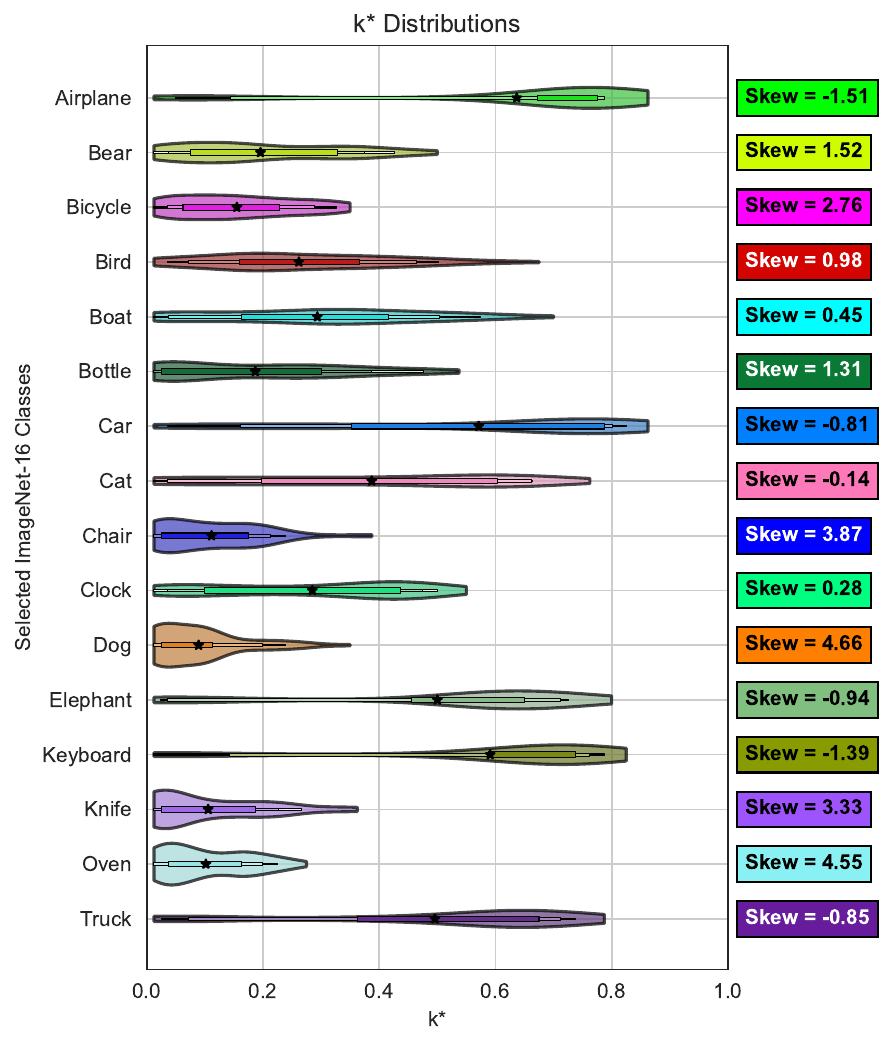}
        \rulesep
        \includegraphics[width=0.49\linewidth]{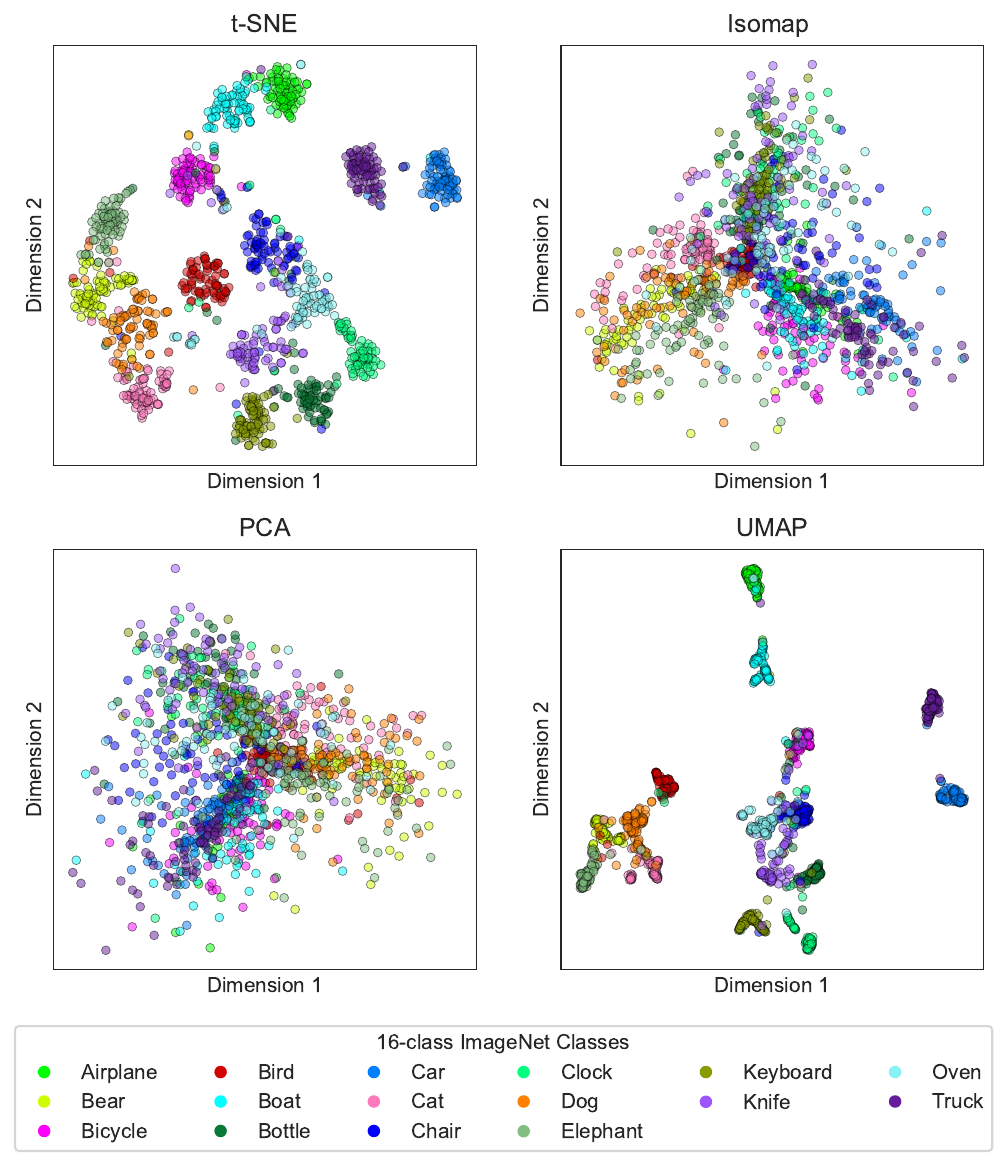}
        \caption{
               Visualization of the distribution of samples in latent space using, \figleft~{k*~distribution}, and \figright~Dimensionality Reduction techniques like t-SNE \figtopleft, Isomap \figtopright, PCA \figbottomleft, and UMAP \figbottomright~ of all classes of 16-class-ImageNet for the Logit Layer of Adversarially Trained WideResNet-50 \cite{dong2020Benchmarking} (see \Tableref{tab:adversarial}).
        }
    \end{figure*}

\end{document}